\documentclass{article}

\PassOptionsToPackage{numbers, compress}{natbib}
 \usepackage[preprint]{neurips_2026}

\usepackage[utf8]{inputenc} % allow utf-8 input
\usepackage[T1]{fontenc}    % use 8-bit T1 fonts
\usepackage{hyperref}       % hyperlinks
\usepackage{url}            % simple URL typesetting
\usepackage{booktabs}       % professional-quality tables
\usepackage{amsfonts}       % blackboard math symbols
\usepackage{nicefrac}       % compact symbols for 1/2, etc.
\usepackage{microtype}      % microtypography
\usepackage{xcolor}         % colors
\usepackage{amsmath}
\usepackage{amsfonts}
\usepackage{amssymb}
\usepackage{bbm}
\usepackage{graphicx}
\usepackage{subcaption}
\usepackage{booktabs}
\usepackage{multirow}
\usepackage[table]{xcolor}
\definecolor{lightcyan}{RGB}{224,255,255}
\title{Rethinking Data Selection as Distribution Matching on Semantic Manifolds}

\author{%
\begin{tabular}{c}
\textbf{Peng Sun}$^{1}$ \quad
\textbf{Yi Yang}$^{1}$ \quad
\textbf{Mingshuo Chen}$^{2}$ \quad
\textbf{Chunxiao Li}$^{3}$ \\
\textbf{Yao Zhu}$^{4}$ \quad
\textbf{Huawen Shen}$^{5}$ \quad
\textbf{Tianfan Fu}$^{1}$\thanks{Corresponding author.}
\end{tabular}
\\[0.8em]
\small
$^{1}$Nanjing University \quad
$^{2}$Beijing University of Posts and Telecommunications
\\
\small
$^{3}$Fudan University \quad
$^{4}$Zhejiang University \quad
$^{5}$Institute of Information Engineering
}

\begin{document}

\maketitle

\begin{abstract}
Instruction tuning of large language models and large vision language models relies on massive training data, which incurs substantial computational and storage costs. 
Meanwhile, training data often contains redundancy, noise, and low quality samples. 
Existing data selection methods mostly rely on diversity, importance, or a combination of both, without explicitly constraining the distributional consistency between the selected subset and the full data in the embedding space.
We propose QMAC, a quality aware multiband characteristic function matching method that formulates data selection as a joint optimization problem of distribution matching and quality filtering. 
QMAC matches the empirical characteristic functions of the selected subset and the full data across multiple frequency scales, capturing both global semantic coverage and local perturbations, while using task related quality scores to filter noisy and low quality samples. 
We systematically evaluate QMAC on 4 data types, 8 target models, and 44 benchmarks. 
Experimental results show that QMAC outperforms existing data selection baselines on standard image-text instruction data, substantially surpasses random selection on standard text instruction, image-text reasoning, and text reasoning data, and exceeds full data training in multiple model and budget combinations. The code will be made publicly available upon acceptance of the paper.
\end{abstract}

\section{Introduction}
\label{sec:introduction}

As the scale of training data for large language models and large vision language models continues to grow, instruction tuning on the full dataset incurs substantial computational and storage costs~\cite{safaei2025filter,zhang2025survey}. 
Meanwhile, large training datasets often contain redundancy, noise, and low quality samples, meaning that more data does not necessarily lead to better training outcomes~\cite{chen2023alpagasus,liu2025less}. Data selection therefore aims to identify an efficient subset from the full dataset, reducing training cost while approaching or even surpassing the performance of full data training~\cite{dohmatob2025less,liu2025less,xia2024less}. We observe that instruction tuning data does not form unstructured scatter points in the high dimensional embedding space. 
Instead, it tends to concentrate around several low dimensional semantic manifolds. 
Detailed data analysis is provided in Appendix~\ref{app:data_analysis}. 
This suggests that a desirable data selection method should not only identify valuable samples, but also preserve the smooth distributional structure of the original data along the manifolds under a limited budget~\cite{bukharin-etal-2024-data,chen-etal-2025-mig,yang2025measuring}.

\begin{figure}[t]
  \centering
  \begin{subfigure}[t]{0.32\linewidth}
    \centering
    \includegraphics[width=\linewidth]{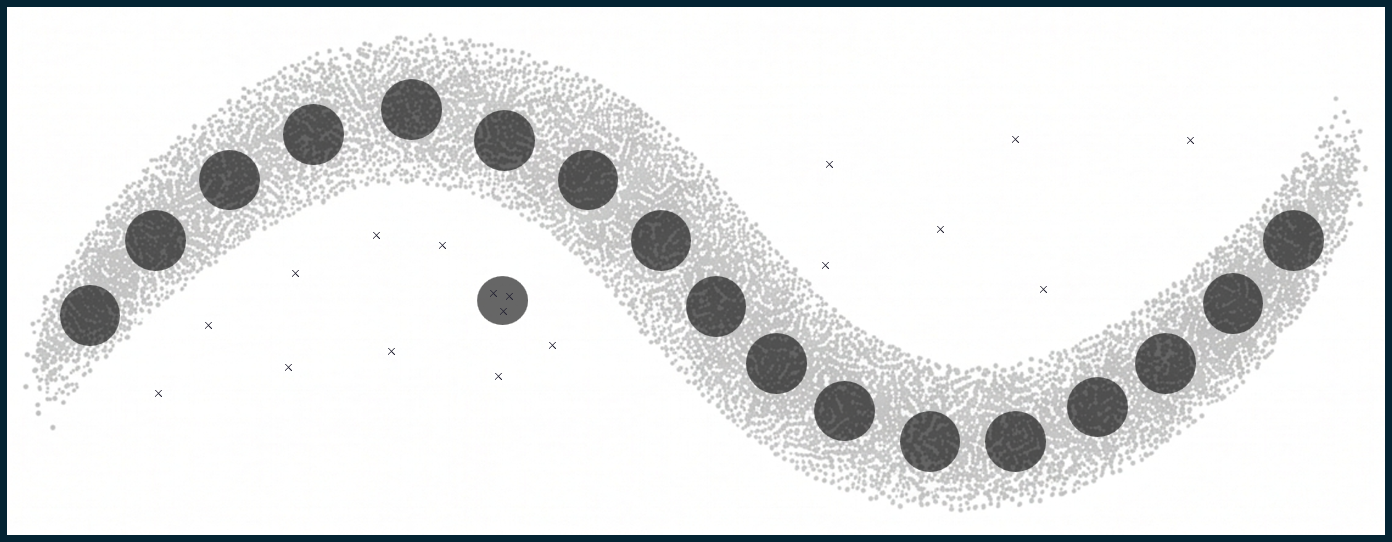}
    \caption{Diversity}
    \label{fig:manifold_diversity}
  \end{subfigure}
  \hfill
  \begin{subfigure}[t]{0.32\linewidth}
    \centering
    \includegraphics[width=\linewidth]{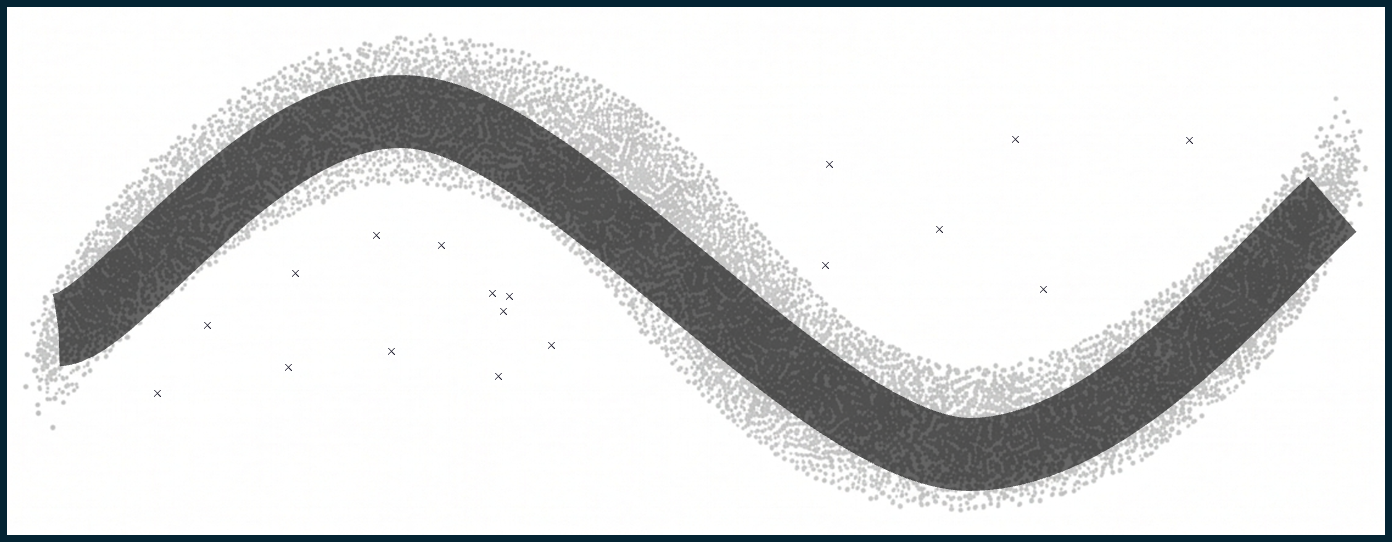}
    \caption{Importance}
    \label{fig:manifold_importance}
  \end{subfigure}
  \hfill
  \begin{subfigure}[t]{0.32\linewidth}
    \centering
    \includegraphics[width=\linewidth]{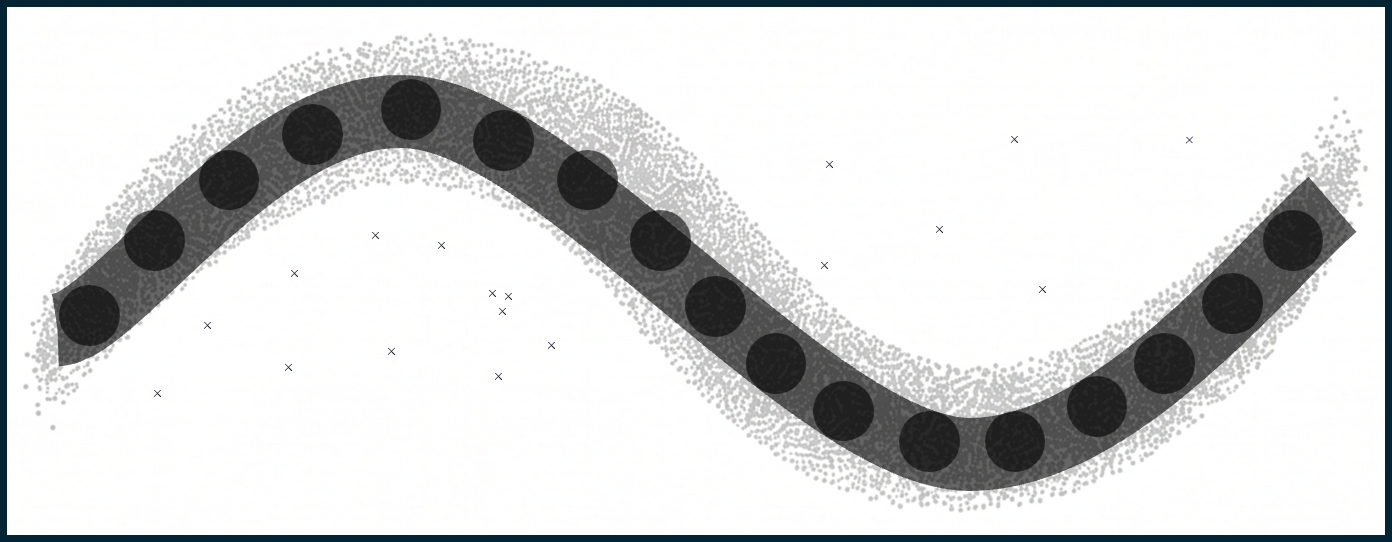}
    \caption{Hybrid}
    \label{fig:manifold_hybrid}
  \end{subfigure}

  \caption{Conceptual illustration of existing data selection paradigms. Existing methods often preserve coverage indirectly through diversity or importance heuristics, which may fail to maintain distributional consistency on the underlying manifolds.}
  \label{fig:motivation}
\end{figure}

Figure~\ref{fig:motivation} illustrates three representative data selection paradigms. 
Gray points denote the manifold distribution of the full dataset in the embedding space, while the darkest shaded regions indicate where selected samples concentrate.
Diversity based methods usually cover the data distribution by selecting discrete representative points through clustering combined with deduplication or intra cluster sampling~\cite{abbas2023semdedup,bi2025prism,deb2025fishersft,maharana2023d2,sorscher2022beyond}.
However, such discrete approximation struggles to preserve smooth density variations on continuous manifolds, and may retain low quality or outlier samples. 
Importance based methods estimate sample value using signals such as loss, gradients, attention changes, model feedback, or quality scores~\cite{he2025learning,hu2025donod,liu2025less,ma2025mllm,paul2021deep,sun2026does,toibazar2025trust,wang2026opus,wu2026scalselect,wu2024icons,xia2024less}. 
Their selections, however, can be biased toward local high utility regions induced by the scoring function, thereby damaging global coverage. 
Hybrid methods combine importance and diversity signals, but they often still rely on clustering to maintain coverage~\cite{chen2024your,lee2024concept,yan2025coido,yu2025mastering}. 
Existing methods therefore do not directly enforce distributional consistency between the selected subset and the full dataset on the underlying manifolds. 
The key limitation is the lack of a direct distribution matching objective that aligns the empirical distributions of the subset and the full data in the embedding space while filtering low quality samples.

To address this issue, we propose QMAC, a quality aware multiband characteristic function matching method for data selection. 
Instead of preserving diversity indirectly through clustering or neighborhood heuristics, QMAC directly matches the empirical distribution of the candidate subset to that of the full dataset in the embedding space. 
Specifically, QMAC uses multiband characteristic function distance to compare the two distributions across different frequency scales, capturing both global semantic coverage and local perturbations. 
It also incorporates task related quality scores as auxiliary signals to filter low quality and noisy samples.

Notably, existing data selection methods are usually designed for a single data type or a limited range of data types~\cite{bi2025prism,wu2026scalselect,xia2024less,yan2025coido}, making it difficult to form a unified data selection framework. 
In contrast, QMAC shows stable advantages across four settings: standard image-text instruction data, standard text instruction data, image-text reasoning data, and text reasoning data. 
It not only outperforms existing data selection baselines on standard image-text instruction data, but also consistently surpasses random selection in other settings and even outperforms full-data training across multiple model–budget combinations, demonstrating broad applicability across data types.
Our main contributions are as follows:
\begin{itemize}
\item We revisit data selection from the perspective of semantic manifold preservation, and identify a key limitation of existing diversity based, importance based, and hybrid methods: they only indirectly approximate distribution preservation, without explicitly enforcing distributional consistency in the embedding space.

\item We propose QMAC, which explicitly matches the empirical distributions of the selected subset and the full dataset through multiband characteristic function distance, while using task related quality scores to filter low quality samples, thereby preserving the manifold distribution and improving sample quality.

\item We systematically evaluate QMAC on 4 data types, 8 target models, and 44 benchmarks. 
Results show that QMAC achieves more stable advantages over random selection and existing data selection baselines, demonstrating its potential as a general data selection framework.
\end{itemize}

\section{Method}
\label{methodmethod}

\subsection{Preliminaries: Characteristic Function Distance}
\label{sec:prelim_cfd}

Characteristic functions~\cite{ansari2020characteristic} provide a frequency domain representation of probability distributions. 
For a distribution $P$ defined on $\mathbb{R}^d$, its characteristic function is given by
\begin{equation}
    \varphi_P(\mathbf{t})
    =
    \mathbb{E}_{\mathbf{x}\sim P}
    \left[
        \exp\left(\mathrm{i}\langle \mathbf{t}, \mathbf{x}\rangle\right)
    \right],
    \qquad
    \mathbf{t}\in\mathbb{R}^d,
\end{equation}
where $\mathrm{i}=\sqrt{-1}$ and $\mathbf{t}$ denotes a frequency vector. 
Characteristic functions uniquely determine probability distributions. 
That is, two distributions are identical if their characteristic functions agree over the entire frequency space. 
This property makes characteristic functions a natural tool for measuring discrepancies between distributions.

Characteristic Function Distance (CFD)~\cite{ansari2020characteristic} measures the discrepancy between two distributions $P$ and $Q$ through the weighted squared difference between their characteristic functions. 
It is defined as
\begin{equation}
    \operatorname{CFD}_{\omega}^{2}(P,Q)
    =
    \mathbb{E}_{\mathbf{t}\sim\omega}
    \left[
        \left|
            \varphi_P(\mathbf{t})-\varphi_Q(\mathbf{t})
        \right|^2
    \right],
\end{equation}
where $\omega$ is a weighting distribution over the frequency space and determines the frequency regions where the two distributions are compared. 
When the support of $\omega$ covers $\mathbb{R}^d$, we have $\operatorname{CFD}_{\omega}(P,Q)=0$ if and only if $P=Q$.

In the finite sample setting, given a sample set $X=\{\mathbf{x}_i\}_{i=1}^{n}$, the empirical characteristic function is
\begin{equation}
    \widehat{\varphi}_{X}(\mathbf{t})
    =
    \frac{1}{n}
    \sum_{i=1}^{n}
    \exp\left(\mathrm{i}\langle \mathbf{t}, \mathbf{x}_i\rangle\right).
\end{equation}
By drawing $M$ frequency points $\{\mathbf{t}_m\}_{m=1}^{M}$ from $\omega$, we obtain the empirical estimate of CFD as
\begin{equation}
    \widehat{\operatorname{CFD}}_{\omega}^{2}(X,Y)
    =
    \frac{1}{M}
    \sum_{m=1}^{M}
    \left|
        \widehat{\varphi}_{X}(\mathbf{t}_m)
        -
        \widehat{\varphi}_{Y}(\mathbf{t}_m)
    \right|^2.
\end{equation}
This formulation avoids explicit density estimation in high dimensional spaces. 
It only requires comparing empirical distributions at a finite number of randomly sampled frequency points, which gives a computational cost that scales linearly with the number of samples. 
This makes CFD well suited for subset selection on large data collections.

\subsection{Problem Formulation}
\label{sec:problem_formulation}

Given a dataset $D$, data selection seeks a budgeted subset $D'\subseteq D$ with $|D'|=K$, such that training on $D'$ approaches or even surpasses the performance of training on $D$. 
Let $f(\cdot)$ denote the resulting model performance. 
We formulate the objective as
\begin{equation}
    \max_{D'\subseteq D} f(D')
    \quad
    \mathrm{s.t.}
    \quad
    |D'| = K .
\end{equation}
To address this problem, we propose QMAC, which formulates data selection as a joint problem of distribution matching and sample quality filtering. 
The details of our method are introduced in the following sections.

\subsection{QMAC: Quality Aware Multiband Adaptive CFD Selection}
\label{sec:qmac}

We introduce QMAC, a quality aware multiband CFD method for data selection. 
Given a fixed budget $K$, QMAC learns sample selection variables that encourage the selected subset to match the full dataset distribution in the embedding space while achieving higher data quality than random selection.

\begin{figure}[t]
  \centering
  \includegraphics[width=1.0\linewidth]{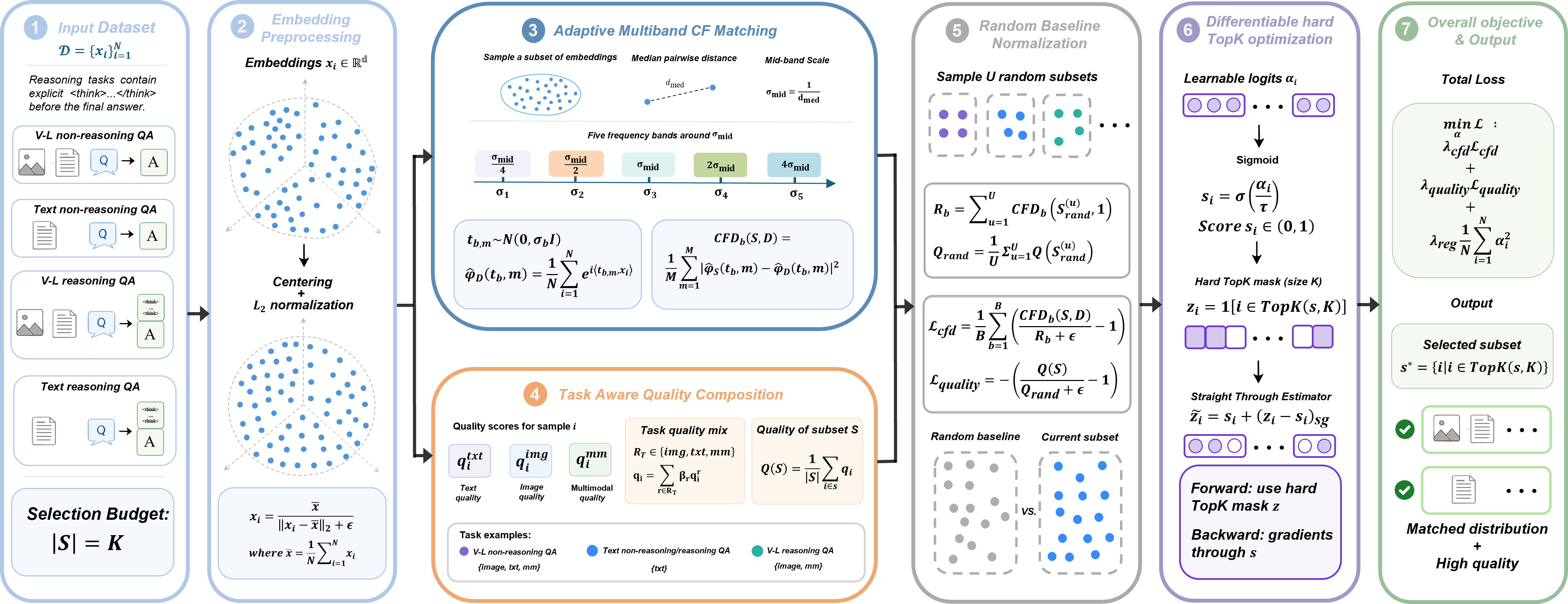}
  \caption{Overview of the QMAC pipeline.}
  \label{fig:qmac_pipeline}
\end{figure}

Figure~\ref{fig:qmac_pipeline} illustrates the overall pipeline, which consists of seven steps: input dataset specification, embedding preprocessing, adaptive multiband characteristic function matching, task aware quality composition, random baseline normalization, differentiable hard TopK optimization, and final objective optimization with subset output. This section focuses on describing the main procedure of QMAC, while detailed experimental settings and hyperparameters are provided in Appendix~\ref{app:hyperparameters}.

\paragraph{Embedding preprocessing}

Given a dataset $D=\{x_i\}_{i=1}^{N}$, we first compute its embedding representation $X=\{\mathbf{x}_i\}_{i=1}^{N}$, where $\mathbf{x}_i\in\mathbb{R}^{d}$. 
To stabilize frequency projection, we center the embeddings and apply $\ell_2$ normalization:
\begin{equation}
    \widetilde{\mathbf{x}}_i
    =
    \frac{\mathbf{x}_i-\bar{\mathbf{x}}}
    {\|\mathbf{x}_i-\bar{\mathbf{x}}\|_2+\epsilon},
    \qquad
    \bar{\mathbf{x}}
    =
    \frac{1}{N}
    \sum_{i=1}^{N}
    \mathbf{x}_i .
\end{equation}
In the sequel, we use $\mathbf{x}_i$ to denote the preprocessed embedding for simplicity.

\paragraph{Adaptive multiband characteristic function matching}

QMAC uses multiple frequency bands to match the full distribution at different scales. 
We estimate the median pairwise distance $d_{\mathrm{med}}$ from a small random subset of embeddings and set
\begin{equation}
    \sigma_{\mathrm{mid}}=\frac{1}{d_{\mathrm{med}}},
    \qquad
    \{\sigma_b\}_{b=1}^{5}
    =
    \left\{
    \frac{\sigma_{\mathrm{mid}}}{4},
    \frac{\sigma_{\mathrm{mid}}}{2},
    \sigma_{\mathrm{mid}},
    2\sigma_{\mathrm{mid}},
    4\sigma_{\mathrm{mid}}
    \right\}.
\end{equation}
For each band $b$, we sample $M$ frequency vectors
\begin{equation}
    \mathbf{t}_{b,m}
    \sim
    \mathcal{N}(\mathbf{0},\sigma_b^2 I),
    \qquad
    m=1,\ldots,M .
\end{equation}
Let $z_i\in\{0,1\}$ indicate whether the $i$th sample is selected. 
At frequency $\mathbf{t}_{b,m}$, the empirical characteristic functions of the full dataset and selected subset are
\begin{equation}
    \widehat{\varphi}_{D}(\mathbf{t}_{b,m})
    =
    \frac{1}{N}
    \sum_{i=1}^{N}
    e^{\mathrm{i}\langle \mathbf{t}_{b,m},\mathbf{x}_i\rangle},
    \qquad
    \widehat{\varphi}_{S}(\mathbf{t}_{b,m})
    =
    \frac{
    \sum_{i=1}^{N}
    z_i e^{\mathrm{i}\langle \mathbf{t}_{b,m},\mathbf{x}_i\rangle}
    }{
    \sum_{i=1}^{N} z_i
    } .
\end{equation}
The matching error on band $b$ is
\begin{equation}
    \mathrm{CFD}_{b}(S,D)
    =
    \frac{1}{M}
    \sum_{m=1}^{M}
    \left|
    \widehat{\varphi}_{S}(\mathbf{t}_{b,m})
    -
    \widehat{\varphi}_{D}(\mathbf{t}_{b,m})
    \right|^2 .
\end{equation}

\paragraph{Task aware quality composition}

For task $T$, QMAC assigns each sample a task dependent quality score by combining the quality dimensions used by the task:
\begin{equation}
    q_i
    =
    \sum_{r\in\mathcal{R}_{T}}
    \beta_r q_i^r,
    \qquad
    r\in\{\mathrm{img},\mathrm{txt},\mathrm{mm}\}.
\end{equation}
Here, $q_i^{\mathrm{img}}$, $q_i^{\mathrm{txt}}$, and $q_i^{\mathrm{mm}}$ denote image quality, text quality, and overall multimodal quality, respectively. 
$\mathcal{R}_{T}$ is the set of quality dimensions used for task $T$, and $\beta_r$ controls their relative weights. 
The average quality of the selected subset is
\begin{equation}
    Q(S)
    =
    \frac{1}{|S|}
    \sum_{i\in S} q_i .
\end{equation}

\paragraph{Random baseline normalization}

The CFD terms and the quality term can have different scales. 
We normalize them using random subsets under the same budget. 
Specifically, we sample $U$ random subsets $\{S_{\mathrm{rand}}^{(u)}\}_{u=1}^{U}$ with $|S_{\mathrm{rand}}^{(u)}|=K$, and compute
\begin{equation}
    R_b
    =
    \frac{1}{U}
    \sum_{u=1}^{U}
    \mathrm{CFD}_{b}
    \left(
    S_{\mathrm{rand}}^{(u)},D
    \right),
    \qquad
    Q_{\mathrm{rand}}
    =
    \frac{1}{U}
    \sum_{u=1}^{U}
    Q
    \left(
    S_{\mathrm{rand}}^{(u)}
    \right).
\end{equation}
The normalized distribution and quality losses are then defined as
\begin{equation}
    \mathcal{L}_{\mathrm{cfd}}
    =
    \frac{1}{B}
    \sum_{b=1}^{B}
    \left(
    \frac{
    \mathrm{CFD}_{b}(S,D)
    }{
    R_b+\epsilon
    }
    -1
    \right),
    \qquad
    \mathcal{L}_{\mathrm{quality}}
    =
    -
    \left(
    \frac{
    Q(S)
    }{
    Q_{\mathrm{rand}}+\epsilon
    }
    -1
    \right).
\end{equation}
The quality term is negative when the selected subset has higher average quality than random subsets, so minimizing it encourages quality improvement beyond the random baseline.

\paragraph{Differentiable hard TopK optimization}

QMAC learns a selection logit $\alpha_i$ for each sample and maps it to a continuous selection score:
\begin{equation}
    s_i
    =
    \sigma\left(\frac{\alpha_i}{\tau}\right),
    \qquad
    \mathbf{s}=(s_1,\ldots,s_N),
\end{equation}
where $\sigma(\cdot)$ is the sigmoid function and $\tau$ is the temperature. 
To satisfy the fixed budget, we use a hard TopK mask in the forward pass:
\begin{equation}
    z_i
    =
    \mathbf{1}\left[
    i\in\mathrm{TopK}(\mathbf{s},K)
    \right],
    \qquad
    \sum_{i=1}^{N} z_i = K ,
\end{equation}
Since TopK is not differentiable, we use a straight through estimator~\cite{bengio2013estimating},
\begin{equation}
    \widetilde{\mathbf{z}}
    =
    \mathbf{s}
    +
    \left(
    \mathbf{z}-\mathbf{s}
    \right)_{\mathrm{sg}},
\end{equation}
where $\mathrm{sg}$ stops gradients. 
Thus, the forward pass uses the hard selection $\mathbf{z}$, while the backward pass updates $\alpha_i$ through the continuous scores $s_i$.

\paragraph{Overall objective \& Output}

The final optimization objective is
\begin{equation}
    \min_{\boldsymbol{\alpha}}
    \mathcal{L}
    =
    \lambda_{\mathrm{cfd}}\mathcal{L}_{\mathrm{cfd}}
    +
    \lambda_{\mathrm{quality}}\mathcal{L}_{\mathrm{quality}}
    +
    \lambda_{\mathrm{reg}}
    \frac{1}{N}
    \sum_{i=1}^{N}
    \alpha_i^2 .
\end{equation}
After optimization, QMAC returns the $K$ samples with the highest scores:
\begin{equation}
    S^{\star}
    =
    \left\{
    i
    \mid
    i\in\mathrm{TopK}(\mathbf{s},K)
    \right\}.
\end{equation}

\section{Experiments}
\label{sec:experiments}

\subsection{Experimental Setup}
\label{sec:experimental_setup}

\paragraph{Datasets}
We evaluate QMAC on four datasets: Vision-Flan~\cite{xu2024vision} and OpenHermes (from https://huggingface.co/datasets/teknium/openhermes) for standard image-text and text instruction data, and LLaVA-CoT~\cite{xu2025llava} and SmallThoughts~\cite{wu2025concise} for image-text and text reasoning data. 
Reasoning data is identified by the presence of an explicit \verb|<think></think>| segment in the response. Detailed dataset descriptions and preprocessing procedures are provided in Appendix~\ref{app:datasets}.

\paragraph{Target models}
For each dataset, we instruction tune different target models to evaluate QMAC. 
On Vision-Flan, we use LLaVA-V1.5-7B~\cite{liu2024improved} and LLaVA-V1.5-13B~\cite{liu2024improved}. 
On OpenHermes, we use Qwen2.5-3B~\cite{qwen2025qwen25technicalreport} and Llama-3.1-8B~\cite{grattafiori2024llama}. 
On LLaVA-CoT, we use Qwen2-VL-2B-Instruct~\cite{wang2024qwen2} and Llama-3.2-11B-Vision-Instruct~\cite{grattafiori2024llama}. 
On SmallThoughts, we use Qwen2.5-Math-1.5B~\cite{yang2024qwen2} and Qwen2.5-Math-7B~\cite{yang2024qwen2}. 
Detailed training hyperparameters are provided in Appendix~\ref{app:training_hyperparameters}.

\paragraph{Subset budgets}
For image-text datasets, we evaluate subset budgets of $5\%$, $10\%$, and $15\%$. 
For text datasets, we evaluate subset budgets of $10\%$, $20\%$, and $30\%$.

\paragraph{Baselines}
For Vision-Flan, the standard image-text instruction dataset, we compare QMAC with Full Data, Random Selection, XMAS~\cite{naharas2025data}, COINCIDE~\cite{lee2024concept}, SemDeDup~\cite{abbas2023semdedup}, D2 Pruning~\cite{maharana2023d2}, PRISM~\cite{bi2025prism}, CVS~\cite{sun2026does}, ScalSelect~\cite{wu2026scalselect}, CLIP Score~\cite{hessel2021clipscore}, and EL2N~\cite{paul2021deep}. 
These baselines cover diversity based methods, importance based methods, and methods that combine importance and diversity. For the other datasets, we use Full Data and Random Selection as baselines.

\paragraph{Quality and Embedding models}
The quality scores in QMAC are used only as auxiliary signals to filter out clear outliers or low quality samples. 
To reduce computational overhead, we use lightweight models with vLLM~\cite{kwon2023efficient} acceleration for batch scoring. For Vision-Flan, we use q-sit-mini-0.8B~\cite{zhang2025teaching} and Qwen3-VL-4B-Instruct~\cite{bai2025qwen3} to estimate $q^{\mathrm{img}}$, $q^{\mathrm{txt}}$, and $q^{\mathrm{mm}}$, respectively. 
For OpenHermes, we use Qwen3-4B-Instruct-2507~\cite{yang2025qwen3} to estimate $q^{\mathrm{txt}}$. 
For LLaVA-CoT, we use q-sit-mini-0.8B and Qwen3.5-9B~\cite{team2026qwen3} to estimate $q^{\mathrm{img}}$ and $q^{\mathrm{mm}}$. 
For SmallThoughts, we use Qwen3.5-9B to estimate $q^{\mathrm{txt}}$. 
For embedding extraction, we also use lightweight models. 
We use Qwen3-VL-Embedding-2B~\cite{li2026qwen3} for image-text datasets and Qwen3-Embedding-4B~\cite{zhang2025qwen3} for text datasets. The scoring prompts used for quality estimation are provided in Appendix~\ref{app:prompts}.

\paragraph{Evaluation}
Following the design purpose of each dataset, we evaluate the trained models on different benchmarks. Vision-Flan, OpenHermes, LLaVA-CoT, and SmallThoughts involve 12, 12, 12, and 18 benchmarks, respectively, covering 44 benchmarks in total. The specific benchmarks used for each dataset and the corresponding detailed results are provided in Appendix~\ref{app:Evaluation}. To normalize evaluation scales across datasets, we report Average Relative Performance (ARP):
\begin{equation}
    \mathrm{ARP}
    =
    \frac{
    \text{Subset Data Performance}
    }{
    \text{Full Data Performance}
    }
    \times 100 .
\end{equation}

\begin{figure*}[t]
    \centering

    \begin{subfigure}{0.92\textwidth}
        \centering
        \includegraphics[width=\textwidth]{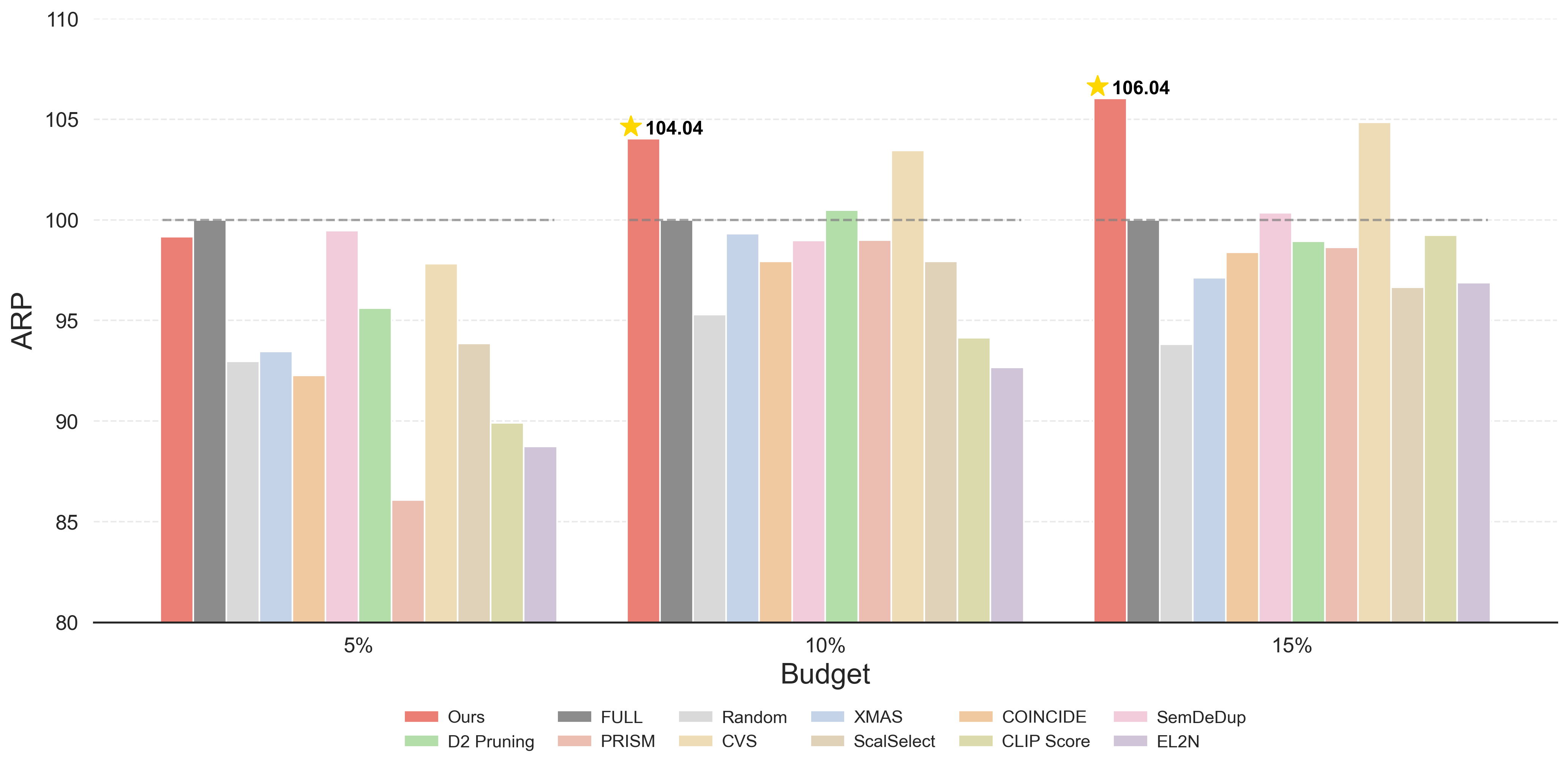}
        \caption{Vision-Flan}
        \label{fig:main_visionflan}
    \end{subfigure}

    \vspace{0.8em}

    \begin{subfigure}{0.32\textwidth}
        \centering
        \includegraphics[width=\textwidth]{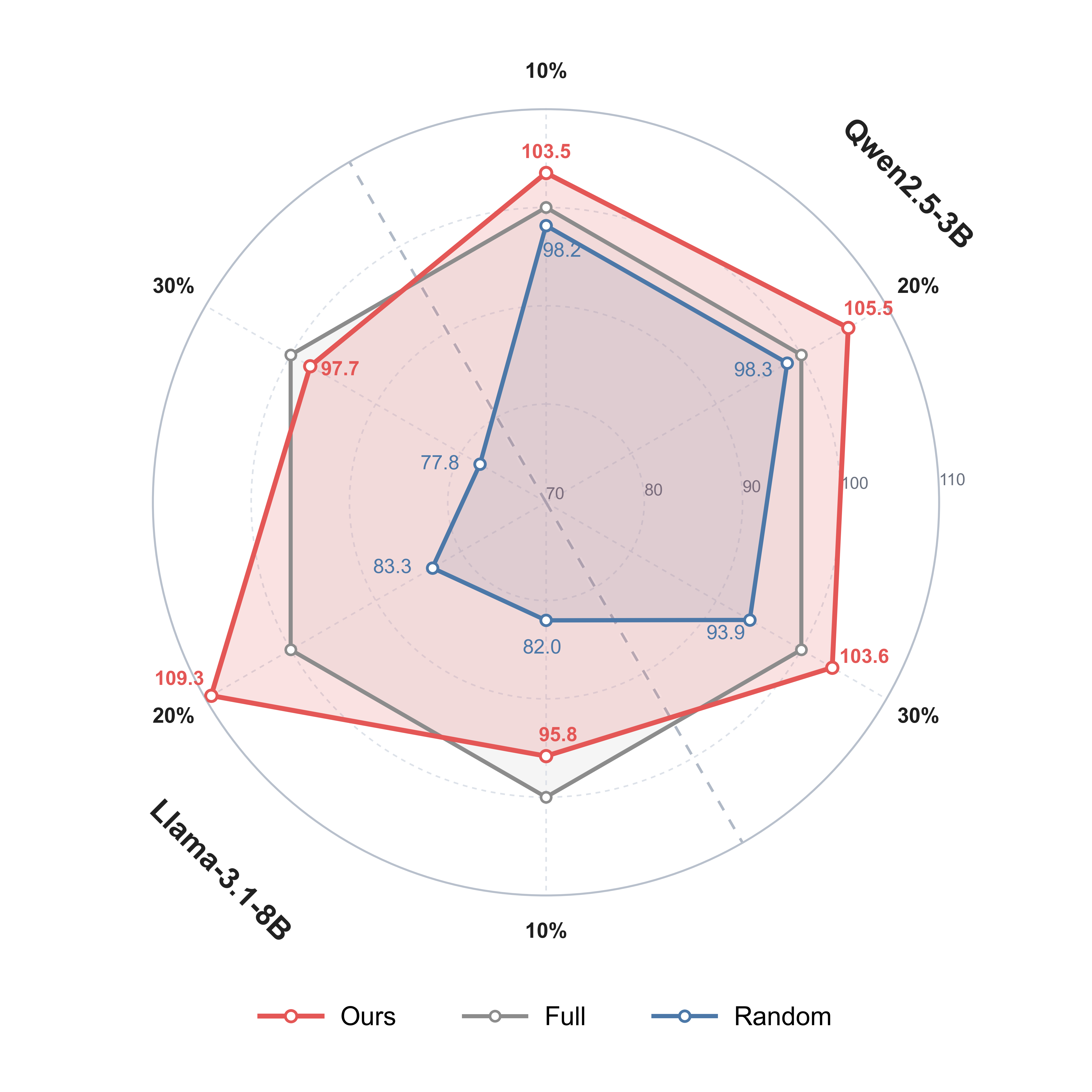}
        \caption{OpenHermes}
        \label{fig:main_openhermes}
    \end{subfigure}
    \hfill
    \begin{subfigure}{0.32\textwidth}
        \centering
        \includegraphics[width=\textwidth]{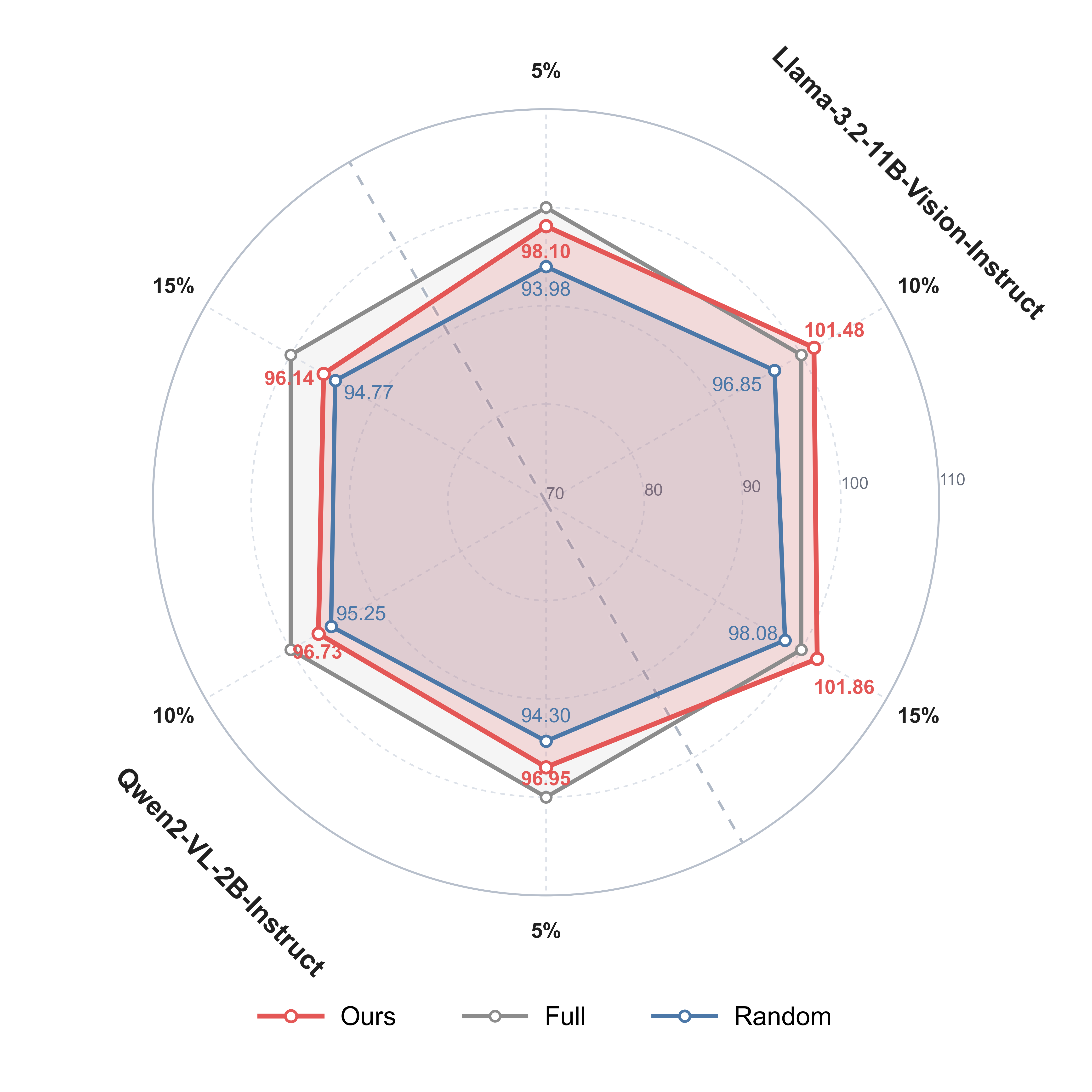}
        \caption{LLaVA-CoT}
        \label{fig:main_llava_cot}
    \end{subfigure}
    \hfill
    \begin{subfigure}{0.32\textwidth}
        \centering
        \includegraphics[width=\textwidth]{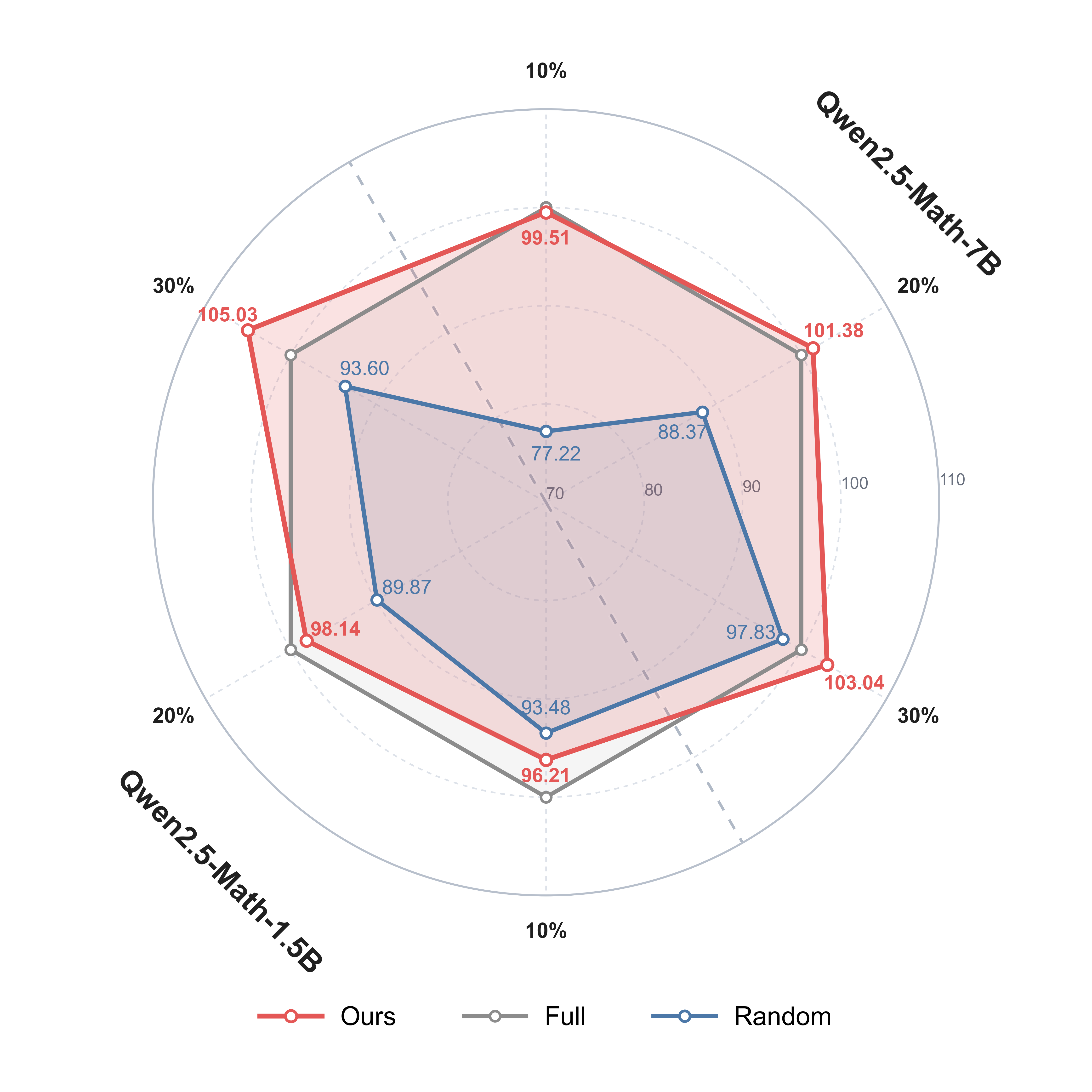}
        \caption{SmallThoughts}
        \label{fig:main_smallthoughts}
    \end{subfigure}

    \caption{
    Performance comparison of QMAC and baseline methods across four datasets.
    On Vision-Flan, QMAC achieves the best performance among all data selection methods at both the 10\% and 15\% budgets and surpasses full-data training at both budgets.
    On OpenHermes, LLaVA-CoT, and SmallThoughts, QMAC consistently outperforms Random Selection by a large margin and surpasses full-data training in half of the evaluated model and budget combinations.
    }
    \label{fig:main_results}
\end{figure*}

\subsection{Main Results}
\label{sec:main_results}

Figure~\ref{fig:main_results} shows the results of QMAC and the baseline methods on four datasets. 
On the standard image-text instruction dataset Vision-Flan, as shown in Figure~\ref{fig:main_visionflan}, QMAC already approaches Full Data performance with only a $5\%$ budget when LLaVA-V1.5-7B is used as the target model. 
When the budget increases to $10\%$ and $15\%$, QMAC reaches ARP of 104.04 and 106.04, respectively, outperforming all data selection baselines and even surpassing Full Data. 
This suggests that jointly modeling distribution matching and task related quality allows QMAC to preserve data coverage while strengthening the effective training signal in the selected subset. 
We further validate QMAC on LLaVA-V1.5-13B, with results provided in Appendix~\ref{sec:detailed_results}.

The results in Figures~\ref{fig:main_results}(\subref{fig:main_openhermes})--(\subref{fig:main_smallthoughts}) further examine the generalization of QMAC to text and reasoning data. Previous studies show that Random Selection is a very strong baseline, and many data selection methods fail to outperform random sampling in practical training~\cite{xia2024rethinking}.
On OpenHermes, LLaVA-CoT, and SmallThoughts, we focus on comparing QMAC with Random Selection and Full Data. 
Across the three datasets, six target models, and three budgets, QMAC consistently outperforms Random Selection and surpasses Full Data in half of the model and budget combinations. 
These results demonstrate that QMAC can reliably identify efficient training subsets. 
On the more challenging mathematical reasoning dataset SmallThoughts, QMAC achieves ARP of 105.03 and 103.04 at the $30\%$ budget on the two target models, respectively. 
This demonstrates that QMAC effectively retains training signals that are more valuable for improving reasoning ability.

It is worth noting that there is currently no unified data selection method that works well across all four data types considered in our experiments. 
Most existing methods are tailored to a single data type or a limited range of data types~\cite{bi2025prism,wu2026scalselect,xia2024less,yan2025coido}.
In contrast, QMAC shows stable advantages across image-text, text, standard instruction, and reasoning settings, suggesting that it can serve as a more general framework for data selection.

\section{Analysis and Ablation Studies}
\label{sec:analysis_ablation}

\subsection{QMAC Yields Better Distributional Quality}
\label{sec:distributional_quality}

\begin{figure}[t]
  \centering
  \begin{subfigure}[t]{0.24\linewidth}
    \centering
    \includegraphics[width=\linewidth]{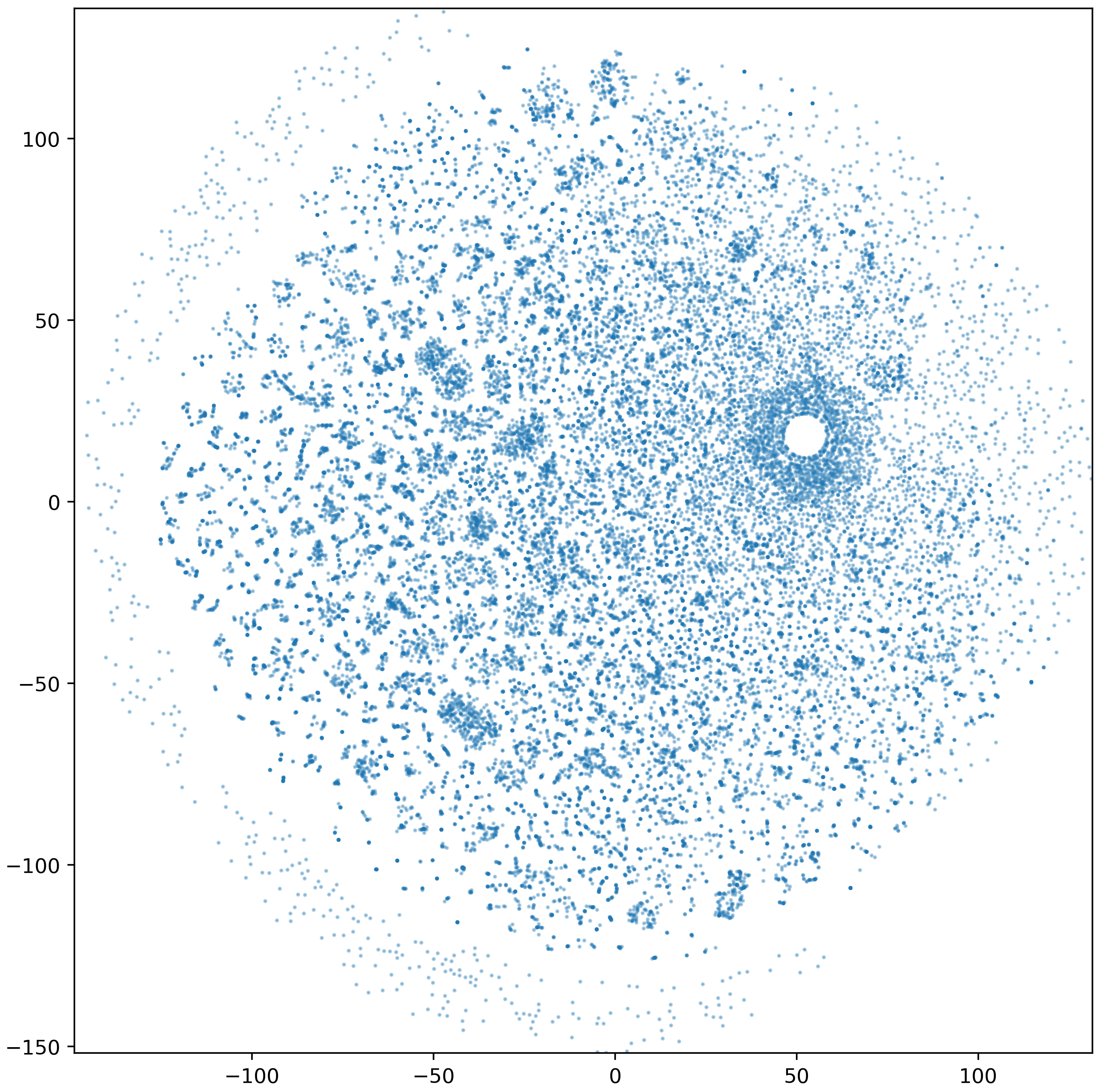}
    \caption{Random}
    \label{fig:tsne_random}
  \end{subfigure}
  \hfill
  \begin{subfigure}[t]{0.24\linewidth}
    \centering
    \includegraphics[width=\linewidth]{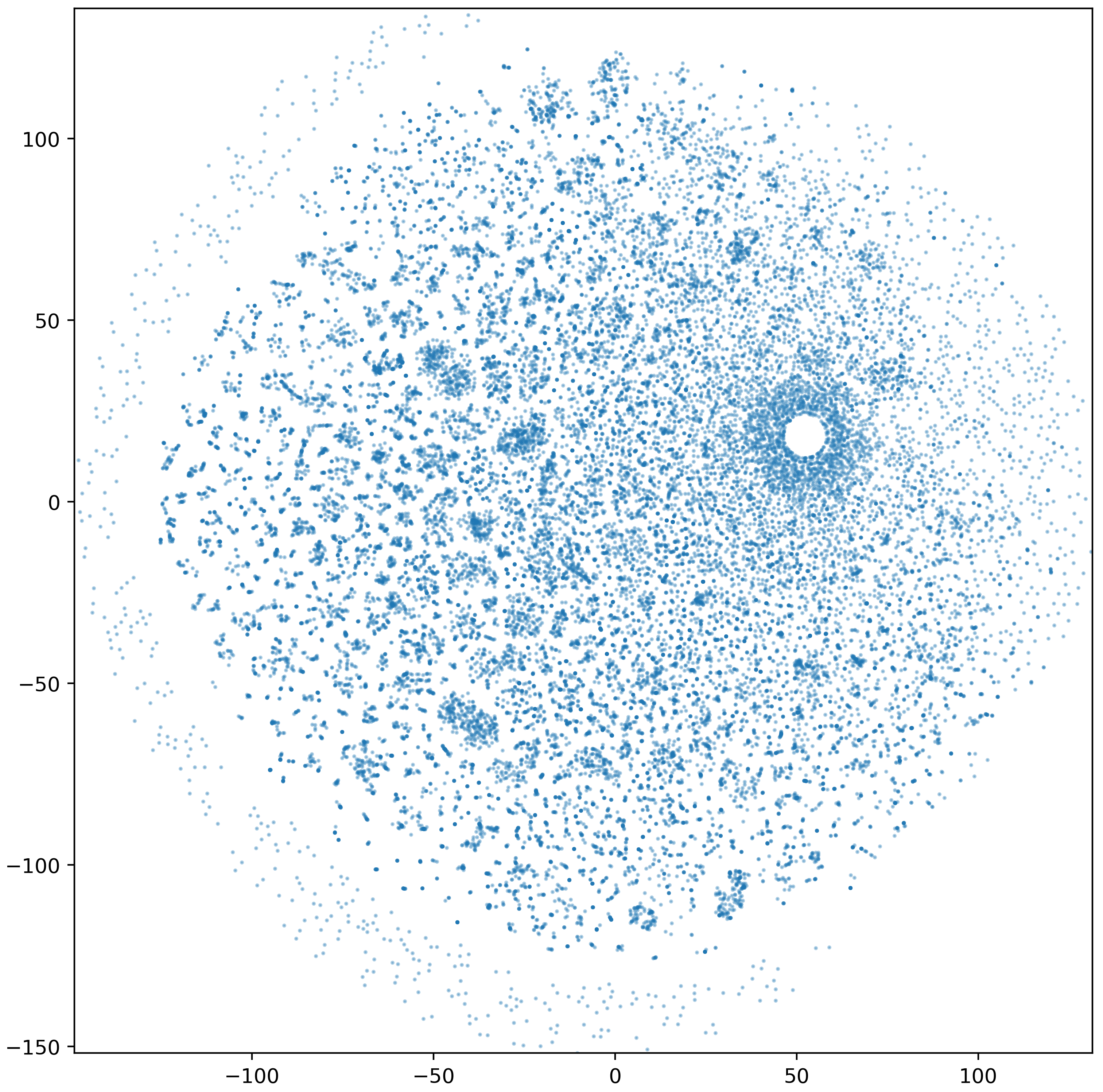}
    \caption{Diversity}
    \label{fig:tsne_diversity}
  \end{subfigure}
  \hfill
  \begin{subfigure}[t]{0.24\linewidth}
    \centering
    \includegraphics[width=\linewidth]{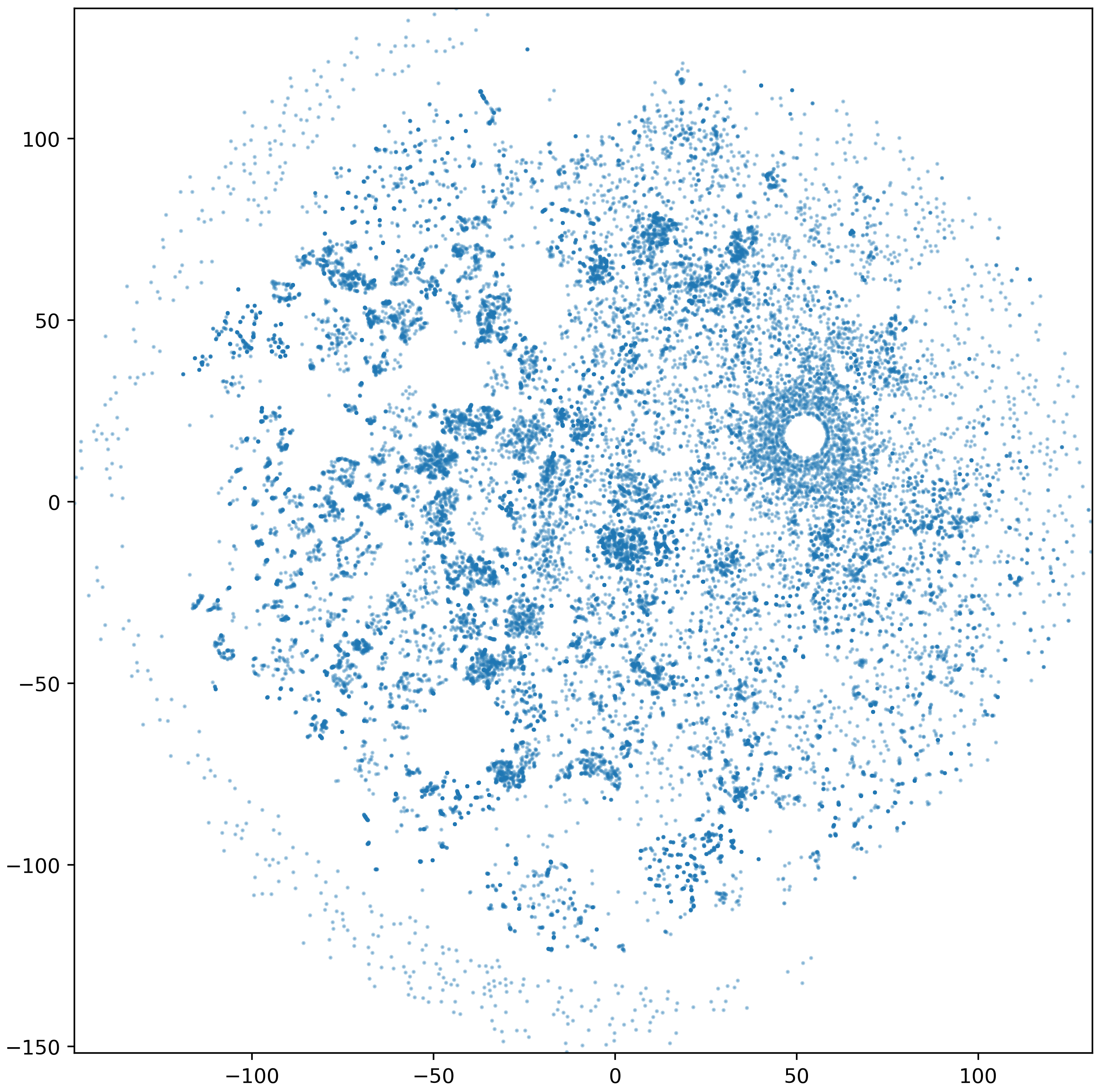}
    \caption{Importance}
    \label{fig:tsne_importance}
  \end{subfigure}
  \hfill
  \begin{subfigure}[t]{0.24\linewidth}
    \centering
    \includegraphics[width=\linewidth]{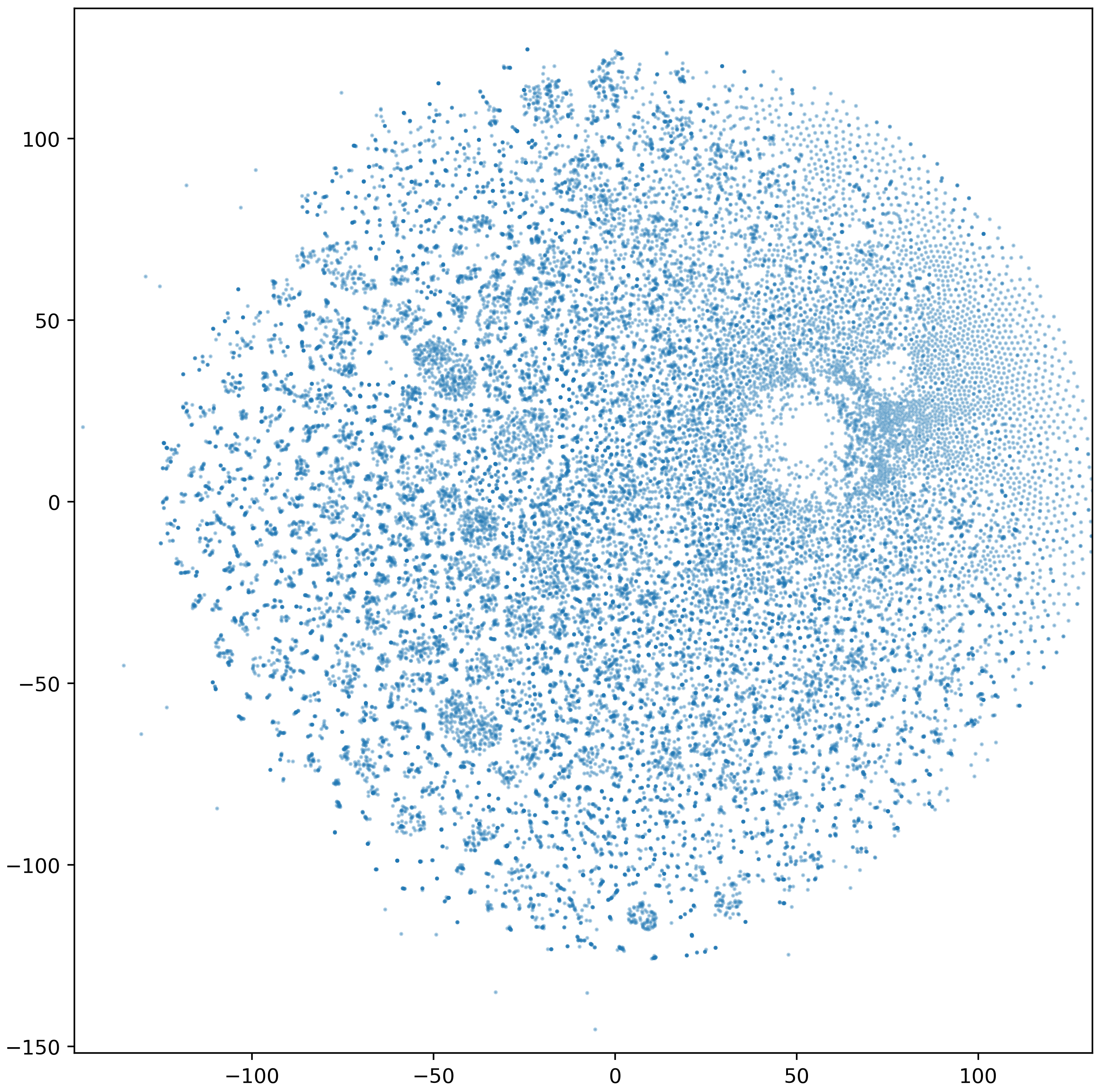}
    \caption{QMAC}
    \label{fig:tsne_qmac}
  \end{subfigure}

  \caption{t-SNE visualization of Vision-Flan subsets at a $15\%$ budget. QMAC yields a more uniform and isotropic distribution.}
  \label{fig:tsne_distribution}
\end{figure}
To qualitatively assess how different methods affect the data distribution, we apply t-SNE~\cite{van2008visualizing} to the embeddings of Vision-Flan subsets selected at a $15\%$ budget. 
Figures~\ref{fig:tsne_distribution}(\subref{fig:tsne_random})--(\subref{fig:tsne_qmac}) correspond to Random Selection, diversity based method COINCIDE, importance based method ScalSelect, and QMAC, respectively. 
We examine distributional quality from three aspects: uniformity, isotropy, and structural naturalness. 
A desirable subset should preserve sample diversity while reducing abnormal patterns such as local density imbalance, directional bias, and peripheral ring structures.

Random Selection, shown in Figure~\ref{fig:tsne_random}, approximately reflects the distribution of the full dataset. 
Although it provides broad global coverage, it also exhibits a clear sparse ring structure near the boundary and uneven local density, indicating that the full dataset itself contains structural bias. 
Diversity based methods, represented by COINCIDE in Figure~\ref{fig:tsne_diversity}, preserve the coverage of full dataset and demonstrates the benefit of maintaining sample diversity. 
However, it still inherits the peripheral ring structure and density discontinuities, suggesting that diversity alone is insufficient to correct structural defects in the original distribution. 
Importance based methods, represented by ScalSelect in Figure~\ref{fig:tsne_importance}, show a more severe degradation. 
Its selected subset is skewed toward one side, becomes globally asymmetric, and exhibits a clear density gradient, with some regions overly concentrated and others sparsely populated. 
The peripheral ring structure also remains and becomes partially fragmented. 
This indicates that importance based sampling can overemphasize high scoring or salient samples, thereby amplifying selection bias and anisotropy.

In contrast, QMAC produces a more uniform and symmetric distribution, as shown in Figure~\ref{fig:tsne_qmac}. 
The peripheral ring structure is substantially weakened, and local density changes become smoother. 
These results show that QMAC improves the uniformity and isotropy of the embedding space while preserving sample coverage and representativeness, leading to a more stable data distribution.

\subsection{Multiband CFD Balances Global Coverage and Local Robustness}
\label{sec:multiband_cfd_analysis}

\begin{figure}[!t]
  \centering
  \begin{subfigure}[t]{0.48\linewidth}
    \centering
    \includegraphics[width=\linewidth]{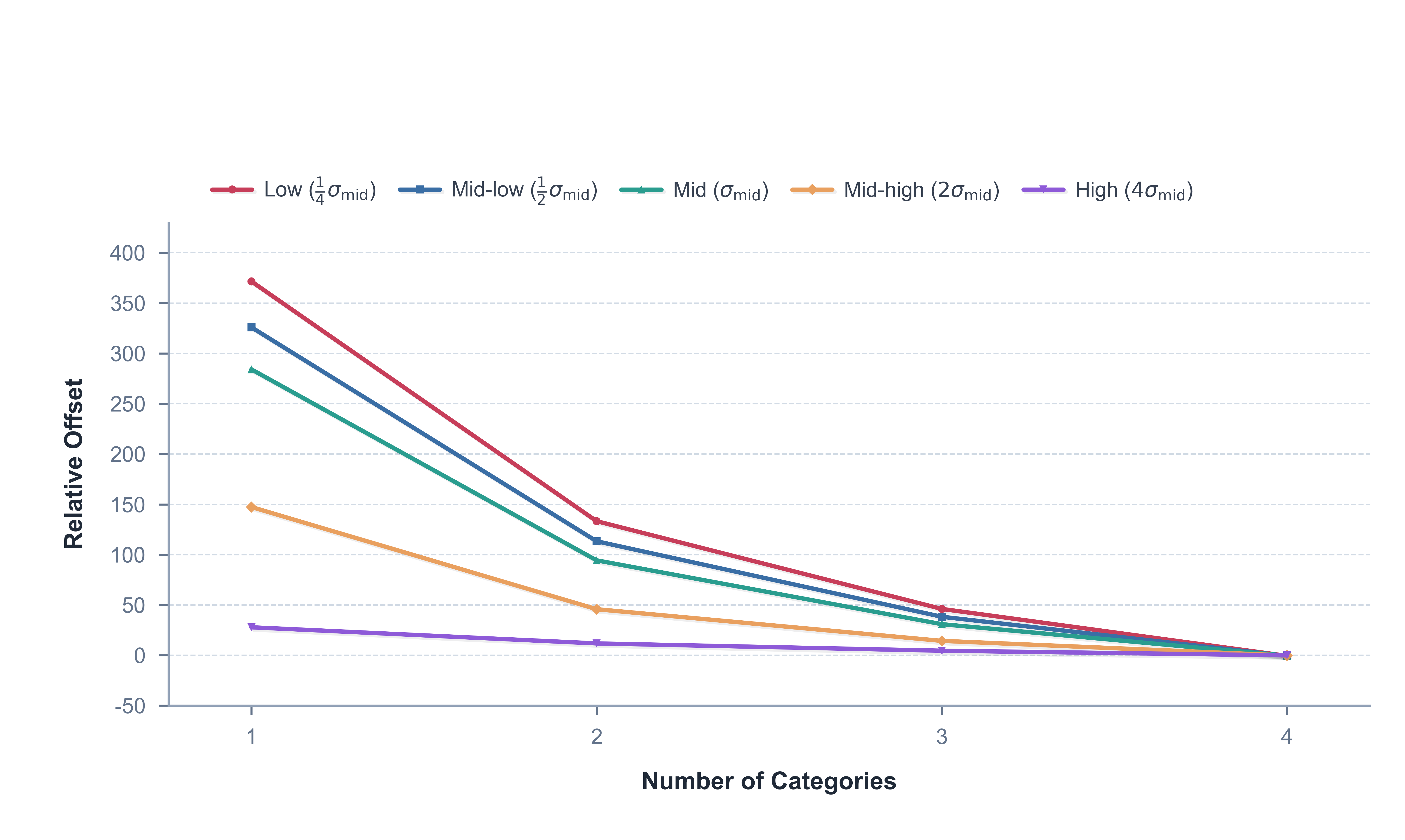}
    \caption{Band sensitivity to task coverage}
    \label{fig:task_coverage_sensitivity}
  \end{subfigure}
  \hfill
  \begin{subfigure}[t]{0.48\linewidth}
    \centering
    \includegraphics[width=\linewidth]{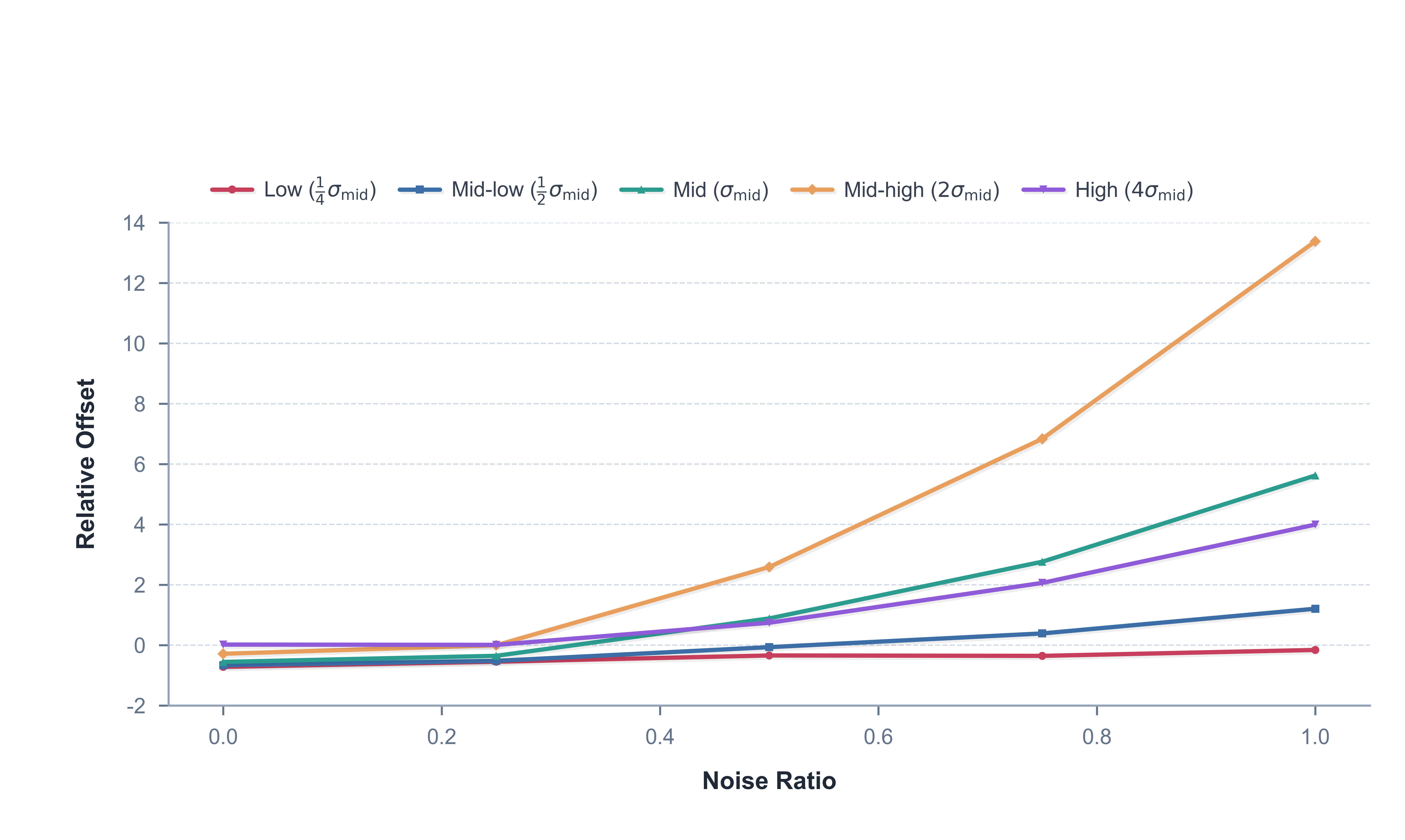}
    \caption{Band sensitivity to noise level}
    \label{fig:noise_sensitivity}
  \end{subfigure}

  \caption{Frequency band sensitivity of QMAC. Low frequency bands are more sensitive to task coverage, whereas high frequency bands are more sensitive to noise.}
  \label{fig:task_noise_ablation}
\end{figure}

We construct a dataset with 10,000 samples, consisting of 9,000 clean samples and 1,000 perturbed samples, with samples uniformly distributed across four task types.
The construction details are provided in Appendix~\ref{app:synthetic_construction}. 
Under a $10\%$ budget, we analyze the five frequency bands used in QMAC. 
The results are shown in Figure~\ref{fig:task_noise_ablation}, where a higher Relative Offset indicates that the selected subset performs worse relative to random selection.

Figure~\ref{fig:task_coverage_sensitivity} shows the effect of task type coverage, with the noise level fixed at $10\%$. 
Low frequency bands are clearly more sensitive to task coverage, indicating that they capture the global structure of the data distribution. 
This helps preserve global consistency and distributional alignment.

Figure~\ref{fig:noise_sensitivity} shows the effect of noise level, where all task types are fully covered. 
High frequency bands are much more sensitive to noise than low frequency bands, while low frequency bands are almost unaffected. 
This suggests that high frequency bands mainly capture local details and perturbations, which helps identify noisy samples and preserve local stability.

Taken together, Figure~\ref{fig:task_coverage_sensitivity} and Figure~\ref{fig:noise_sensitivity} show that the middle frequency bands lie between these two behaviors. 
They retain semantic information while also capturing local perturbations, thereby balancing global consistency and local robustness.

\begin{figure}[!t]
  \centering

  \begin{minipage}[t]{0.31\linewidth}
    \vspace{0pt}
    \centering
    \includegraphics[width=\linewidth]{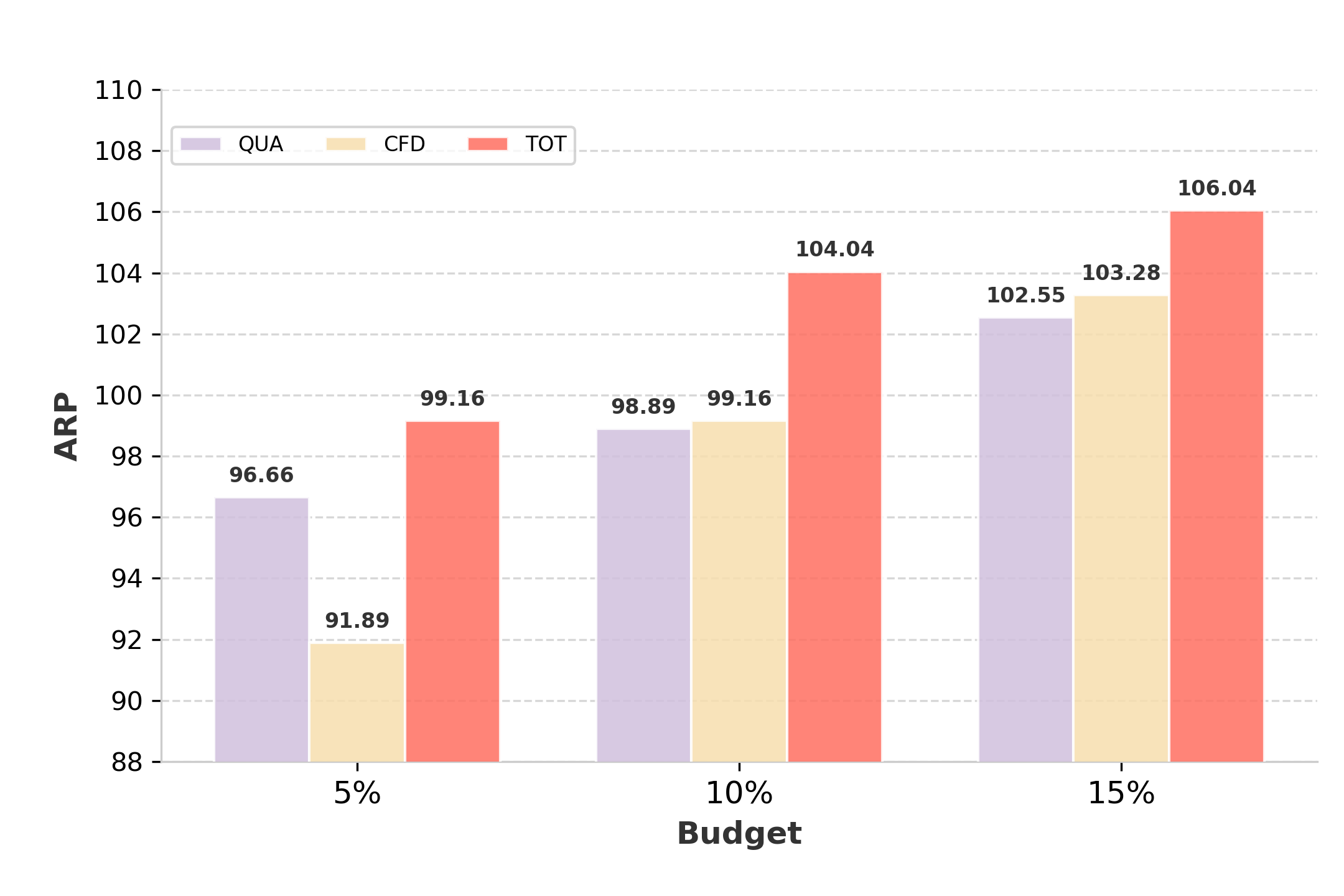}
    \caption{Ablation on quality and distribution constraints. Combining QUA and CFD yields the best performance.}
    \label{fig:constraint_ablation}
  \end{minipage}
  \hfill
  \begin{minipage}[t]{0.31\linewidth}
    \vspace{0pt}
    \centering
    \includegraphics[width=\linewidth]{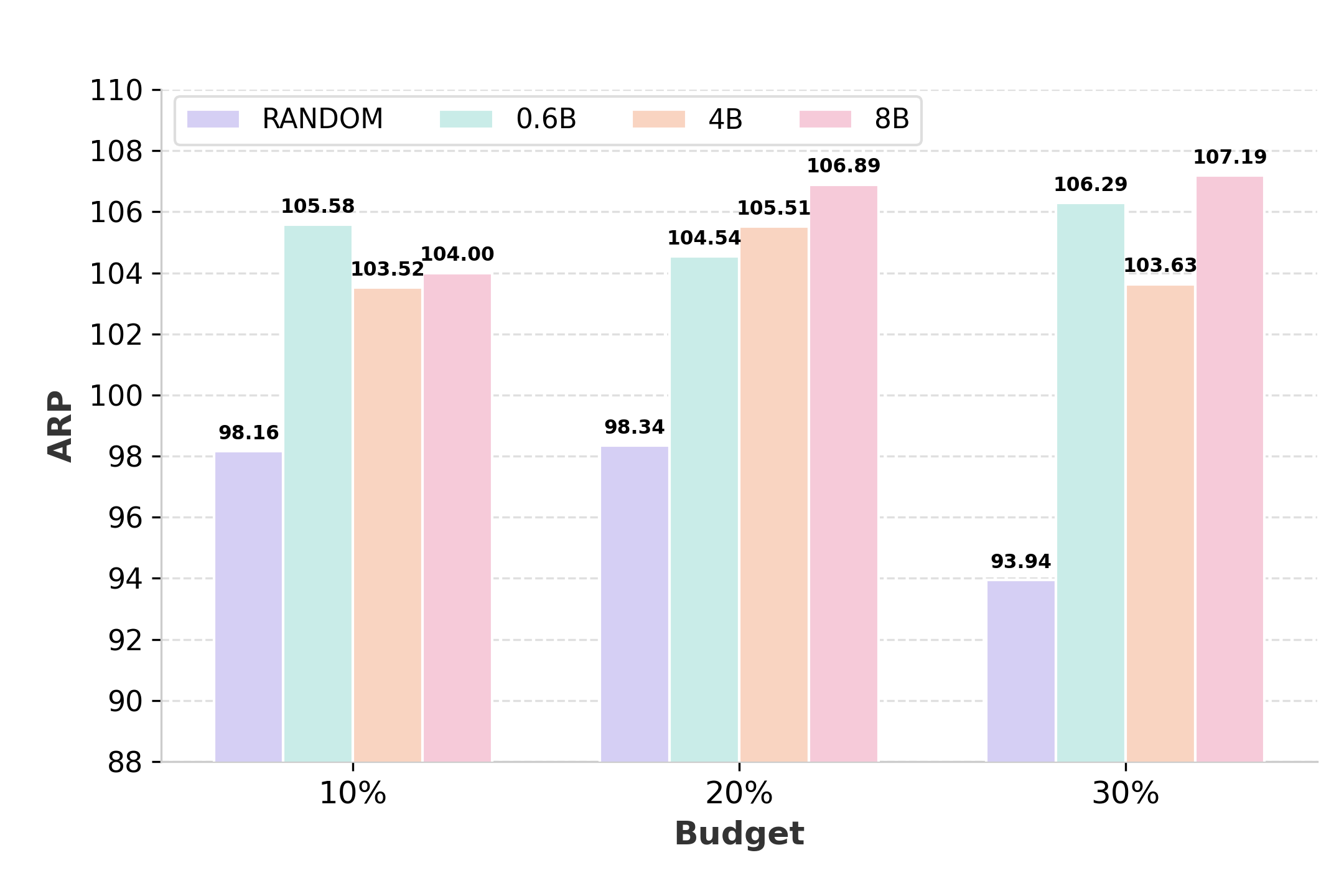}
    \caption{Ablation on embedding models. QMAC remains robust across embedding model scales.}
    \label{fig:embedding_ablation}
  \end{minipage}
  \hfill
  \begin{minipage}[t]{0.31\linewidth}
    \vspace{0pt}
    \centering
    \includegraphics[width=\linewidth]{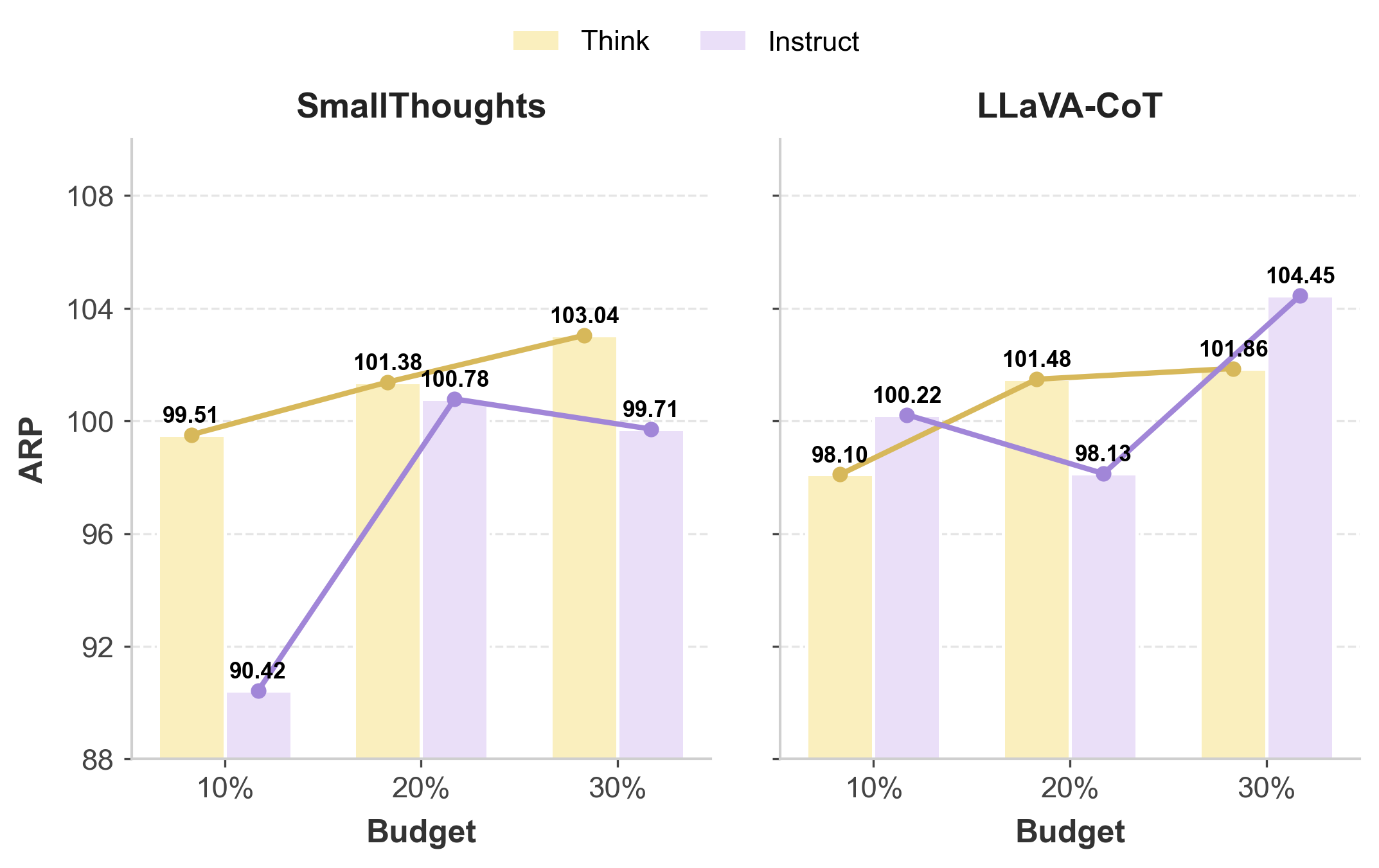}
    \caption{Ablation on quality models. Quality scoring acts as an auxiliary filtering signal.}
    \label{fig:quality_ablation}
  \end{minipage}

\end{figure}
% , and Think provides more stable gains across increasing budgets

\subsection{Ablation on Quality and Distribution Constraints}
\label{sec:quality_distribution_ablation}

We ablate the quality constraint and the distribution constraint in QMAC on Vision-Flan, using LLaVA-V1.5-7B as the target model. 
We compare three variants: quality only (QUA), CFD distribution constraint only (CFD), and their combination (TOT). 
As shown in Figure~\ref{fig:constraint_ablation}, QUA outperforms CFD under the low budget of $5\%$, indicating that when only a very small number of samples can be selected, the effective training signal from high quality samples is more critical. 
As the budget increases, CFD surpasses QUA, suggesting that distribution coverage and diversity become more important for generalization when more samples are available.

TOT achieves the best performance under all budgets. 
This shows that the quality constraint and the distribution constraint are complementary: the former improves sample effectiveness, while the latter preserves distributional representativeness. 
By combining both constraints, QMAC can more stably balance quality and coverage across different subset budgets, leading to better performance.

\subsection{Ablation on Embedding Models}
\label{sec:quality_embedding_ablation}
We ablate the embedding model used by QMAC on OpenHermes, using Qwen2.5-3B as the target model. 
We only vary the parameter scale of Qwen3-Embedding and compare three settings: 0.6B, 4B, and 8B. 
As shown in Figure~\ref{fig:embedding_ablation}, all embedding models substantially outperform Random Selection under all budgets, indicating that QMAC is robust to the scale of the embedding model. 
Overall, larger embedding models tend to yield better performance, and the 8B model achieves the best results at the $20\%$ and $30\%$ budgets. 
However, the 0.6B model also performs strongly across multiple budgets, suggesting that QMAC does not rely on large embedding models and that lightweight models can already provide effective distributional representations.

\subsection{Ablation on Quality Models}
\label{sec:quality_quality_ablation}
We ablate the quality model used by QMAC on LLaVA-CoT and SmallThoughts, using Llama-3.2-11B-Vision-Instruct and Qwen2.5-Math-7B as the target models, respectively. 
With Qwen3.5-9B as the scoring model, we compare two scoring configurations: Think enables reasoning mode and first generates an analysis with multidimensional judgments before producing the quality score, while Instruct disables reasoning mode and directly predicts the score. 
Since Think benefits from both reasoning mode and explicit scoring rationales, its reasoning and evaluation stability are substantially stronger than those of Instruct. 
As shown in Figure~\ref{fig:quality_ablation}, Instruct performs better overall on LLaVA-CoT, while Think performs better overall on SmallThoughts, indicating that quality scores mainly act as auxiliary signals for filtering clearly erroneous or outlier samples. 
Meanwhile, Think improves steadily with larger budgets on both datasets, whereas Instruct shows more fluctuation, suggesting that stronger quality models can provide more stable gains as the subset budget increases.

\section{Conclusion and Limitations}
\label{sec:conclusion_limitations}

\paragraph{Conclusion}
We propose QMAC, a quality aware multiband characteristic function matching method for instruction tuning data selection. 
Unlike existing methods that rely mainly on diversity, importance, or their combination, QMAC explicitly constrains the distributional consistency between the selected subset and the full data in the embedding space, while using task related quality signals to filter low quality samples. 
Experiments show that QMAC achieves stable gains across standard image-text data, standard text data, image-text reasoning data, and text reasoning data. 
It can substantially reduce the amount of training data while approaching or even surpassing the performance of full data training.

\paragraph{Limitations}
First, although QMAC shows robustness across different embedding models and quality models, its selection performance still depends on the reliability of the representation space and scoring signals. 
When the embeddings or quality scores contain systematic bias, the selected subset may also be affected. 
Second, QMAC requires additional embedding extraction, quality scoring, and optimization steps. 
Although these costs are usually much lower than full data training, there remains room to further improve efficiency on larger scale datasets. 
Finally, our experiments focus on instruction tuning. 
The applicability of QMAC to pretraining, reinforcement learning, and other data selection scenarios remains to be further validated.

\newpage
\bibliographystyle{plainnat}
\bibliography{references}

\newpage
%%%%%%%%%%%%%%%%%%%%%%%%%%%%%%%%%%%%%%%%%%%%%%%%%%%%%%%%%%%%

\appendix

\section{Related Work}
\label{sec:related_work}

Data selection for large language models (LLMs) and large vision language models (LVLMs) aims to select a compact yet effective training subset from large training data under a limited budget. 
The objective is to match or even surpass the performance obtained by training on the full training data.
Existing methods can be broadly grouped into three categories: importance based methods, diversity based methods, and hybrid methods that combine both criteria.

\subsection{Importance Based Methods}
\label{sec:rw_importance}

Importance based methods aim to estimate the training value of individual samples and preferentially retain samples that are more beneficial to model learning. 
A common strategy is to use validation signals or training dynamics as proxies for sample utility. 
For example, LESS~\cite{xia2024less}, TIVE~\cite{liu2025less}, ICONS~\cite{wu2024icons}, and OPUS~\cite{wang2026opus} select data according to gradient similarity between training samples and validation samples. 
Another line of work trains a proxy model on a small amount of data, and then identifies informative samples based on their loss or prediction difficulty under the proxy model, including MLLM-Selector~\cite{ma2025mllm}, Token Cleaning~\cite{pang2025token}, and EL2N~\cite{paul2021deep}.

Beyond proxy or validation based signals, some methods estimate importance from the behavior of the target model itself. ScalSelect~~\cite{wu2026scalselect} evaluates sample importance by analyzing the target model's internal attention behavior, particularly its attention distributions. CVS~\cite{sun2026does} and Exploring Learning Complexity for Efficient Downstream Dataset Pruning~\cite{jiang2024exploring} use masking operations and select samples according to the induced changes in output probabilities and losses during model generation. 
DONOD~\cite{hu2025donod} further measures sample value through its influence on model parameter changes after training. 
In addition, several methods introduce explicit scoring models. 
Better Reasoning with Less Data~\cite{xu2025better}, Trust the Model~\cite{toibazar2025trust}, and Learning from the Best~\cite{he2025learning} train quality or utility scorers to assign a score to each sample.

Overall, importance based methods can effectively identify high value samples, but they often emphasize local sample utility and may overlook the global distributional structure of the selected subset.

\subsection{Diversity Based Methods}
\label{sec:rw_diversity}

Diversity based methods emphasize coverage and distributional representativeness in the semantic space, aiming to preserve the overall structure of the original data under a small data budget. 
PRISM~\cite{bi2025prism}, FisherSFT~\cite{deb2025fishersft}, SemDeDup~\cite{abbas2023semdedup}, Self-Sup~\cite{sorscher2022beyond}, and D2 Pruning~\cite{maharana2023d2} operate on sample embeddings through deduplication, clustering, or coverage modeling to improve the diversity and representativeness of the selected subset. 
ICONS~\cite{wu2024icons} and INSTAG~\cite{lu2023instag} construct more representative training subsets by maximizing the coverage of downstream task types.

Compared with importance based methods, diversity based methods better preserve global coverage, but they may retain noisy or low quality samples when representativeness is considered alone.

\subsection{Hybrid Methods}
\label{sec:rw_hybrid}

Hybrid methods aim to balance data coverage and sample utility. 
They seek to avoid the distributional bias caused by purely importance based selection, as well as the neglect of sample quality in purely diversity based selection. 
CoIDO~\cite{yan2025coido} trains a unified scorer to jointly model data quality and diversity. 
DataTailor~\cite{yu2025mastering} designs selection criteria from multiple dimensions, including sample informativeness, intra cluster and inter cluster relations, and the number of dialogue turns. 
COINCIDE~\cite{lee2024concept} extracts training data embeddings from multiple layers of the target model, and samples from both intra cluster and inter cluster perspectives based on the clustering structure. 
Self-Filter~\cite{chen2024your} first trains a data scorer using loss signals, and then introduces diversity constraints through KNN neighborhood penalties and CLIP embedding similarity.

QMAC also falls into the family of hybrid methods, but it differs from existing work in its formulation. 
We model data selection for LLMs and LVLMs as a joint problem of direct distribution matching and quality filtering. 
In the embedding space, QMAC preserves distributional representativeness through multiband characteristic function distance, while using task related quality signals to filter low quality or noisy samples. 
This yields selected subsets that provide both broad coverage and high training value.

\section{Data Analysis}
\label{app:data_analysis}

We analyze the embedding geometry of two representative datasets: Vision-Flan and OpenHermes. For both datasets, we use the centered and $\ell_2$-normalized embeddings, which are the same representations used by QMAC.

We report several geometry statistics in Table~\ref{tab:data_manifold_analysis}. The PCA components for 90\% variance and the PCA effective rank describe the global linear spectral complexity of the embedding distribution. The TwoNN intrinsic dimension estimates the intrinsic dimension from nearest-neighbor distance ratios. We also compute local PCA dimensions by applying PCA within each sample's 30-nearest-neighbor neighborhood and counting the minimum number of local principal components required to explain 90\% of the local variance. Finally, we compare PCA and Isomap neighborhood preservation, which measures how well the nearest-neighbor structure in the original embedding space is preserved after low-dimensional projection.

\begin{table}[t]
\centering
\caption{Embedding geometry statistics for Vision-Flan and OpenHermes. Both datasets exhibit high global PCA complexity but much lower local intrinsic dimensions, suggesting globally diverse yet locally low-dimensional semantic structures.}
\label{tab:data_manifold_analysis}
\footnotesize
\resizebox{0.88\linewidth}{!}{
\begin{tabular}{lcc}
\toprule
Metric & Vision-Flan & OpenHermes \\
\midrule
Embedding dimension & 2,048 & 2,560 \\
PCA components for 90\% variance & 379 & 553 \\
PCA effective rank & 255.06 & 342.27 \\
TwoNN intrinsic dimension & 8.25 & 10.83 \\
Local PCA dimension (mean) & 17.38 & 21.88 \\
Local PCA dimension (median) & 19 & 22 \\
PCA neighborhood preservation & 0.383 & 0.256 \\
Isomap neighborhood preservation & 0.499 & 0.426 \\
\bottomrule
\end{tabular}
}
\end{table}

The results show that both datasets have high global spectral complexity. On Vision-Flan, although the embedding dimension is 2,048, 379 principal components are still required to explain 90\% of the variance, and the effective PCA rank is 255.06. On OpenHermes, the embedding dimension is 2,560, while 553 principal components are required to explain 90\% of the variance and the effective PCA rank reaches 342.27. These results indicate that the data is not concentrated in a single low-dimensional linear subspace.

At the same time, the intrinsic and local dimensions are much smaller than both the ambient embedding dimensions and the global PCA ranks. The TwoNN intrinsic dimension is 8.25 on Vision-Flan and 10.83 on OpenHermes. The local PCA dimensions are also relatively low, with mean values of 17.38 and 21.88 and median values of 19 and 22, respectively. This gap between global PCA complexity and local dimensionality indicates that the data is globally diverse but locally concentrated around much lower-dimensional structures.

The neighborhood preservation results further suggest that this structure is nonlinear. For both datasets, Isomap preserves local neighborhoods better than PCA. On Vision-Flan, the preservation score increases from 0.383 with PCA to 0.499 with Isomap. On OpenHermes, it increases from 0.256 to 0.426. Since PCA captures a global linear projection while Isomap better preserves nonlinear neighborhood geometry, this comparison suggests that the embedding distributions are better described as nonlinear semantic manifolds rather than single global linear subspaces.

Overall, the analysis shows that instruction tuning data does not behave like unstructured high-dimensional scatter points. Instead, it exhibits high global semantic diversity together with low-dimensional local geometry. This observation motivates data selection methods to preserve the distributional structure of the embedding space, rather than selecting only isolated representative points or locally high-scoring samples.
%%%%%%%%%%%%%%%%%%%%%%%%%%%%%%%%%%%%%%%%%%%%%%%%%%%%%%%%%%%%

\section{QMAC Hyperparameter Settings}
\label{app:hyperparameters}

We separate the fixed implementation settings of QMAC from the hyperparameters that are either
optimized by Optuna~\cite{akiba2019optuna} or vary across datasets and budgets. For all experiments, we use five frequency bands and sample
\(M=1024\) frequency vectors per band. The selector temperature is fixed to \(\tau=1.0\), and the
selector scores are computed with a sigmoid transformation. We fix the CFD weight to
\(\lambda_{\mathrm{cfd}}=1.0\) and use it as the reference scale for the distribution matching term.

For selector optimization, we initialize all selector logits to the same value
\begin{equation}
\alpha_i = \log \frac{r}{1-r}, \qquad r = K/N,
\end{equation}
where \(K\) is the subset budget and \(N\) is the size of the full dataset. Therefore, the initial
sigmoid score of every sample is equal to the target selection ratio. For each Optuna trial, we
optimize the selector with Adam for at most 1024 steps. We use early stopping with a patience of
128 steps and restore the best checkpoint. The final selected subset is the hard TopK subset from
the best checkpoint.

For random baseline normalization in Eq.~(13), we use \(U=128\) random subsets sampled under
the same budget. This same value is used for both the CFD baseline and the quality baseline. To set
the frequency scales, we estimate the median pairwise distance from a random 10\% subset of the
preprocessed embeddings.

We do not manually tune the relative weights of the QMAC objective for each dataset. Given the
fixed CFD weight, Optuna automatically tunes the quality weight \(\lambda_{\mathrm{quality}}\), the
logit regularization weight \(\lambda_{\mathrm{reg}}\), and the learning rate of the selector optimizer.
For each dataset and budget, we run Optuna for 128 trials. The quality composition weights
\(\beta_{\mathrm{img}}, \beta_{\mathrm{txt}}, \beta_{\mathrm{mm}}\) are fixed according to the data modality
of each dataset. The resulting settings are summarized in Table~\ref{tab:qmac_hyperparams_all}.

\begin{table*}[t]
\centering
\caption{QMAC hyperparameters and dataset-dependent quality composition weights used for subset selection. Fixed settings shared by all rows include \(M=1024\) frequencies per band, \(\tau=1.0\), \(\lambda_{\mathrm{cfd}}=1.0\), \(U=128\) random subsets for baseline normalization, and at most 1024 selector optimization steps.}
\label{tab:qmac_hyperparams_all}
\resizebox{\textwidth}{!}{
\begin{tabular}{lccccccc}
\toprule
Dataset
& Budget
& lr
& $\lambda_{\mathrm{quality}}$
& $\lambda_{\mathrm{reg}}$
& $\beta_{\mathrm{img}}$
& $\beta_{\mathrm{txt}}$
& $\beta_{\mathrm{mm}}$ \\
\midrule
Vision-Flan
& 5\%
& $4.1430 \times 10^{-3}$
& 8.0246
& $1.6038 \times 10^{-4}$
& 0.2
& 0.2
& 0.6 \\

Vision-Flan
& 10\%
& $2.0674 \times 10^{-3}$
& 12.7972
& $2.0257 \times 10^{-7}$
& 0.2
& 0.2
& 0.6 \\

Vision-Flan
& 15\%
& $1.3220 \times 10^{-3}$
& 12.3325
& $6.2026 \times 10^{-4}$
& 0.2
& 0.2
& 0.6 \\

OpenHermes
& 10\%
& $6.8755 \times 10^{-4}$
& 14.1393
& $4.1548 \times 10^{-7}$
& 0
& 1
& 0 \\

OpenHermes
& 20\%
& $3.9690 \times 10^{-3}$
& 16.5022
& $4.0876 \times 10^{-4}$
& 0
& 1
& 0 \\

OpenHermes
& 30\%
& $2.3940 \times 10^{-3}$
& 82.6297
& $9.7213 \times 10^{-4}$
& 0
& 1
& 0 \\

LLaVA-CoT
& 5\%
& $1.6904 \times 10^{-3}$
& 22.1084
& $8.9601 \times 10^{-6}$
& 0.2
& 0
& 0.8 \\

LLaVA-CoT
& 10\%
& $1.1613 \times 10^{-3}$
& 96.4032
& $1.3174 \times 10^{-5}$
& 0.2
& 0
& 0.8 \\

LLaVA-CoT
& 15\%
& $4.3996 \times 10^{-3}$
& 50.9842
& $1.7158 \times 10^{-7}$
& 0.2
& 0
& 0.8 \\

SmallThoughts
& 10\%
& $4.2200 \times 10^{-4}$
& 5.9691
& $1.9048 \times 10^{-6}$
& 0
& 1
& 0 \\

SmallThoughts
& 20\%
& $3.9684 \times 10^{-3}$
& 16.8139
& $2.7250 \times 10^{-4}$
& 0
& 1
& 0 \\

SmallThoughts
& 30\%
& $3.7824 \times 10^{-3}$
& 24.4853
& $2.3954 \times 10^{-7}$
& 0
& 1
& 0 \\
\bottomrule
\end{tabular}
}
\end{table*}

\section{Datasets and Preprocessing Details}
\label{app:datasets}

\subsection{Datasets}

\paragraph{Vision-Flan}
Vision-Flan~\cite{xu2024vision} is a high-quality human-annotated visual instruction tuning dataset designed to improve the generalization ability of vision-language models across diverse visual tasks. It is constructed from publicly available academic vision datasets and covers 187 fine-grained visual tasks, each paired with expert-written and carefully validated task instructions. The dataset reduces synthetic bias and hallucination risk while providing broad task coverage, containing approximately 1,000 samples per task and about 186K samples in total.

\paragraph{OpenHermes}
OpenHermes (from https://huggingface.co/datasets/teknium/openhermes) is a fully open-source text instruction tuning dataset composed of approximately 242K entries. It primarily consists of GPT-4-generated instruction-response data collected from a broad mixture of open datasets across the AI landscape, including GPTeacher, WizardLM, Airoboros GPT-4, Camel-AI domain expert datasets, CodeAlpaca, GPT4-LLM, and Unnatural Instructions. The dataset covers diverse instruction-following scenarios such as general instruction, role-playing, coding, and domain-specific tasks.

\paragraph{LLaVA-CoT}
LLaVA-CoT-100k~\cite{xu2025llava} is an image-text reasoning instruction dataset designed to improve the structured reasoning ability of vision-language models. It integrates approximately 99K image-question-answer pairs from a mixture of general-purpose and science-targeted VQA datasets, including ShareGPT4V, ChartQA, A-OKVQA, AI2D, GeoQA+, ScienceQA, DocVQA, PISC, CLEVR, and CLEVR-Math. Each sample is annotated with a structured reasoning response consisting of summary, caption, reasoning, and conclusion stages.

\paragraph{SmallThoughts}
SmallThoughts~\cite{wu2025concise} is a text reasoning instruction dataset designed to provide concise and difficulty-adaptive reasoning traces for supervised fine-tuning. It contains 100K question-solution pairs, where each solution is rewritten into a compact chain-of-thought style response whose length is adapted to the estimated problem difficulty. Compared with conventional long reasoning traces, SmallThoughts provides shorter yet complete reasoning supervision. It focuses on efficient mathematical and general reasoning.

\subsection{Preprocessing Procedures}

For Vision-Flan and SmallThoughts, we directly use the original datasets without additional filtering or format conversion. For LLaVA-CoT, we remove samples without associated images and retain 98,572 image-text reasoning samples. To unify the response format with the reasoning-data setting used in our experiments, we wrap the reasoning part and the final answer part of each response with \verb|<think></think>| and \verb|<answer></answer>|, respectively. For OpenHermes, since the performance gain of text instruction tuning becomes marginal after the data scale exceeds 100K samples, we randomly sample 100K examples from the original dataset and apply no further processing.

\section{Training Hyperparameters}
\label{app:training_hyperparameters}

Table~\ref{tab:training_hyperparameters} summarizes the training hyperparameters used for all target models. 
For LLaVA-V1.5-7B~\cite{liu2024improved} and LLaVA-V1.5-13B~\cite{liu2024improved}, we follow the hyperparameter settings from the official LLaVA training code. 

For Qwen2.5-3B~\cite{qwen2025qwen25technicalreport} and Llama-3.1-8B~\cite{grattafiori2024llama}, we use different learning rates according to the fine-tuning strategy. 
Specifically, LoRA fine-tuning uses a learning rate of $1\times10^{-4}$, while full-parameter fine-tuning uses a smaller learning rate of $1\times10^{-5}$ for stable optimization.

For Qwen2-VL-2B-Instruct~\cite{wang2024qwen2} and Llama-3.2-11B-Vision-Instruct~\cite{grattafiori2024llama}, since both models are already instruction-tuned checkpoints, we use a conservative learning rate of $1\times10^{-5}$ for full-parameter fine-tuning. 
We set the batch size to 64 for these two models because using a batch size of 128 leads to too few optimization steps within one epoch, both for the full dataset and for subset training, making the training insufficient.

For Qwen2.5-Math-1.5B~\cite{yang2024qwen2} and Qwen2.5-Math-7B~\cite{yang2024qwen2}, we follow the hyperparameter choices from the original setting, but train for two epochs instead of three. 
We found that training for three epochs can lead to overfitting. 
Moreover, the original training protocol selects the best checkpoint during training rather than directly using the final checkpoint after three epochs, so we adopt a shorter training schedule for a fair and stable evaluation.
\begin{table*}[t]
\centering
\small
\resizebox{\textwidth}{!}{
\begin{tabular}{lccccccccc}
\toprule
\textbf{Model} & \textbf{Strategy} & \textbf{Epoch} & \textbf{Batch size} & \textbf{LR} & \textbf{Optimizer} & \textbf{LR scheduler} & \textbf{Warmup ratio} & \textbf{LoRA rank} & \textbf{LoRA alpha} \\
\midrule
LLaVA-V1.5-7B & LoRA & 1 & 128 & $2\times10^{-4}$ & AdamW & Cosine & 0.03 & 128 & 256 \\
LLaVA-V1.5-13B & LoRA & 1 & 128 & $2\times10^{-4}$ & AdamW & Cosine & 0.03 & 128 & 256 \\
Qwen2.5-3B & LoRA & 1 & 128 & $1\times10^{-4}$ & AdamW & Cosine & 0.03 & 32 & 64 \\
Llama-3.1-8B & Full & 1 & 128 & $1\times10^{-5}$ & AdamW & Cosine & 0.03 & -- & -- \\
Qwen2-VL-2B-Instruct & Full & 1 & 64 & $1\times10^{-5}$ & AdamW & Cosine & 0.03 & -- & -- \\
Llama-3.2-11B-Vision-Instruct & Full & 1 & 64 & $1\times10^{-5}$ & AdamW & Cosine & 0.03 & -- & -- \\
Qwen2.5-Math-1.5B & Full & 2 & 128 & $5\times10^{-5}$ & AdamW & Cosine & 0.03 & -- & -- \\
Qwen2.5-Math-7B & Full & 2 & 128 & $5\times10^{-5}$ & AdamW & Cosine & 0.03 & -- & -- \\
\bottomrule
\end{tabular}
}
\caption{Training hyperparameters for target models.}
\label{tab:training_hyperparameters}
\end{table*}

\section{Quality Scoring Prompts}
\label{app:prompts}

\begin{figure}[t]
    \centering
    \includegraphics[width=0.5\linewidth]{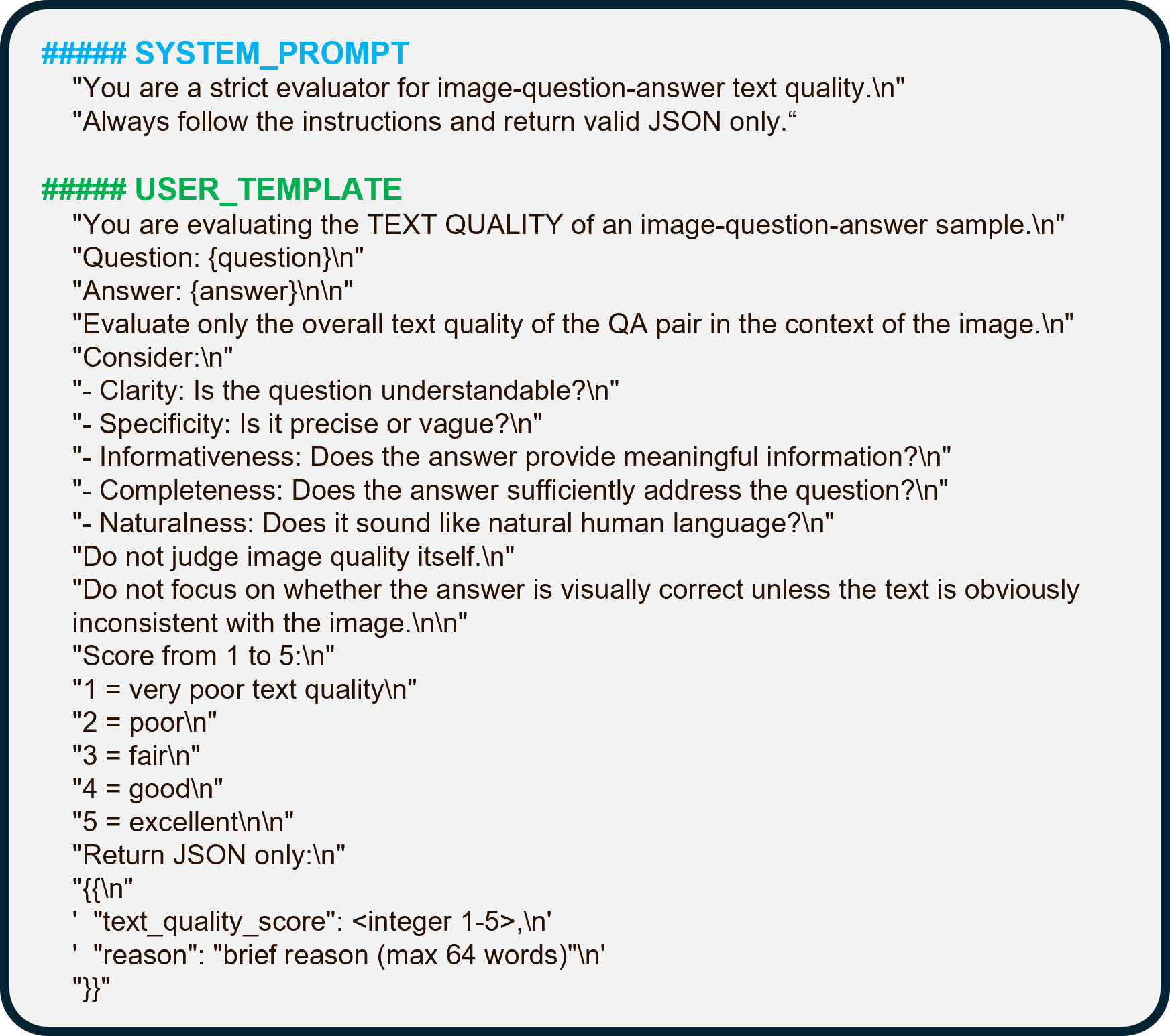}
    \caption{Scoring prompt for estimating the text-quality score $q^{txt}$ on Vision-Flan.}
    \label{fig:prompt_vf_text}
\end{figure}

\begin{figure}[t]
    \centering
    \includegraphics[width=0.5\linewidth]{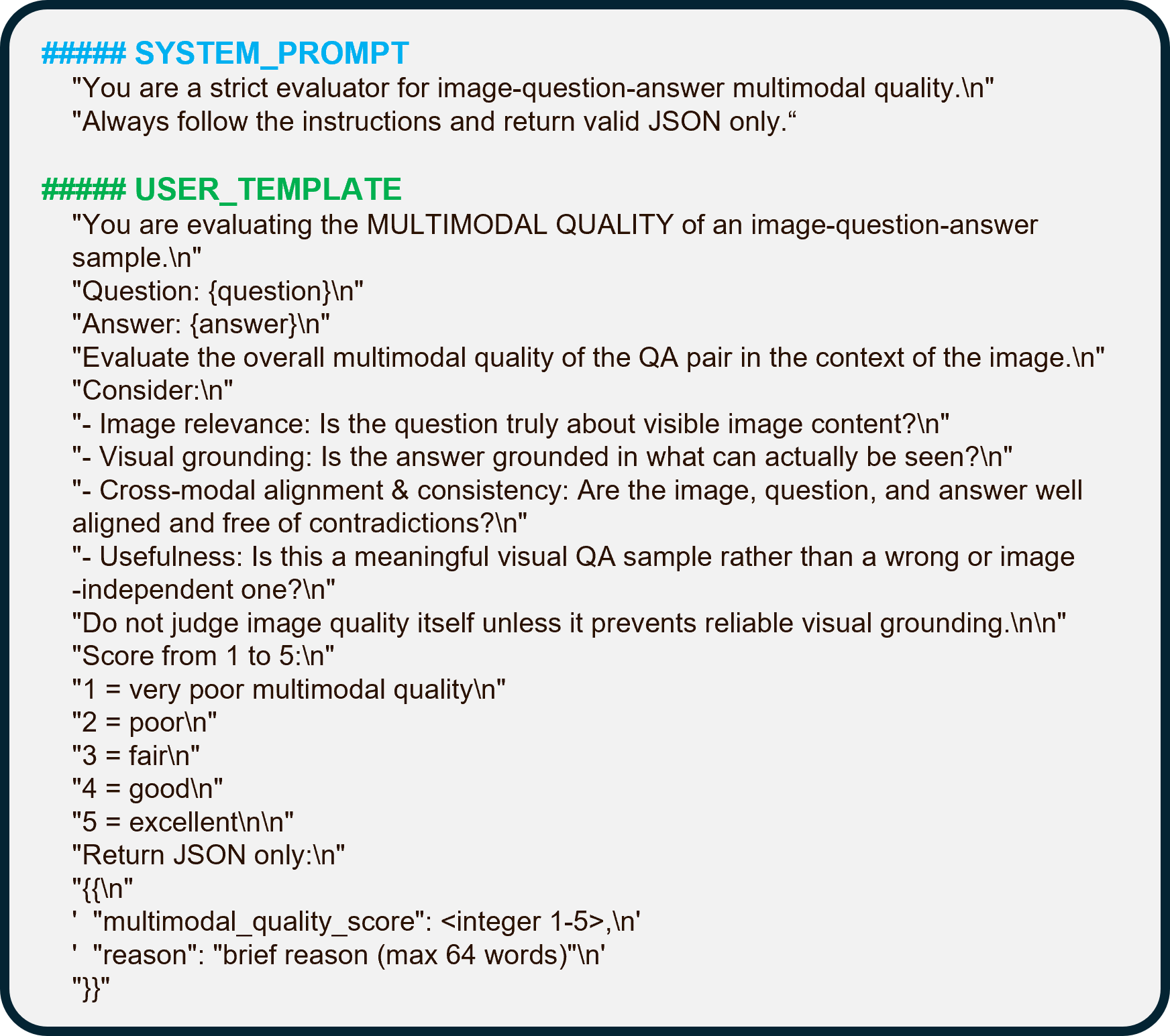}
    \caption{Scoring prompt for estimating the multimodal-quality score $q^{mm}$ on Vision-Flan.}
    \label{fig:prompt_vf_mm}
\end{figure}

\begin{figure}[t]
    \centering
    \includegraphics[width=0.5\linewidth]{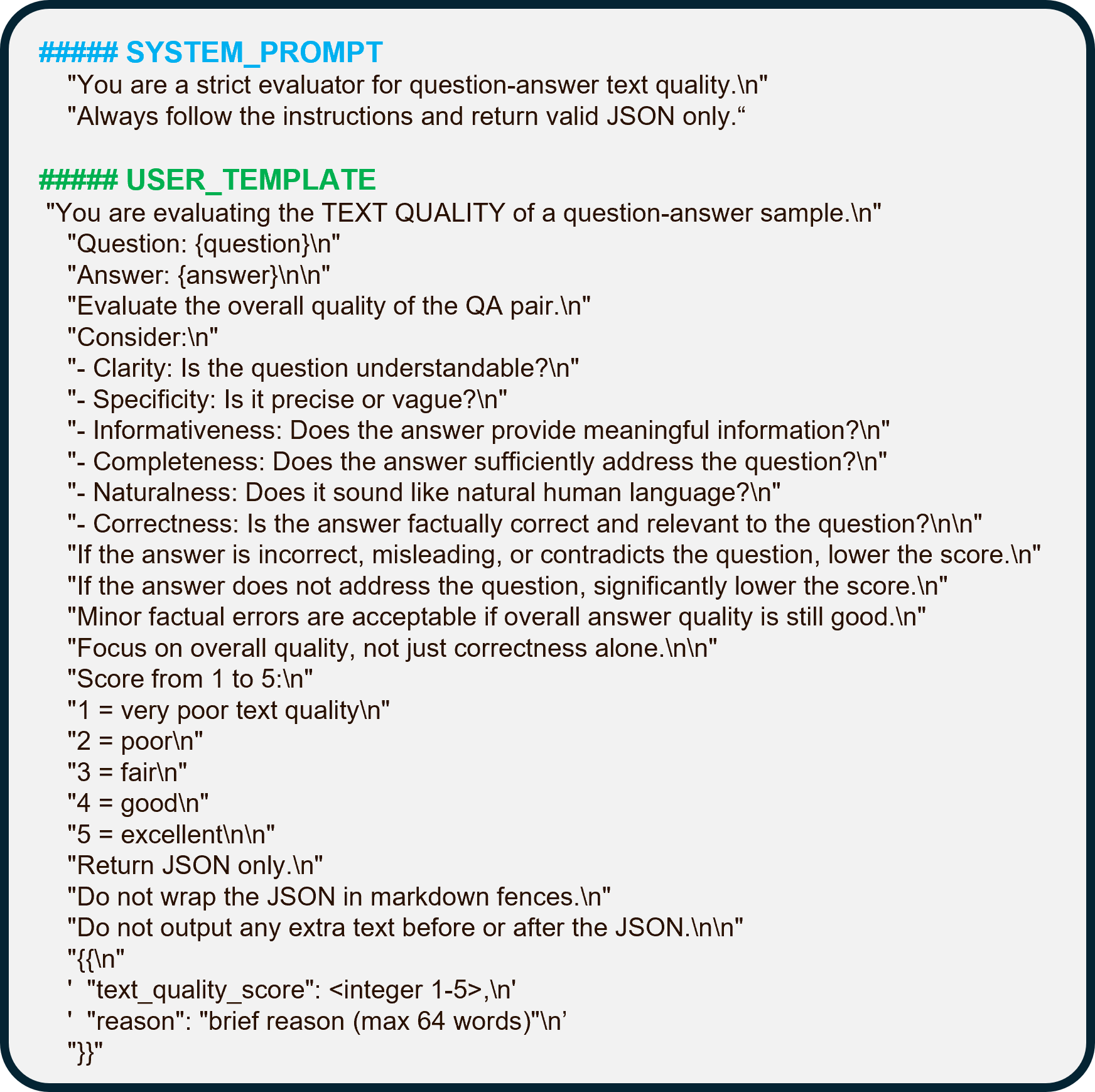}
    \caption{Scoring prompt for estimating the text-quality score $q^{txt}$ on OpenHermes.}
    \label{fig:prompt_op_text}
\end{figure}

\begin{figure}[t]
    \centering
    \includegraphics[width=0.8\linewidth]{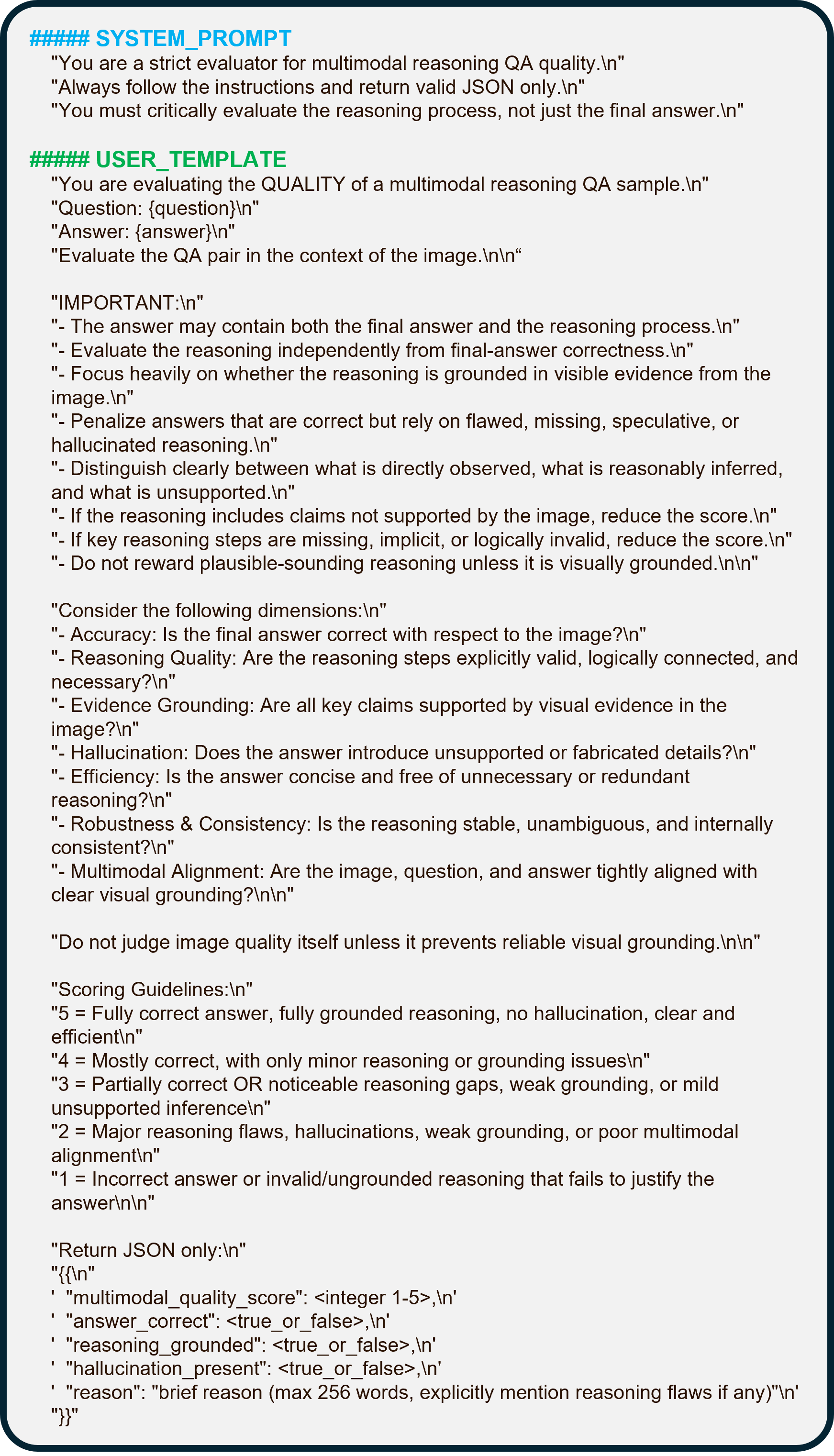}
    \caption{Scoring prompt for estimating the image-text reasoning quality score $q^{mm}$ on LLaVA-CoT.}
    \label{fig:prompt_llavacot}
\end{figure}

\begin{figure}[t]
    \centering
    \includegraphics[width=0.8\linewidth]{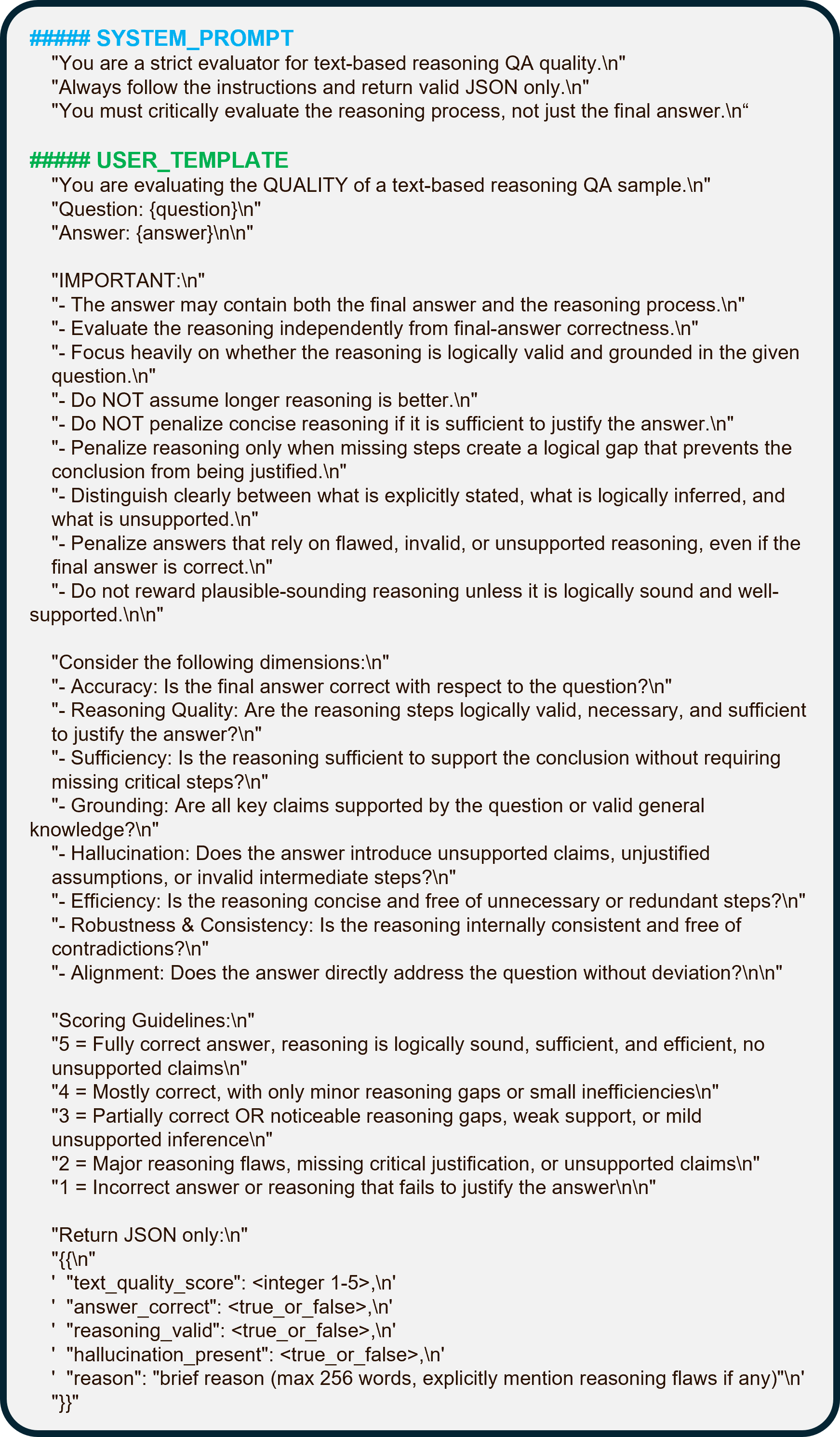}
    \caption{Scoring prompt for estimating the text-reasoning quality score $q^{txt}$ on SmallThoughts.}
    \label{fig:prompt_smallthoughts}
\end{figure}

We provide the scoring prompts used for quality estimation in QMAC. Specifically,
Figure~\ref{fig:prompt_vf_text} and Figure~\ref{fig:prompt_vf_mm} present the prompts used to estimate
$q^{txt}$ and $q^{mm}$ for Vision-Flan, respectively. Figure~\ref{fig:prompt_op_text} presents the prompt used to estimate $q^{txt}$ for OpenHermes.
Figure~\ref{fig:prompt_llavacot} presents the prompt used to estimate $q^{mm}$ for LLaVA-CoT. Figure~\ref{fig:prompt_smallthoughts} presents the prompt used to estimate $q^{txt}$ for SmallThoughts. Although the prompts specify a JSON-style output format with a reason field, we stop
generation when the model starts to emit the reason field marker and extract the preceding
quality score. This avoids decoding the explanatory text and reduces the computational overhead of
quality estimation.

\section{Evaluation}
\label{app:Evaluation}

\subsection{Evaluation Benchmarks}
\label{sec:benchmarks}

We choose benchmarks according to the modality and target capability of each dataset. For Vision-Flan, we use benchmarks covering general multimodal understanding, visual question answering, OCR, document understanding, chart understanding, and science-diagram reasoning. For OpenHermes, we use benchmarks covering instruction following, knowledge, reading comprehension, coding, mathematical reasoning, and multi-hop reasoning. For LLaVA-CoT, we combine multimodal mathematical and logical reasoning benchmarks with general vision-language benchmarks to evaluate both reasoning ability and overall multimodal robustness. For Qwen2-VL-2B-Instruct trained on LLaVA-CoT, we observe that full-data training still yields extremely poor performance on We-Math~\cite{qiao2025we}, LogicVista~\cite{xiao2024logicvista}, DynaMath~\cite{zou2024dynamath}, and MMStar~\cite{chen2024we}, making these benchmarks less informative for comparing data selection methods under this target model. Therefore, for this model we evaluate on the remaining eight benchmarks. For SmallThoughts, we focus on mathematical and structured reasoning benchmarks across school-level, competition-level, exam-style, STEM, word-problem, and tabular settings. The full benchmark list is shown in Table~\ref{tab:eval_benchmarks}.

\subsection{Detailed Evaluation Results}
\label{sec:detailed_results}
For the main experiments, Table~\ref{tab:full_comparison_vf} reports the full comparison on Vision-Flan, including QMAC and all data selection baselines. 
Table~\ref{tab:llava_v15_13b_vf} further reports the Vision-Flan results using LLaVA-V1.5-13B as the target model. 
Tables~\ref{tab:openhermes_qwen25_3b} and~\ref{tab:openhermes_llama31_8b} report the detailed OpenHermes results using Qwen2.5-3B and Llama-3.1-8B, respectively. 
Tables~\ref{tab:llava_cot_llama32_11b_vision} and~\ref{tab:llava_cot_qwen2vl_2b} report the detailed LLaVA-CoT results using Llama-3.2-11B-Vision-Instruct and Qwen2-VL-2B-Instruct, respectively. 
Tables~\ref{tab:smallthoughts_qwen25_math_7b} and~\ref{tab:smallthoughts_qwen25_math_15b} report the detailed SmallThoughts results using Qwen2.5-Math-7B and Qwen2.5-Math-1.5B, respectively.

We also provide detailed benchmark-level results for the ablation studies discussed in the main paper. Table~\ref{tab:openhermes_constraint_ablation} corresponds to the quality and distribution constraint ablation in Section~\ref{sec:quality_distribution_ablation}. Table~\ref{tab:openhermes_embedding_ablation} corresponds to the embedding model ablation in Section~\ref{sec:quality_embedding_ablation}.  Tables~\ref{tab:smallthoughts_quality_ablation} and~\ref{tab:llavacot_quality_ablation} correspond to the quality model ablation in Section~\ref{sec:quality_quality_ablation}. 

\section{Dataset for Band Analysis}
\label{app:synthetic_construction}
Animals-10 (from https://huggingface.co/datasets/Rapidata/Animals-10) is an animal classification dataset that contains only images of different animal categories. We construct a synthetic image-text instruction dataset from it to analyze the sensitivity of different frequency bands.
Specifically, we select five animal categories, including butterfly, chicken, dog, horse, and squirrel.
For each category, we sample 1,800 images as clean samples and another 200 images as source samples for perturbation, resulting in 10,000 samples in total, with 9,000 clean samples and 1,000 perturbed samples.

\begin{figure}[t]
    \centering
    \includegraphics[width=0.75\linewidth]{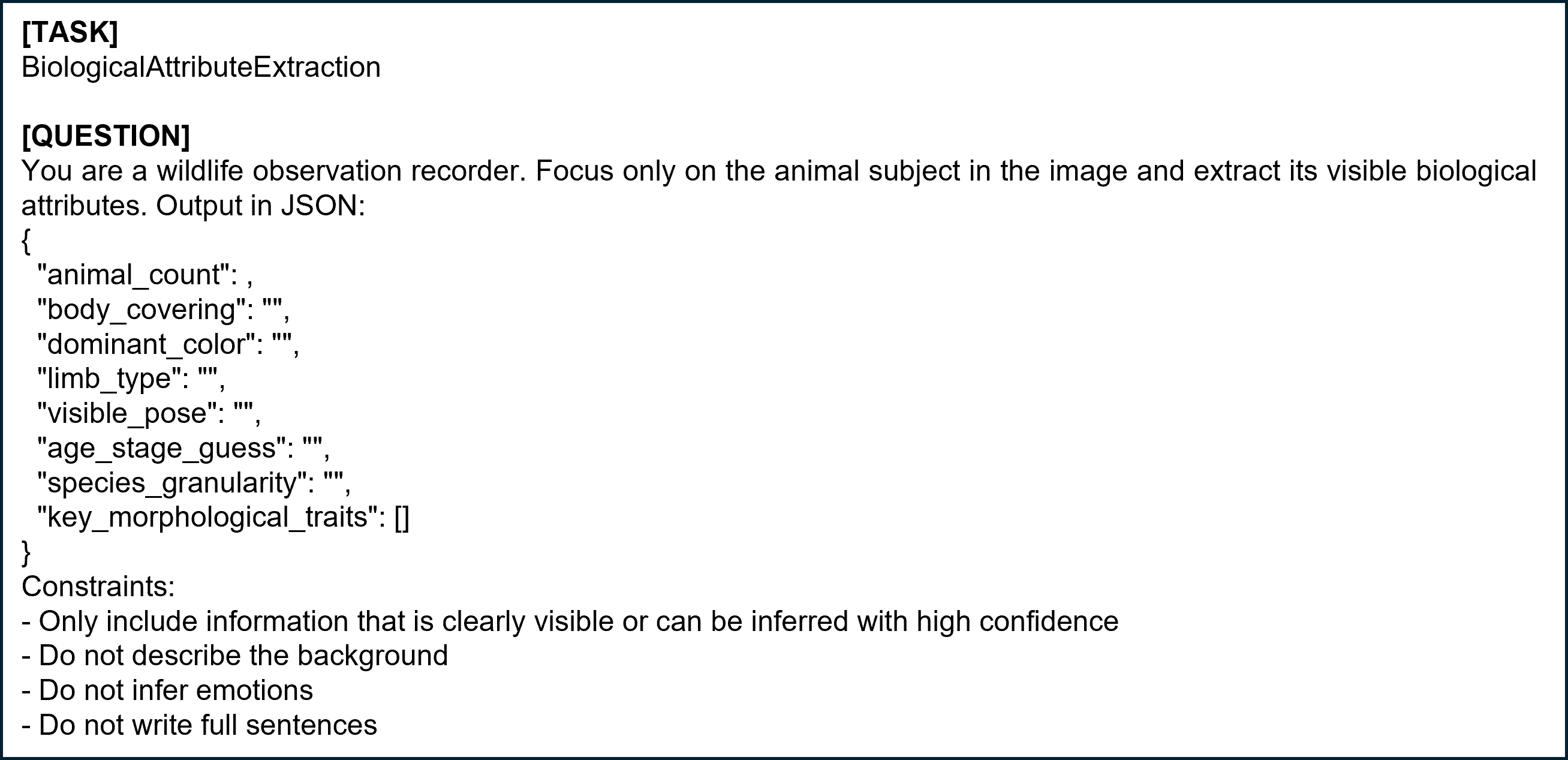}
    \caption{Example of BiologicalAttributeExtraction.}
    \label{fig:task1}
\end{figure}

\begin{figure}[t]
    \centering
    \includegraphics[width=0.75\linewidth]{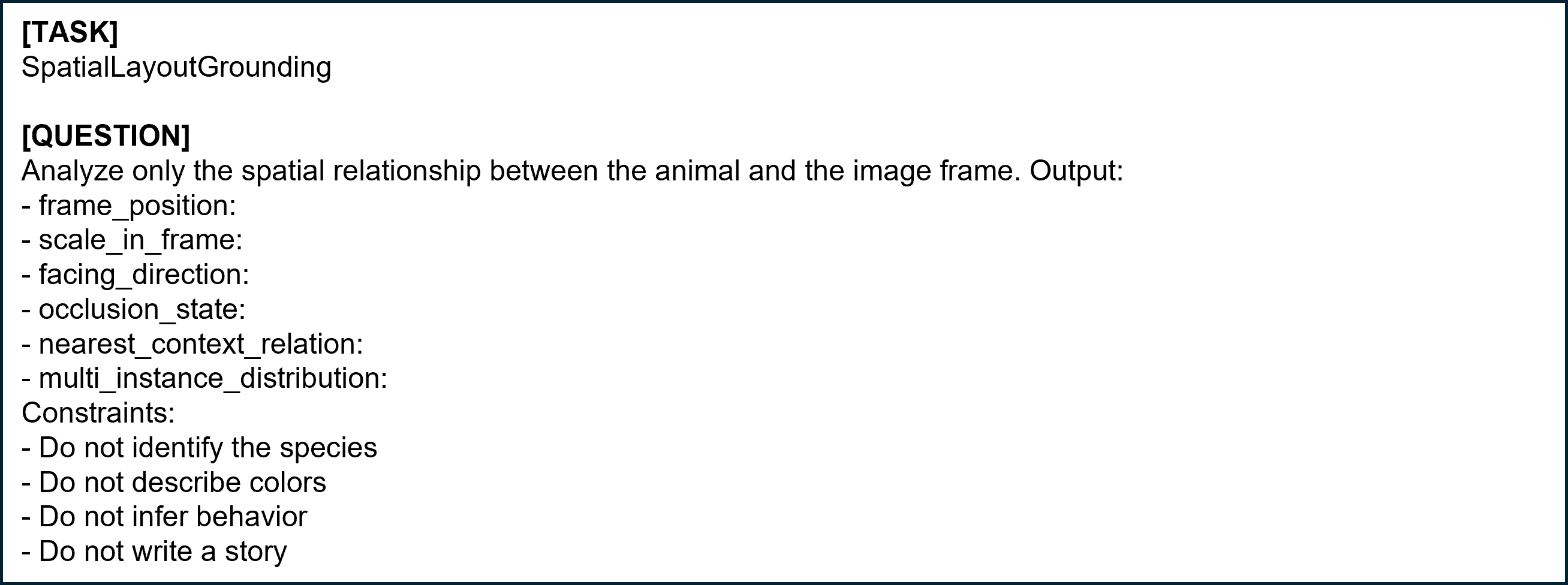}
    \caption{Example of SpatialLayoutGrounding.}
    \label{fig:task2}
\end{figure}

\begin{figure}[t]
    \centering
    \includegraphics[width=0.75\linewidth]{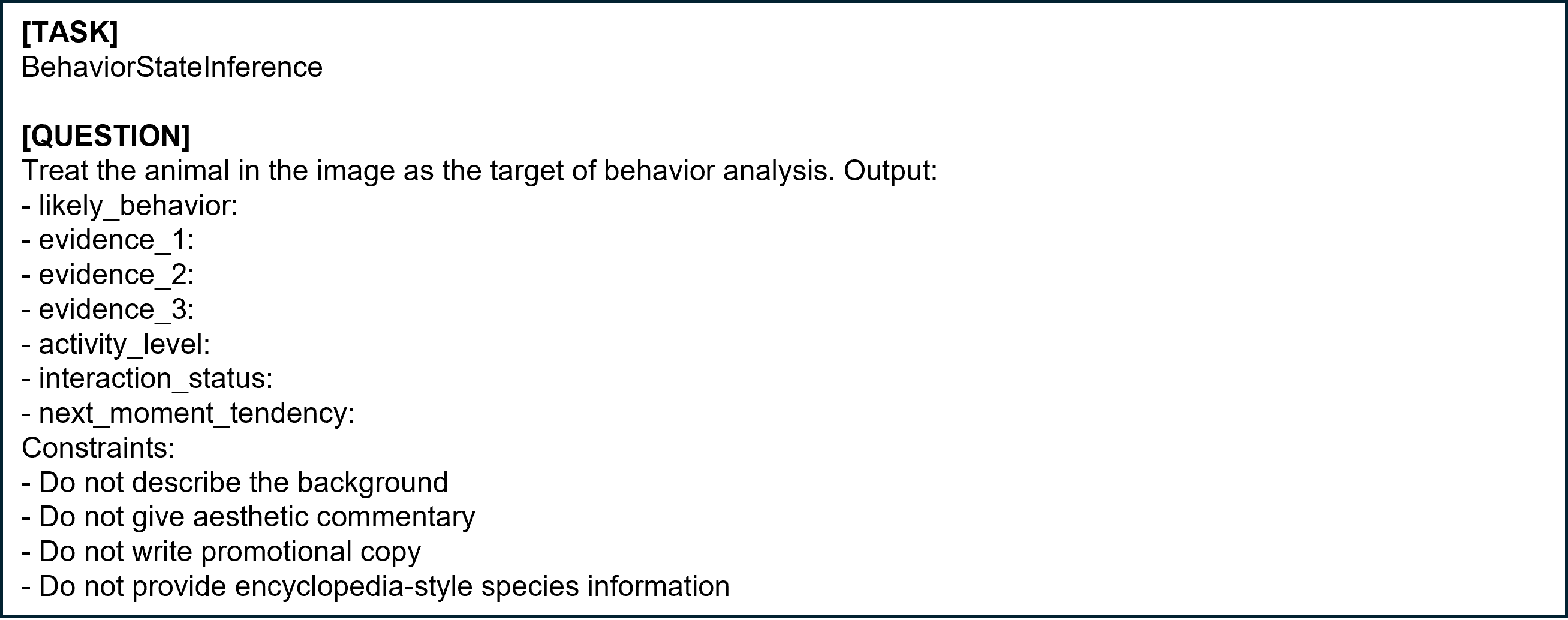}
    \caption{Example of BehaviorStateInference.}
    \label{fig:task3}
\end{figure}

\begin{figure}[t]
    \centering
    \includegraphics[width=0.75\linewidth]{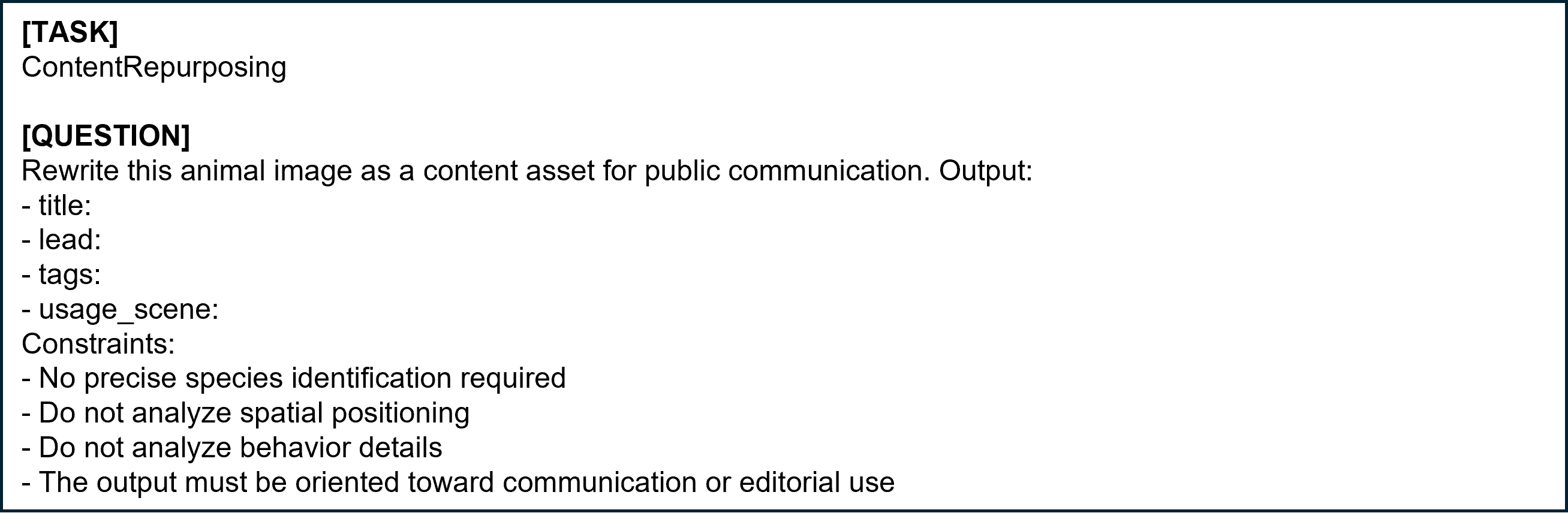}
    \caption{Example of ContentRepurposing.}
    \label{fig:task4}
\end{figure}

For instruction construction, we design four distinct question types:
BiologicalAttributeExtraction, SpatialLayoutGrounding, BehaviorStateInference, and ContentRepurposing.
Examples of these four question types are shown in Figure~\ref{fig:task1}, Figure~\ref{fig:task2}, Figure~\ref{fig:task3}, and Figure~\ref{fig:task4}, respectively.
These tasks emphasize different visual-language capabilities, including attribute recognition, spatial grounding, behavior understanding, and content transformation.
This design introduces clear task-level semantic distinctions, which helps form distinguishable regions in the embedding space.

To construct perturbed samples, we apply degradation to both the image and text sides.
For images, we introduce visual corruptions including noise, blur, JPEG compression, and resizing artifacts.
For text instructions, we apply perturbations such as word shuffling, dropping, and insertion.
The perturbed samples keep the original category and question type annotations, while the image and instruction are replaced with their degraded versions.

\section{Time Cost Analysis}
\label{app:time_cost}

We provide a concrete reference example for the computational cost of different
data selection methods. In the full experimental pipeline, some quality scoring,
instruction-tuning, and evaluation runs were conducted on different GPU types,
including NVIDIA A6000 and NVIDIA H200 GPUs, depending on model scale and
resource availability. Therefore, instead of claiming a single uniform compute
environment for all experiments, we report a controlled and directly comparable
example for the data selection stage.

Specifically, all compared data selection methods are evaluated on the
Vision-Flan dataset with 186K candidate samples under the same hardware
environment: four NVIDIA A6000 GPUs. We report the end-to-end GPU hours required
by the data selection stage, including the corresponding selection procedure but
excluding the subsequent instruction-tuning and evaluation costs. As shown in
Table~\ref{tab:time_cost}, QMAC introduces a moderate computational overhead
compared with faster selection baselines such as PRISM and ScalSelect. However,
these lower-cost methods obtain weaker downstream performance in our experiments.
XMAS, on the other hand, requires substantially higher cost than QMAC, consuming
35.10 GPU hours, which is about 1.40$\times$ the cost of QMAC. COINCIDE and CVS
have slightly lower selection cost than QMAC, but they are either tied to
specific vision-language selection assumptions or do not directly optimize
distributional consistency between the selected subset and the full dataset.

These results suggest that QMAC achieves a favorable cost--performance trade-off:
it is more efficient than the most expensive competitive baseline, while its
additional cost over cheaper baselines is compensated by stronger downstream
performance. Moreover, the selection cost is a one-time preprocessing cost and is
substantially smaller than the cost of repeatedly instruction tuning large target
models on the full dataset.

\newpage

\begin{table}[t]
\centering
\small
\caption{Evaluation benchmarks used for each dataset.}
\label{tab:eval_benchmarks}
\begin{tabular}{cl}
\toprule
\textbf{Dataset} & \textbf{Benchmark} \\
\midrule
\multirow[c]{12}{*}{Vision-Flan}
& GQA~\cite{hudson2019gqa} \\
& VizWiz~\cite{gurari2018vizwiz} \\
& TextVQA~\cite{singh2019towards} \\
& ScienceQA-IMG~\cite{saikh2022scienceqa} \\
& MME~\cite{liang2024survey} \\
& MMBench~\cite{liu2024mmbench} \\
& AI2D~\cite{kembhavi2016diagram} \\
& ChartQA~\cite{masry2022chartqa} \\
& DocVQA~\cite{mathew2021docvqa} \\
& InfoVQA~\cite{mathew2022infographicvqa} \\
& MMStar~\cite{chen2024we} \\
& OCRBench~\cite{liu2024ocrbench} \\
\midrule

\multirow[c]{12}{*}{OpenHermes}
& ARC-E~\cite{clark2018think} \\
& ARC-C~\cite{clark2018think} \\
& GPQA~\cite{rein2023gpqa} \\
& IFEval~\cite{zhou2023instruction} \\
& HumanEval~\cite{chen2021evaluating} \\
& TriviaQA~\cite{joshi2017triviaqa} \\
& SQuAD 2.0~\cite{rajpurkar-etal-2018-know} \\
& DROP~\cite{dua-etal-2019-drop} \\
& MATH~\cite{hendrycks2021measuring} \\
& MMLU~\cite{hendrycks2020measuring} \\
& GSM8K~\cite{cobbe2021training} \\
& HotpotQA~\cite{yang-etal-2018-hotpotqa} \\
\midrule

\multirow[c]{12}{*}{LLaVA-CoT}
& MATH-Vision~\cite{wang2024measuring} \\
& We-Math~\cite{qiao2025we} \\
& LogicVista~\cite{xiao2024logicvista} \\
& DynaMath~\cite{zou2024dynamath} \\
& MMStar~\cite{chen2024we} \\
& MME~\cite{liang2024survey} \\
& MMBench-EN~\cite{liu2024mmbench} \\
& ScienceQA-IMG~\cite{saikh2022scienceqa} \\
& AI2D~\cite{kembhavi2016diagram} \\
& ChartQA~\cite{masry2022chartqa} \\
& InfoVQA~\cite{mathew2022infographicvqa} \\
& OCRBench~\cite{liu2024ocrbench} \\
\midrule

\multirow[c]{18}{*}{SmallThoughts}
& AMC23~\cite{yang2024qwen2} \\
& AQuA~\cite{ling-etal-2017-program} \\
& ASDiv~\cite{miao-etal-2020-diverse} \\
& CARP-EN~\cite{zhang2023evaluating} \\
& CMATH~\cite{wei2023cmath} \\
& CN Middle School~\cite{yang2024qwen2} \\
& College Math~\cite{tang2024mathscale} \\
& GaoKao Math Cloze~\cite{zhong-etal-2024-agieval} \\
& Gaokao 2023 EN~\cite{yang2024qwen2} \\
& GSM8K~\cite{cobbe2021training} \\
& MATH~\cite{hendrycks2021measuring} \\
& MAWPS~\cite{koncel-kedziorski-etal-2016-mawps} \\
& Minerva Math~\cite{lewkowycz2022solving} \\
& MMLU-STEM~\cite{hendrycks2020measuring} \\
& OlympiadBench~\cite{he2024olympiadbench} \\
& SAT Math~\cite{zhong-etal-2024-agieval} \\
& SVAMP~\cite{patel2021nlp} \\
& TabMWP~\cite{lu2022dynamic} \\
\bottomrule
\end{tabular}
\end{table}

\begin{table*}[htbp]
  \centering
  \caption{Main experimental results on Vision-Flan using LLaVA-V1.5-7B as the target model. The best result in each column is highlighted in bold.}
  \label{tab:full_comparison_vf}

  \begin{subtable}{\textwidth}
    \centering
    \caption{5\% data subset.}
    \label{tab:vfsubset_5}
    \resizebox{\textwidth}{!}{%
      \begin{tabular}{lcccccccccccccc}
      \toprule
      \textbf{Method} & \textbf{GQA} & \textbf{VizWiz} & \textbf{TextVQA} & \textbf{SQA-I} & \textbf{MME} & \textbf{MMB-CN} & \textbf{MMB-EN} & \textbf{AI2D} & \textbf{ChartQA} & \textbf{DocVQA} & \textbf{InfoVQA} & \textbf{MMStar} & \textbf{OCRBench} & \textbf{ARP} \\
      \midrule
        Full Data & 47.30 & 55.88 & 35.83 & 61.43 & 1270.40 & 50.26 & 55.67 & 52.14 & 15.68 & 15.80 & 15.53 & 35.32 & 26.20 & 100.00 \\
        \midrule
        Random & 42.93 & 55.79 & 36.54 & 60.34 & 1098.02 & 36.40 & 26.19 & 37.82 & 15.96 & 16.46 & 19.99 & 34.77 & 27.80 & 92.96 \\
        XMAS & 40.38 & 54.15 & 37.06 & 62.42 & 705.48 & 39.95 & 39.95 & 41.22 & 15.40 & 17.98 & 19.80 & 34.37 & 27.50 & 93.46 \\
        COINCIDE & 42.15 & 53.48 & \textbf{38.00} & 55.03 & 1000.71 & 29.12 & 41.24 & 36.40 & 16.64 & 17.89 & 17.79 & 33.51 & 28.70 & 92.26 \\
        D2Prune & 41.02 & 55.99 & 37.47 & 60.59 & 928.95 & 36.94 & 44.93 & 38.54 & 16.60 & 17.68 & 20.58 & 33.47 & 27.90 & 95.61 \\
        EL2N & \textbf{43.26} & 54.93 & 35.94 & 54.34 & 898.77 & 39.78 & 18.90 & 34.03 & 15.80 & 16.23 & 18.36 & 34.13 & 28.20 & 88.74 \\
        SemDeDup & 43.09 & 55.72 & 37.31 & 63.11 & 1115.77 & 41.32 & 48.71 & 41.16 & \textbf{17.80} & 19.02 & 19.42 & 33.09 & 27.80 & \textbf{99.45} \\
        CLIP Score & 39.24 & 54.77 & 31.80 & 57.71 & 549.88 & 38.06 & 40.38 & 37.63 & 15.92 & 16.78 & 19.69 & 33.75 & \textbf{29.20} & 89.92 \\
        CVS & 38.42 & \textbf{57.46} & 37.98 & 63.41 & 800.81 & \textbf{45.88} & \textbf{52.66} & 39.44 & 15.12 & 18.48 & \textbf{20.74} & \textbf{35.59} & 27.80 & 97.82 \\
        PRISM & 29.97 & 52.06 & 26.73 & 49.33 & 1106.50 & 35.13 & 16.67 & 29.99 & 17.00 & \textbf{19.56} & 19.35 & 35.26 & 27.90 & 86.08 \\
        SCAL & 39.52 & 53.96 & 36.08 & 62.87 & 838.18 & 38.49 & 43.64 & 43.13 & 15.28 & 17.67 & 19.07 & 34.01 & 27.50 & 93.86 \\
        \rowcolor{lightcyan}
        QMAC (Ours) & 42.56 & 54.18 & 37.41 & \textbf{63.81} & \textbf{1173.75} & 41.58 & 52.32 & \textbf{43.56} & 16.32 & 17.96 & 19.18 & 32.86 & 28.00 & 99.16 \\
        \bottomrule
      \end{tabular}%
    }
  \end{subtable}

  \vspace{10pt}

  \begin{subtable}{\textwidth}
    \centering
    \caption{10\% data subset.}
    \label{tab:vfsubset_10}
    \resizebox{\textwidth}{!}{%
      \begin{tabular}{lcccccccccccccc}
      \toprule
      \textbf{Method} & \textbf{GQA} & \textbf{VizWiz} & \textbf{TextVQA} & \textbf{SQA-I} & \textbf{MME} & \textbf{MMB-CN} & \textbf{MMB-EN} & \textbf{AI2D} & \textbf{ChartQA} & \textbf{DocVQA} & \textbf{InfoVQA} & \textbf{MMStar} & \textbf{OCRBench} & \textbf{ARP} \\
      \midrule
        Full Data & 47.30 & 55.88 & 35.83 & 61.43 & 1270.40 & 50.26 & 55.67 & 52.14 & 15.68 & 15.80 & 15.53 & 35.32 & 26.20 & 100.00 \\
        \midrule
        Random & 43.31 & 56.02 & 36.13 & 60.92 & 1152.77 & 39.61 & 31.88 & 44.50 & 16.52 & 17.27 & 18.08 & 34.36 & 27.90 & 95.30 \\
        XMAS & 42.73 & 55.82 & 36.55 & 64.40 & 1008.41 & 45.19 & 54.38 & 43.95 & 16.20 & 18.39 & 18.19 & 34.28 & 28.50 & 99.31 \\
        COINCIDE & 43.33 & 57.38 & 38.70 & 60.83 & 1032.00 & 43.56 & 41.24 & 40.16 & 15.84 & 18.96 & 19.25 & 34.27 & 29.00 & 97.93 \\
        D2Prune & 43.47 & 56.40 & 37.81 & 63.01 & 1146.50 & \textbf{47.68} & 51.80 & 43.01 & 16.44 & 17.38 & 18.78 & 34.73 & 29.00 & 100.49 \\
        EL2N & \textbf{44.08} & 56.99 & 38.10 & 47.99 & 1088.62 & 39.52 & 31.70 & 29.92 & 15.64 & 18.78 & 18.78 & 34.94 & 28.30 & 92.67 \\
        SemDeDup & 43.93 & 56.39 & 37.20 & 58.65 & 1236.25 & 47.51 & 45.96 & 39.31 & 15.72 & 18.89 & 17.58 & 36.00 & 28.50 & 98.97 \\
        CLIP Score & 41.02 & 53.63 & 38.11 & 59.84 & 725.09 & 34.28 & 30.50 & 35.56 & 16.36 & 20.29 & 22.72 & 34.90 & \textbf{29.10} & 94.13 \\
        CVS & 39.49 & 57.66 & \textbf{38.99} & 66.78 & 930.81 & 46.13 & 54.30 & 43.98 & 16.24 & 21.18 & 21.72 & \textbf{37.70} & 28.70 & 103.45 \\
        PRISM & 35.21 & 48.80 & 37.87 & 61.97 & 751.61 & 38.49 & 44.16 & 44.88 & 16.44 & \textbf{23.76} & \textbf{23.37} & 35.92 & 28.80 & 98.99 \\
        SCAL & 42.73 & 56.26 & 37.68 & 60.88 & 1101.46 & 44.42 & \textbf{54.81} & 44.49 & 14.84 & 18.29 & 16.71 & 33.56 & 27.80 & 97.94 \\
        \rowcolor{lightcyan}
        QMAC (Ours) & 42.85 & \textbf{57.78} & 38.59 & \textbf{66.98} & \textbf{1242.66} & 46.99 & 51.37 & \textbf{47.60} & \textbf{17.08} & 19.07 & 19.59 & 36.03 & 28.60 & \textbf{104.04} \\
        \bottomrule
      \end{tabular}%
    }
  \end{subtable}

  \vspace{10pt}

  \begin{subtable}{\textwidth}
    \centering
    \caption{15\% data subset.}
    \label{tab:vfsubset_15}
    \resizebox{\textwidth}{!}{%
      \begin{tabular}{lcccccccccccccc}
      \toprule
      \textbf{Method} & \textbf{GQA} & \textbf{VizWiz} & \textbf{TextVQA} & \textbf{SQA-I} & \textbf{MME} & \textbf{MMB-CN} & \textbf{MMB-EN} & \textbf{AI2D} & \textbf{ChartQA} & \textbf{DocVQA} & \textbf{InfoVQA} & \textbf{MMStar} & \textbf{OCRBench} & \textbf{ARP} \\
      \midrule
        Full Data & 47.30 & 55.88 & 35.83 & 61.43 & 1270.40 & 50.26 & 55.67 & 52.14 & 15.68 & 15.80 & 15.53 & 35.32 & 26.20 & 100.00 \\
        \midrule
        Random & 43.19 & 56.33 & 35.41 & 59.59 & 1230.04 & 42.44 & 38.06 & 41.06 & 15.52 & 16.26 & 16.26 & 32.27 & 27.60 & 93.82 \\
        XMAS & 44.68 & 56.52 & 37.19 & 61.18 & 1012.67 & 40.72 & 46.99 & 44.62 & 15.84 & 17.83 & 17.88 & 34.95 & 27.50 & 97.13 \\
        COINCIDE & 44.28 & 56.44 & 37.52 & 55.73 & 1221.20 & 45.96 & 39.26 & 37.50 & 17.24 & 19.03 & 18.32 & 35.59 & 28.80 & 98.39 \\
        D2Prune & 43.96 & 57.16 & 38.40 & 61.63 & 1059.64 & 44.67 & 49.48 & 44.66 & 15.80 & 17.62 & 18.60 & 34.13 & 28.30 & 98.94 \\
        EL2N & 45.81 & 57.45 & 38.52 & 53.69 & 1105.20 & 46.22 & 39.18 & 43.39 & 14.68 & 19.06 & 17.35 & 34.56 & 28.40 & 96.87 \\
        SemDeDup & \textbf{50.00} & 51.57 & 37.30 & 59.69 & \textbf{1247.46} & 43.90 & 45.45 & 43.13 & 17.04 & 18.66 & 18.93 & 35.24 & 28.00 & 100.35 \\
        CLIP Score & 41.33 & 55.10 & 38.44 & \textbf{65.25} & 640.09 & 39.60 & 41.67 & 48.41 & 16.68 & 20.19 & 21.96 & 37.57 & \textbf{29.30} & 99.24 \\
        CVS & 43.46 & \textbf{57.93} & \textbf{40.57} & 63.46 & 1123.40 & 47.17 & 51.03 & 42.23 & 16.00 & \textbf{21.36} & \textbf{21.98} & \textbf{37.64} & 29.00 & 104.85 \\
        PRISM & 39.25 & 52.24 & 38.62 & 61.23 & 770.61 & 41.67 & 41.07 & 43.78 & \textbf{17.96} & 21.12 & 21.00 & 36.22 & 29.10 & 98.64 \\
        SCAL & 43.15 & 57.90 & 38.47 & 57.56 & 1062.93 & 46.91 & 43.99 & 36.40 & 15.96 & 18.82 & 16.69 & 35.53 & 27.70 & 96.65 \\
        \rowcolor{lightcyan}
        QMAC (Ours) & 45.26 & 56.10 & 39.85 & 63.66 & 1239.03 & \textbf{51.63} & \textbf{57.99} & \textbf{48.77} & 16.00 & 20.75 & 19.57 & 34.89 & 29.20 & \textbf{106.04} \\
        \bottomrule
      \end{tabular}%
    }
  \end{subtable}
\end{table*}

% Table 6
\begin{table*}[htbp]
  \centering
  \caption{Main experimental results on Vision-Flan using LLaVA-V1.5-13B as the target model. The best result in each column is highlighted in bold.}
  
  \label{tab:llava_v15_13b_vf}

  \begin{subtable}{\textwidth}
    \centering
    \caption{5\% data subset.}
    \label{tab:llava_v15_13b_5}
    \resizebox{\textwidth}{!}{%
      \begin{tabular}{lcccccccccccccc}
      \toprule
      \textbf{Method} & \textbf{GQA} & \textbf{VizWiz} & \textbf{TextVQA} & \textbf{SQA-I} & \textbf{MME} & \textbf{MMB-CN} & \textbf{MMB-EN} & \textbf{AI2D} & \textbf{ChartQA} & \textbf{DocVQA} & \textbf{InfoVQA} & \textbf{MMStar} & \textbf{OCRBench} & \textbf{ARP} \\
      \midrule
        Full Data & 49.02 & 55.92 & 35.82 & 70.60 & 1318.89 & 56.44 & 60.74 & 57.67 & 19.44 & 20.03 & 19.87 & 35.65 & 29.50 & 100.00 \\
      \midrule
        Random & \textbf{43.02} & \textbf{56.82} & 34.35 & 63.46 & 1178.91 & \textbf{36.25} & 41.23 & \textbf{45.66} & 14.88 & 14.87 & 18.17 & \textbf{32.17} & 29.00 & 85.12 \\
        \rowcolor{lightcyan}
        QMAC (Ours) & 42.21 & 56.00 & \textbf{36.76} & \textbf{65.05} & \textbf{1263.59} & 34.85 & \textbf{43.91} & 38.27 & \textbf{16.56} & \textbf{18.39} & \textbf{19.20} & 30.93 & \textbf{29.70} & \textbf{87.56} \\
      \bottomrule
      \end{tabular}%
    }
  \end{subtable}

  \vspace{10pt}

  \begin{subtable}{\textwidth}
    \centering
    \caption{10\% data subset.}
    \label{tab:llava_v15_13b_10}
    \resizebox{\textwidth}{!}{%
      \begin{tabular}{lcccccccccccccc}
      \toprule
      \textbf{Method} & \textbf{GQA} & \textbf{VizWiz} & \textbf{TextVQA} & \textbf{SQA-I} & \textbf{MME} & \textbf{MMB-CN} & \textbf{MMB-EN} & \textbf{AI2D} & \textbf{ChartQA} & \textbf{DocVQA} & \textbf{InfoVQA} & \textbf{MMStar} & \textbf{OCRBench} & \textbf{ARP} \\
      \midrule
        Full Data & 49.02 & 55.92 & 35.82 & 70.60 & 1318.89 & 56.44 & 60.74 & 57.67 & 19.44 & 20.03 & 19.87 & 35.65 & 29.50 & 100.00 \\
      \midrule
        Random & 44.98 & \textbf{58.13} & 36.43 & 68.26 & 1176.93 & 47.60 & \textbf{55.56} & 49.63 & \textbf{16.56} & \textbf{16.97} & \textbf{19.08} & 31.35 & 29.60 & 92.26 \\
        \rowcolor{lightcyan}
        QMAC (Ours) & \textbf{45.00} & 53.16 & \textbf{36.99} & \textbf{68.62} & \textbf{1277.66} & \textbf{47.94} & 54.82 & \textbf{52.53} & 15.76 & 16.70 & 18.32 & \textbf{33.02} & \textbf{30.00} & \textbf{92.42} \\
      \bottomrule
      \end{tabular}%
    }
  \end{subtable}

  \vspace{10pt}

  \begin{subtable}{\textwidth}
    \centering
    \caption{15\% data subset.}
    \label{tab:llava_v15_13b_15}
    \resizebox{\textwidth}{!}{%
      \begin{tabular}{lcccccccccccccc}
      \toprule
      \textbf{Method} & \textbf{GQA} & \textbf{VizWiz} & \textbf{TextVQA} & \textbf{SQA-I} & \textbf{MME} & \textbf{MMB-CN} & \textbf{MMB-EN} & \textbf{AI2D} & \textbf{ChartQA} & \textbf{DocVQA} & \textbf{InfoVQA} & \textbf{MMStar} & \textbf{OCRBench} & \textbf{ARP} \\
      \midrule
        Full Data & 49.02 & 55.92 & 35.82 & 70.60 & 1318.89 & 56.44 & 60.74 & 57.67 & 19.44 & 20.03 & 19.87 & 35.65 & 29.50 & 100.00 \\
      \midrule
        Random & 45.25 & 56.96 & \textbf{38.03} & 65.05 & 1176.15 & 42.70 & 46.99 & 48.32 & 16.60 & 18.38 & 18.23 & \textbf{34.86} & 29.40 & 91.14 \\
        \rowcolor{lightcyan}
        QMAC (Ours) & \textbf{45.79} & \textbf{58.32} & 37.56 & \textbf{69.61} & \textbf{1274.37} & \textbf{44.16} & \textbf{49.39} & \textbf{51.52} & \textbf{16.92} & \textbf{20.32} & \textbf{20.51} & 33.34 & \textbf{30.00} & \textbf{94.89} \\
      \bottomrule
      \end{tabular}%
    }
  \end{subtable}

\end{table*}

% To force this to appear as Table 7 in a standalone compilation, uncomment the next line.
\begin{table*}[htbp]
  \centering
  \caption{Main experimental results on OpenHermes using Qwen2.5-3B as the target model. The best result in each column is highlighted in bold.}
  \label{tab:openhermes_qwen25_3b}

  \begin{subtable}{\textwidth}
    \centering
    \caption{10\% data subset.}
    \label{tab:openhermes_qwen25_3b_a}
    \resizebox{\textwidth}{!}{%
      \begin{tabular}{lcccccccccccccc}
      \toprule
      \textbf{Method} & \textbf{ARC-E} & \textbf{ARC-C} & \textbf{GPQA} & \textbf{IFEval-P} & \textbf{IFEval-I} & \textbf{HumanEval} & \textbf{TriviaQA} & \textbf{SQuAD 2.0} & \textbf{DROP} & \textbf{MATH} & \textbf{MMLU} & \textbf{GSM8K} & \textbf{HotpotQA} & \textbf{ARP} \\
      \midrule
        Full Data & 89.77 & 83.05 & 22.73 & 29.94 & 41.25 & 61.59 & 47.22 & 15.20 & 54.62 & 36.20 & 50.72 & 75.51 & 8.79 & 100.00 \\
      \midrule
        Random & 86.60 & \textbf{80.00} & \textbf{9.60} & 24.40 & 36.93 & \textbf{65.24} & \textbf{47.40} & 13.60 & 57.11 & 35.86 & 44.16 & \textbf{76.26} & 16.05 & 98.16 \\
        \rowcolor{lightcyan}
        QMAC (Ours) & \textbf{87.30} & 77.29 & 9.09 & \textbf{26.43} & \textbf{38.25} & 64.43 & 47.28 & \textbf{16.46} & \textbf{61.34} & \textbf{36.36} & \textbf{46.96} & 75.44 & \textbf{19.00} & \textbf{103.52} \\
      \bottomrule
      \end{tabular}%
    }
  \end{subtable}

  \vspace{10pt}

  \begin{subtable}{\textwidth}
    \centering
    \caption{20\% data subset.}
    \label{tab:openhermes_qwen25_3b_b}
    \resizebox{\textwidth}{!}{%
      \begin{tabular}{lcccccccccccccc}
      \toprule
      \textbf{Method} & \textbf{ARC-E} & \textbf{ARC-C} & \textbf{GPQA} & \textbf{IFEval-P} & \textbf{IFEval-I} & \textbf{HumanEval} & \textbf{TriviaQA} & \textbf{SQuAD 2.0} & \textbf{DROP} & \textbf{MATH} & \textbf{MMLU} & \textbf{GSM8K} & \textbf{HotpotQA} & \textbf{ARP} \\
      \midrule
        Full Data & 89.77 & 83.05 & 22.73 & 29.94 & 41.25 & 61.59 & 47.22 & 15.20 & 54.62 & 36.20 & 50.72 & 75.51 & 8.79 & 100.00 \\
      \midrule
        Random & \textbf{88.89} & \textbf{82.03} & 9.09 & 25.51 & 36.81 & \textbf{64.63} & \textbf{47.75} & 12.16 & 54.24 & \textbf{35.88} & 49.90 & \textbf{76.50} & 16.00 & 98.34 \\
        \rowcolor{lightcyan}
        QMAC (Ours) & 88.71 & \textbf{82.03} & \textbf{13.13} & \textbf{29.39} & \textbf{41.25} & 64.02 & 47.67 & \textbf{14.83} & \textbf{60.42} & \textbf{35.88} & \textbf{50.43} & 75.74 & \textbf{18.12} & \textbf{105.51} \\
      \bottomrule
      \end{tabular}%
    }
  \end{subtable}

  \vspace{10pt}

  \begin{subtable}{\textwidth}
    \centering
    \caption{30\% data subset.}
    \label{tab:openhermes_qwen25_3b_c}
    \resizebox{\textwidth}{!}{%
      \begin{tabular}{lcccccccccccccc}
      \toprule
      \textbf{Method} & \textbf{ARC-E} & \textbf{ARC-C} & \textbf{GPQA} & \textbf{IFEval-P} & \textbf{IFEval-I} & \textbf{HumanEval} & \textbf{TriviaQA} & \textbf{SQuAD 2.0} & \textbf{DROP} & \textbf{MATH} & \textbf{MMLU} & \textbf{GSM8K} & \textbf{HotpotQA} & \textbf{ARP} \\
      \midrule
        Full Data & 89.77 & 83.05 & 22.73 & 29.94 & 41.25 & 61.59 & 47.22 & 15.20 & 54.62 & 36.20 & 50.72 & 75.51 & 8.79 & 100.00 \\
      \midrule
        Random & 88.89 & 81.02 & 10.10 & 26.99 & 38.73 & 61.59 & \textbf{47.87} & 12.04 & 56.49 & \textbf{36.50} & 49.04 & 75.66 & 10.06 & 93.94 \\
        \rowcolor{lightcyan}
        QMAC (Ours) & \textbf{89.42} & \textbf{81.69} & \textbf{11.62} & \textbf{27.36} & \textbf{40.05} & \textbf{64.02} & 47.70 & \textbf{14.25} & \textbf{60.28} & 36.22 & \textbf{49.85} & \textbf{76.04} & \textbf{17.71} & \textbf{103.63} \\
      \bottomrule
      \end{tabular}%
    }
  \end{subtable}
\end{table*}
% To force this to appear as Table 8 in a standalone compilation, uncomment the next line.
% \setcounter{table}{7}
\begin{table*}[htbp]
  \centering
  \caption{Main experimental results on OpenHermes using Llama-3.1-8B as the target model. The best result in each column is highlighted in bold.}
  \label{tab:openhermes_llama31_8b}

  \begin{subtable}{\textwidth}
    \centering
    \caption{10\% data subset.}
    \label{tab:openhermes_llama31_8b_a}
    \resizebox{\textwidth}{!}{%
      \begin{tabular}{lcccccccccccccc}
      \toprule
      \textbf{Method} & \textbf{ARC-E} & \textbf{ARC-C} & \textbf{GPQA} & \textbf{IFEval-P} & \textbf{IFEval-I} & \textbf{HumanEval} & \textbf{TriviaQA} & \textbf{SQuAD 2.0} & \textbf{DROP} & \textbf{MATH} & \textbf{MMLU} & \textbf{GSM8K} & \textbf{HotpotQA} & \textbf{ARP} \\
      \midrule
        Full Data & 76.37 & 63.39 & 23.23 & 32.53 & 43.53 & 30.49 & 60.02 & 19.71 & 11.59 & 7.92 & 38.97 & 2.20 & 0.84 & 100.00 \\
      \midrule
        Random & 65.51 & 51.86 & 8.59 & 28.65 & 39.21 & 26.22 & 60.70 & 10.60 & 13.63 & 3.44 & 43.44 & 1.67 & \textbf{0.79} & 82.01 \\
        \rowcolor{lightcyan}
        QMAC (Ours) & \textbf{70.37} & \textbf{57.63} & \textbf{14.14} & \textbf{33.64} & \textbf{44.96} & \textbf{29.88} & \textbf{61.21} & \textbf{12.23} & \textbf{19.70} & \textbf{6.60} & \textbf{46.09} & \textbf{1.90} & 0.63 & \textbf{95.82} \\
      \bottomrule
      \end{tabular}%
    }
  \end{subtable}

  \vspace{10pt}

  \begin{subtable}{\textwidth}
    \centering
    \caption{20\% data subset.}
    \label{tab:openhermes_llama31_8b_b}
    \resizebox{\textwidth}{!}{%
      \begin{tabular}{lcccccccccccccc}
      \toprule
      \textbf{Method} & \textbf{ARC-E} & \textbf{ARC-C} & \textbf{GPQA} & \textbf{IFEval-P} & \textbf{IFEval-I} & \textbf{HumanEval} & \textbf{TriviaQA} & \textbf{SQuAD 2.0} & \textbf{DROP} & \textbf{MATH} & \textbf{MMLU} & \textbf{GSM8K} & \textbf{HotpotQA} & \textbf{ARP} \\
      \midrule
        Full Data & 76.37 & 63.39 & 23.23 & 32.53 & 43.53 & 30.49 & 60.02 & 19.71 & 11.59 & 7.92 & 38.97 & 2.20 & 0.84 & 100.00 \\
      \midrule
        Random & 65.26 & 49.49 & 2.02 & 30.13 & 41.37 & 25.00 & \textbf{61.23} & 11.38 & 17.65 & 4.90 & 31.39 & 2.05 & 0.79 & 83.35 \\
        \rowcolor{lightcyan}
        QMAC (Ours) & \textbf{82.72} & \textbf{69.15} & \textbf{15.15} & \textbf{32.53} & \textbf{43.76} & \textbf{28.66} & 60.98 & \textbf{12.57} & \textbf{30.08} & \textbf{5.64} & \textbf{47.89} & \textbf{2.73} & \textbf{0.85} & \textbf{109.34} \\
      \bottomrule
      \end{tabular}%
    }
  \end{subtable}

  \vspace{10pt}

  \begin{subtable}{\textwidth}
    \centering
    \caption{30\% data subset.}
    \label{tab:openhermes_llama31_8b_c}
    \resizebox{\textwidth}{!}{%
      \begin{tabular}{lcccccccccccccc}
      \toprule
      \textbf{Method} & \textbf{ARC-E} & \textbf{ARC-C} & \textbf{GPQA} & \textbf{IFEval-P} & \textbf{IFEval-I} & \textbf{HumanEval} & \textbf{TriviaQA} & \textbf{SQuAD 2.0} & \textbf{DROP} & \textbf{MATH} & \textbf{MMLU} & \textbf{GSM8K} & \textbf{HotpotQA} & \textbf{ARP} \\
      \midrule
        Full Data & 76.37 & 63.39 & 23.23 & 32.53 & 43.53 & 30.49 & 60.02 & 19.71 & 11.59 & 7.92 & 38.97 & 2.20 & 0.84 & 100.00 \\
      \midrule
        Random & 53.09 & 47.12 & 5.56 & 29.39 & 39.69 & 29.88 & \textbf{61.38} & 4.72 & 11.99 & 4.28 & \textbf{36.04} & \textbf{1.74} & \textbf{0.91} & 77.76 \\
        \rowcolor{lightcyan}
        QMAC (Ours) & \textbf{77.43} & \textbf{63.05} & \textbf{11.11} & \textbf{31.79} & \textbf{41.97} & \textbf{32.93} & 60.64 & \textbf{15.94} & \textbf{26.44} & \textbf{5.46} & 33.34 & 1.52 & 0.72 & \textbf{97.70} \\
      \bottomrule
      \end{tabular}%
    }
  \end{subtable}
\end{table*}

% Table 9
% Requires in preamble:
% \usepackage[table]{xcolor}
% \definecolor{lightcyan}{RGB}{224,255,255}
\begin{table*}[htbp]
  \centering
  \caption{Main experimental results on LLaVA-CoT using Llama-3.2-11B-Vision-Instruct as the target model. The best result in each column is highlighted in bold.}
  \label{tab:llava_cot_llama32_11b_vision}

  \begin{subtable}{\textwidth}
    \centering
    \caption{5\% data subset.}
    \label{tab:llava_cot_llama32_11b_vision_a}
    \resizebox{\textwidth}{!}{%
      \begin{tabular}{lccccccccccccc}
      \toprule
      \textbf{Method} & \textbf{MATH-Vision} & \textbf{We-Math} & \textbf{LogicVista} & \textbf{DynaMath} & \textbf{MMStar} & \textbf{MME} & \textbf{MMB-EN} & \textbf{SQA-I} & \textbf{AI2D} & \textbf{ChartQA} & \textbf{InfoVQA} & \textbf{OCRBench} & \textbf{ARP} \\
      \midrule
        Full Data & 17.01 & 31.90 & 29.91 & 16.23 & 56.99 & 1549.25 & 77.92 & 95.44 & 76.94 & 85.68 & 62.29 & 73.50 & 100.00 \\
      \midrule
        Random & 14.08 & 27.99 & 26.12 & \textbf{16.07} & 52.35 & 1405.82 & 76.37 & 90.18 & 73.83 & 83.88 & 65.16 & 71.50 & 93.98 \\
        \rowcolor{lightcyan}
        QMAC (Ours) & \textbf{15.06} & \textbf{35.06} & \textbf{29.46} & 15.71 & \textbf{53.56} & \textbf{1470.20} & \textbf{77.49} & \textbf{90.98} & \textbf{74.50} & \textbf{84.40} & \textbf{66.38} & \textbf{72.00} & \textbf{98.10} \\
      \bottomrule
      \end{tabular}%
    }
  \end{subtable}

  \vspace{10pt}

  \begin{subtable}{\textwidth}
    \centering
    \caption{10\% data subset.}
    \label{tab:llava_cot_llama32_11b_vision_b}
    \resizebox{\textwidth}{!}{%
      \begin{tabular}{lccccccccccccc}
      \toprule
      \textbf{Method} & \textbf{MATH-Vision} & \textbf{We-Math} & \textbf{LogicVista} & \textbf{DynaMath} & \textbf{MMStar} & \textbf{MME} & \textbf{MMB-EN} & \textbf{SQA-I} & \textbf{AI2D} & \textbf{ChartQA} & \textbf{InfoVQA} & \textbf{OCRBench} & \textbf{ARP} \\
      \midrule
        Full Data & 17.01 & 31.90 & 29.91 & 16.23 & 56.99 & 1549.25 & 77.92 & 95.44 & 76.94 & 85.68 & 62.29 & 73.50 & 100.00 \\
      \midrule
        Random & 15.89 & 29.20 & 30.80 & 15.19 & 50.67 & 1471.44 & 76.55 & \textbf{91.77} & 74.48 & 84.12 & \textbf{66.30} & \textbf{74.20} & 96.85 \\
        \rowcolor{lightcyan}
        QMAC (Ours) & \textbf{17.43} & \textbf{39.25} & \textbf{31.70} & \textbf{15.77} & \textbf{53.85} & \textbf{1542.21} & \textbf{77.15} & 91.42 & \textbf{75.81} & \textbf{84.96} & 65.43 & 71.70 & \textbf{101.48} \\
      \bottomrule
      \end{tabular}%
    }
  \end{subtable}

  \vspace{10pt}

  \begin{subtable}{\textwidth}
    \centering
    \caption{15\% data subset.}
    \label{tab:llava_cot_llama32_11b_vision_c}
    \resizebox{\textwidth}{!}{%
      \begin{tabular}{lccccccccccccc}
      \toprule
      \textbf{Method} & \textbf{MATH-Vision} & \textbf{We-Math} & \textbf{LogicVista} & \textbf{DynaMath} & \textbf{MMStar} & \textbf{MME} & \textbf{MMB-EN} & \textbf{SQA-I} & \textbf{AI2D} & \textbf{ChartQA} & \textbf{InfoVQA} & \textbf{OCRBench} & \textbf{ARP} \\
      \midrule
        Full Data & 17.01 & 31.90 & 29.91 & 16.23 & 56.99 & 1549.25 & 77.92 & 95.44 & 76.94 & 85.68 & 62.29 & 73.50 & 100.00 \\
      \midrule
        Random & 15.56 & 29.54 & \textbf{32.37} & 15.02 & 53.44 & \textbf{1530.66} & 77.75 & 91.03 & 74.09 & \textbf{85.44} & 66.42 & 74.80 & 98.08 \\
        \rowcolor{lightcyan}
        QMAC (Ours) & \textbf{17.27} & \textbf{37.82} & 31.92 & \textbf{15.87} & \textbf{54.56} & 1510.22 & \textbf{78.61} & \textbf{91.87} & \textbf{75.62} & 85.20 & \textbf{66.87} & \textbf{75.20} & \textbf{101.86} \\
      \bottomrule
      \end{tabular}%
    }
  \end{subtable}

\end{table*}

% Table 10
\begin{table*}[htbp]
  \centering
  \caption{Main experimental results on LLaVA-CoT using Qwen2-VL-2B-Instruct as the target model. The best result in each column is highlighted in bold.}
  \label{tab:llava_cot_qwen2vl_2b}

  \begin{subtable}{\textwidth}
    \centering
    \caption{5\% data subset.}
    \label{tab:llava_cot_qwen2vl_2b_a}
    \resizebox{\textwidth}{!}{%
      \begin{tabular}{lccccccccccccc}
      \toprule
      \textbf{Method} & \textbf{MATH-Vision} & \textbf{MME} & \textbf{MMB-EN} & \textbf{SQA-I} & \textbf{AI2D} & \textbf{ChartQA} & \textbf{InfoVQA} & \textbf{OCRBench} & \textbf{ARP} \\
      \midrule
        Full Data & 9.05 & 1350.72 & 64.86 & 77.89 & 63.18 & 73.44 & 49.67 & 71.00 & 100.00 \\
      \midrule
        Random & 9.74 & \textbf{1233.91} & 58.08 & 71.19 & 59.29 & 69.00 & \textbf{45.24} & \textbf{67.90} & 94.30 \\
        \rowcolor{lightcyan}
        QMAC (Ours) & \textbf{11.48} & 1191.27 & \textbf{60.40} & \textbf{71.79} & \textbf{59.59} & \textbf{70.12} & 45.19 & 67.10 & \textbf{96.95} \\
      \bottomrule
      \end{tabular}%
    }
  \end{subtable}

  \vspace{10pt}

  \begin{subtable}{\textwidth}
    \centering
    \caption{10\% data subset.}
    \label{tab:llava_cot_qwen2vl_2b_b}
    \resizebox{\textwidth}{!}{%
      \begin{tabular}{lccccccccccccc}
      \toprule
      \textbf{Method} & \textbf{MATH-Vision} & \textbf{MME} & \textbf{MMB-EN} & \textbf{SQA-I} & \textbf{AI2D} & \textbf{ChartQA} & \textbf{InfoVQA} & \textbf{OCRBench} & \textbf{ARP} \\
      \midrule
        Full Data & 9.05 & 1350.72 & 64.86 & 77.89 & 63.18 & 73.44 & 49.67 & 71.00 & 100.00 \\
      \midrule
        Random & \textbf{9.80} & 1216.21 & 61.08 & 69.26 & 59.52 & 71.04 & 45.98 & 68.90 & 95.25 \\
        \rowcolor{lightcyan}
        QMAC (Ours) & \textbf{9.80} & \textbf{1223.31} & \textbf{62.03} & \textbf{72.19} & \textbf{60.46} & \textbf{71.68} & \textbf{47.28} & \textbf{69.70} & \textbf{96.73} \\
      \bottomrule
      \end{tabular}%
    }
  \end{subtable}

  \vspace{10pt}

  \begin{subtable}{\textwidth}
    \centering
    \caption{15\% data subset.}
    \label{tab:llava_cot_qwen2vl_2b_c}
    \resizebox{\textwidth}{!}{%
      \begin{tabular}{lccccccccccccc}
      \toprule
      \textbf{Method} & \textbf{MATH-Vision} & \textbf{MME} & \textbf{MMB-EN} & \textbf{SQA-I} & \textbf{AI2D} & \textbf{ChartQA} & \textbf{InfoVQA} & \textbf{OCRBench} & \textbf{ARP} \\
      \midrule
        Full Data & 9.05 & 1350.72 & 64.86 & 77.89 & 63.18 & 73.44 & 49.67 & 71.00 & 100.00 \\
      \midrule
        Random & 8.68 & 1282.13 & 59.97 & \textbf{71.99} & 59.49 & 71.52 & 46.41 & \textbf{69.20} & 94.77 \\
        \rowcolor{lightcyan}
        QMAC (Ours) & \textbf{9.08} & \textbf{1286.90} & \textbf{62.20} & 71.64 & \textbf{60.43} & \textbf{72.32} & \textbf{46.86} & 69.00 & \textbf{96.14} \\
      \bottomrule
      \end{tabular}%
    }
  \end{subtable}

\end{table*}

% Table 11

\begin{table*}[htbp]
  \centering
  \caption{Main experimental results on SmallThoughts using Qwen2.5-Math-7B as the target model. The best result in each column is highlighted in bold.}
  \label{tab:smallthoughts_qwen25_math_7b}

  \begin{subtable}{\textwidth}
    \centering
    \caption{10\% data subset.}
    \label{tab:smallthoughts_qwen25_math_7b_10}
    \resizebox{\textwidth}{!}{%
      \begin{tabular}{lccccccccccccccccccc}
      \toprule
      \textbf{Method} & \textbf{AMC23} & \textbf{AQuA} & \textbf{ASDiv} & \textbf{CARP-EN} & \textbf{CMATH} & \textbf{CN Middle School} & \textbf{College Math} & \textbf{GaoKao Math Cloze} & \textbf{Gaokao 2023 EN} & \textbf{GSM8K} & \textbf{MATH} & \textbf{MAWPS} & \textbf{Minerva Math} & \textbf{MMLU-STEM} & \textbf{OlympiadBench} & \textbf{SAT Math} & \textbf{SVAMP} & \textbf{TabMWP} & \textbf{ARP} \\
      \midrule
        Full Data & 17.50 & 28.30 & 56.40 & 29.70 & 58.50 & 12.90 & 11.50 & 15.30 & 24.90 & 42.0 & 26.30 & 59.0 & 6.60 & 21.20 & 7.30 & 28.10 & 47.0 & 27.30 & 100.0 \\
      \midrule
        Random & 7.50 & \textbf{21.30} & 42.20 & 24.40 & 33.0 & 9.90 & 9.90 & 12.70 & 20.50 & 34.0 & 22.20 & 47.20 & 5.10 & 24.50 & 5.60 & 21.90 & 35.0 & 17.20 & 77.22 \\
        \rowcolor{lightcyan}
        QMAC (Ours) & \textbf{17.50} & 19.30 & \textbf{50.80} & \textbf{31.10} & \textbf{46.20} & \textbf{21.80} & \textbf{12.0} & \textbf{18.60} & \textbf{25.70} & \textbf{39.60} & \textbf{24.40} & \textbf{55.60} & \textbf{5.90} & \textbf{25.70} & \textbf{6.80} & \textbf{25.0} & \textbf{42.10} & \textbf{23.90} & \textbf{99.51} \\
      \bottomrule
      \end{tabular}%
    }
  \end{subtable}

  \vspace{10pt}

  \begin{subtable}{\textwidth}
    \centering
    \caption{20\% data subset.}
    \label{tab:smallthoughts_qwen25_math_7b_20}
    \resizebox{\textwidth}{!}{%
      \begin{tabular}{lccccccccccccccccccc}
      \toprule
      \textbf{Method} & \textbf{AMC23} & \textbf{AQuA} & \textbf{ASDiv} & \textbf{CARP-EN} & \textbf{CMATH} & \textbf{CN Middle School} & \textbf{College Math} & \textbf{GaoKao Math Cloze} & \textbf{Gaokao 2023 EN} & \textbf{GSM8K} & \textbf{MATH} & \textbf{MAWPS} & \textbf{Minerva Math} & \textbf{MMLU-STEM} & \textbf{OlympiadBench} & \textbf{SAT Math} & \textbf{SVAMP} & \textbf{TabMWP} & \textbf{ARP} \\
      \midrule
        Full Data & 17.50 & 28.30 & 56.40 & 29.70 & 58.50 & 12.90 & 11.50 & 15.30 & 24.90 & 42.0 & 26.30 & 59.0 & 6.60 & 21.20 & 7.30 & 28.10 & 47.0 & 27.30 & 100.0 \\
      \midrule
        Random & \textbf{15.0} & \textbf{24.40} & 55.70 & 26.70 & 47.80 & \textbf{14.90} & 10.90 & 16.90 & 20.30 & 39.60 & 24.0 & 61.80 & \textbf{7.0} & 9.30 & \textbf{7.60} & 3.10 & 48.0 & 24.20 & 88.37 \\
        \rowcolor{lightcyan}
        QMAC (Ours) & \textbf{15.0} & 21.30 & \textbf{65.60} & \textbf{31.20} & \textbf{57.80} & 9.90 & \textbf{11.90} & \textbf{18.60} & \textbf{24.70} & \textbf{44.0} & \textbf{24.20} & \textbf{70.90} & 4.80 & \textbf{26.20} & \textbf{7.60} & \textbf{28.10} & \textbf{57.50} & \textbf{28.10} & \textbf{101.38} \\
      \bottomrule
      \end{tabular}%
    }
  \end{subtable}

  \vspace{10pt}

  \begin{subtable}{\textwidth}
    \centering
    \caption{30\% data subset.}
    \label{tab:smallthoughts_qwen25_math_7b_30}
    \resizebox{\textwidth}{!}{%
      \begin{tabular}{lccccccccccccccccccc}
      \toprule
      \textbf{Method} & \textbf{AMC23} & \textbf{AQuA} & \textbf{ASDiv} & \textbf{CARP-EN} & \textbf{CMATH} & \textbf{CN Middle School} & \textbf{College Math} & \textbf{GaoKao Math Cloze} & \textbf{Gaokao 2023 EN} & \textbf{GSM8K} & \textbf{MATH} & \textbf{MAWPS} & \textbf{Minerva Math} & \textbf{MMLU-STEM} & \textbf{OlympiadBench} & \textbf{SAT Math} & \textbf{SVAMP} & \textbf{TabMWP} & \textbf{ARP} \\
      \midrule
        Full Data & 17.50 & 28.30 & 56.40 & 29.70 & 58.50 & 12.90 & 11.50 & 15.30 & 24.90 & 42.0 & 26.30 & 59.0 & 6.60 & 21.20 & 7.30 & 28.10 & 47.0 & 27.30 & 100.0 \\
      \midrule
        Random & \textbf{22.50} & \textbf{24.80} & 59.70 & 28.20 & 55.70 & \textbf{15.80} & 11.20 & 11.0 & 21.0 & 39.20 & 24.20 & 63.10 & 5.50 & \textbf{26.20} & 7.70 & \textbf{15.60} & 51.30 & \textbf{28.20} & 97.83 \\
        \rowcolor{lightcyan}
        QMAC (Ours) & 17.50 & 22.0 & \textbf{62.30} & \textbf{32.10} & \textbf{59.20} & 12.90 & \textbf{13.60} & \textbf{14.40} & \textbf{24.40} & \textbf{46.20} & \textbf{26.30} & \textbf{64.80} & \textbf{9.60} & 24.50 & \textbf{7.90} & 12.50 & \textbf{54.30} & 26.70 & \textbf{103.04} \\
      \bottomrule
      \end{tabular}%
    }
  \end{subtable}

\end{table*}

% Table 12
\begin{table*}[htbp]
  \centering
  \caption{Main experimental results on SmallThoughts using Qwen2.5-Math-1.5B as the target model. The best result in each column is highlighted in bold.}
  \label{tab:smallthoughts_qwen25_math_15b}

  \begin{subtable}{\textwidth}
    \centering
    \caption{10\% data subset.}
    \label{tab:smallthoughts_qwen25_math_15b_10}
    \resizebox{\textwidth}{!}{%
      \begin{tabular}{lccccccccccccccccccc}
      \toprule
      \textbf{Method} & \textbf{AMC23} & \textbf{AQuA} & \textbf{ASDiv} & \textbf{CARP-EN} & \textbf{CMATH} & \textbf{CN Middle School} & \textbf{College Math} & \textbf{GaoKao Math Cloze} & \textbf{Gaokao 2023 EN} & \textbf{GSM8K} & \textbf{MATH} & \textbf{MAWPS} & \textbf{Minerva Math} & \textbf{MMLU-STEM} & \textbf{OlympiadBench} & \textbf{SAT Math} & \textbf{SVAMP} & \textbf{TabMWP} & \textbf{ARP} \\
      \midrule
        Full Data & 12.50 & 16.10 & 35.10 & 21.70 & 41.50 & 18.80 & 7.20 & 11.0 & 16.40 & 25.30 & 16.60 & 34.70 & 4.0 & 25.0 & 3.90 & 15.60 & 22.10 & 15.0 & 100.0 \\
      \midrule
        Random & 5.0 & 8.70 & \textbf{42.80} & \textbf{23.80} & \textbf{39.0} & 9.90 & \textbf{9.20} & 5.90 & \textbf{16.10} & \textbf{23.20} & \textbf{16.20} & \textbf{47.70} & 4.0 & 16.20 & \textbf{3.90} & 9.40 & \textbf{36.60} & \textbf{17.0} & 93.48 \\
        \rowcolor{lightcyan}
        QMAC (Ours) & \textbf{10.0} & \textbf{15.40} & 36.30 & \textbf{23.80} & 32.20 & \textbf{12.90} & 8.60 & \textbf{7.60} & 15.30 & 18.70 & 15.10 & 44.10 & \textbf{4.40} & \textbf{18.30} & \textbf{3.90} & \textbf{15.60} & 34.60 & 12.50 & \textbf{96.21} \\
      \bottomrule
      \end{tabular}%
    }
  \end{subtable}

  \vspace{10pt}

  \begin{subtable}{\textwidth}
    \centering
    \caption{20\% data subset.}
    \label{tab:smallthoughts_qwen25_math_15b_20}
    \resizebox{\textwidth}{!}{%
      \begin{tabular}{lccccccccccccccccccc}
      \toprule
      \textbf{Method} & \textbf{AMC23} & \textbf{AQuA} & \textbf{ASDiv} & \textbf{CARP-EN} & \textbf{CMATH} & \textbf{CN Middle School} & \textbf{College Math} & \textbf{GaoKao Math Cloze} & \textbf{Gaokao 2023 EN} & \textbf{GSM8K} & \textbf{MATH} & \textbf{MAWPS} & \textbf{Minerva Math} & \textbf{MMLU-STEM} & \textbf{OlympiadBench} & \textbf{SAT Math} & \textbf{SVAMP} & \textbf{TabMWP} & \textbf{ARP} \\
      \midrule
        Full Data & 12.50 & 16.10 & 35.10 & 21.70 & 41.50 & 18.80 & 7.20 & 11.0 & 16.40 & 25.30 & 16.60 & 34.70 & 4.0 & 25.0 & 3.90 & 15.60 & 22.10 & 15.0 & 100.0 \\
      \midrule
        Random & 7.50 & 9.10 & 36.20 & \textbf{23.70} & 36.70 & 9.90 & \textbf{8.80} & \textbf{9.30} & 12.70 & 21.90 & 14.60 & 40.50 & \textbf{4.0} & 14.70 & \textbf{4.90} & 9.40 & 29.50 & 14.10 & 89.87 \\
        \rowcolor{lightcyan}
        QMAC (Ours) & \textbf{12.50} & \textbf{13.80} & \textbf{38.60} & 22.60 & \textbf{38.50} & \textbf{11.90} & 8.30 & 8.50 & \textbf{15.30} & \textbf{23.60} & \textbf{15.70} & \textbf{44.0} & \textbf{4.0} & \textbf{23.70} & 3.40 & \textbf{12.50} & \textbf{33.0} & \textbf{14.80} & \textbf{98.14} \\
      \bottomrule
      \end{tabular}%
    }
  \end{subtable}

  \vspace{10pt}

  \begin{subtable}{\textwidth}
    \centering
    \caption{30\% data subset.}
    \label{tab:smallthoughts_qwen25_math_15b_30}
    \resizebox{\textwidth}{!}{%
      \begin{tabular}{lccccccccccccccccccc}
      \toprule
      \textbf{Method} & \textbf{AMC23} & \textbf{AQuA} & \textbf{ASDiv} & \textbf{CARP-EN} & \textbf{CMATH} & \textbf{CN Middle School} & \textbf{College Math} & \textbf{GaoKao Math Cloze} & \textbf{Gaokao 2023 EN} & \textbf{GSM8K} & \textbf{MATH} & \textbf{MAWPS} & \textbf{Minerva Math} & \textbf{MMLU-STEM} & \textbf{OlympiadBench} & \textbf{SAT Math} & \textbf{SVAMP} & \textbf{TabMWP} & \textbf{ARP} \\
      \midrule
        Full Data & 12.50 & 16.10 & 35.10 & 21.70 & 41.50 & 18.80 & 7.20 & 11.0 & 16.40 & 25.30 & 16.60 & 34.70 & 4.0 & 25.0 & 3.90 & 15.60 & 22.10 & 15.0 & 100.0 \\
      \midrule
        Random & \textbf{12.50} & 13.80 & 32.10 & 21.40 & \textbf{38.50} & 11.90 & \textbf{8.30} & 8.50 & 15.30 & 22.10 & 14.40 & 34.60 & \textbf{4.80} & \textbf{23.70} & \textbf{3.40} & 12.50 & 25.0 & \textbf{14.70} & 93.60 \\
        \rowcolor{lightcyan}
        QMAC (Ours) & \textbf{12.50} & \textbf{16.50} & \textbf{38.0} & \textbf{22.10} & 36.30 & \textbf{17.80} & 8.20 & \textbf{11.0} & \textbf{15.80} & \textbf{26.20} & \textbf{16.20} & \textbf{42.70} & \textbf{4.80} & 22.10 & 1.90 & \textbf{28.10} & \textbf{29.20} & 13.80 & \textbf{105.03} \\
      \bottomrule
      \end{tabular}%
    }
  \end{subtable}

\end{table*}
% Table 13
\begin{table*}[htbp]
  \centering
  \caption{Ablation of quality and distribution constraints on Vision-Flan using LLaVA-V1.5-7B as the target model. We compare quality-only selection (QUA), the CFD distribution constraint alone, and their combination (TOT). The best result in each column is highlighted in bold.}
  \label{tab:openhermes_constraint_ablation}

  \begin{subtable}{\textwidth}
    \centering
    \caption{5\% data subset.}
    \label{tab:openhermes_constraint_ablation_5}
    \resizebox{\textwidth}{!}{%
      \begin{tabular}{lcccccccccccccc}
      \toprule
      \textbf{Method} & \textbf{GQA} & \textbf{VizWiz} & \textbf{TextVQA} & \textbf{SQA-I} & \textbf{MME} & \textbf{MMB-CN} & \textbf{MMB-EN} & \textbf{AI2D} & \textbf{ChartQA} & \textbf{DocVQA} & \textbf{InfoVQA} & \textbf{MMStar} & \textbf{OCRBench} & \textbf{ARP} \\
      \midrule
        QUA & 41.95 & \textbf{54.39} & 37.30 & 57.76 & 1064.46 & \textbf{42.10} & 38.32 & 38.96 & \textbf{16.32} & 17.99 & \textbf{21.57} & 33.00 & \textbf{29.10} & 96.66 \\
        CFD & 41.35 & 53.88 & \textbf{37.56} & 57.28 & 870.97 & 37.12 & 25.43 & 34.03 & 15.64 & \textbf{18.50} & 21.38 & \textbf{34.16} & 28.30 & 91.89 \\
        TOT & \textbf{42.56} & 54.18 & 37.41 & \textbf{63.81} & \textbf{1173.75} & 41.58 & \textbf{52.32} & \textbf{43.56} & \textbf{16.32} & 17.96 & 19.18 & 32.86 & 28.00 & \textbf{99.16} \\
      \bottomrule
      \end{tabular}%
    }
  \end{subtable}

  \vspace{10pt}

  \begin{subtable}{\textwidth}
    \centering
    \caption{10\% data subset.}
    \label{tab:openhermes_constraint_ablation_10}
    \resizebox{\textwidth}{!}{%
      \begin{tabular}{lcccccccccccccc}
      \toprule
      \textbf{Method} & \textbf{GQA} & \textbf{VizWiz} & \textbf{TextVQA} & \textbf{SQA-I} & \textbf{MME} & \textbf{MMB-CN} & \textbf{MMB-EN} & \textbf{AI2D} & \textbf{ChartQA} & \textbf{DocVQA} & \textbf{InfoVQA} & \textbf{MMStar} & \textbf{OCRBench} & \textbf{ARP} \\
      \midrule
        QUA & 43.18 & 55.45 & 37.62 & 54.98 & 1229.67 & \textbf{48.28} & 38.92 & 37.34 & 15.76 & \textbf{19.63} & \textbf{21.38} & 33.83 & 28.30 & 98.89 \\
        CFD & \textbf{43.46} & 56.63 & 38.43 & 63.51 & \textbf{1253.62} & 40.73 & 50.00 & 44.47 & 14.68 & 18.22 & 18.34 & 34.37 & 27.80 & 99.16 \\
        TOT & 42.85 & \textbf{57.78} & \textbf{38.59} & \textbf{66.98} & 1242.66 & 46.99 & \textbf{51.37} & \textbf{47.60} & \textbf{17.08} & 19.07 & 19.59 & \textbf{36.03} & \textbf{28.60} & \textbf{104.04} \\
      \bottomrule
      \end{tabular}%
    }
  \end{subtable}

  \vspace{10pt}

  \begin{subtable}{\textwidth}
    \centering
    \caption{15\% data subset.}
    \label{tab:openhermes_constraint_ablation_15}
    \resizebox{\textwidth}{!}{%
      \begin{tabular}{lcccccccccccccc}
      \toprule
      \textbf{Method} & \textbf{GQA} & \textbf{VizWiz} & \textbf{TextVQA} & \textbf{SQA-I} & \textbf{MME} & \textbf{MMB-CN} & \textbf{MMB-EN} & \textbf{AI2D} & \textbf{ChartQA} & \textbf{DocVQA} & \textbf{InfoVQA} & \textbf{MMStar} & \textbf{OCRBench} & \textbf{ARP} \\
      \midrule
        QUA & 43.67 & 56.88 & 37.64 & 54.19 & \textbf{1295.19} & 48.88 & 50.52 & 43.46 & 16.36 & 20.68 & 19.55 & \textbf{35.14} & \textbf{29.30} & 102.55 \\
        CFD & 44.37 & \textbf{58.32} & 38.77 & \textbf{64.45} & 1194.43 & 46.39 & 56.44 & 46.60 & \textbf{16.40} & 18.74 & 19.25 & 34.72 & 28.50 & 103.28 \\
        TOT & \textbf{45.26} & 56.10 & \textbf{39.85} & 63.66 & 1239.03 & \textbf{51.63} & \textbf{57.99} & \textbf{48.77} & 16.00 & \textbf{20.75} & \textbf{19.57} & 34.89 & 29.20 & \textbf{106.04} \\
      \bottomrule
      \end{tabular}%
    }
  \end{subtable}

\end{table*}

% Tables 14--16 with unified visual width and font size.
% Required packages:
% \usepackage{booktabs}

% Table 14
\begin{table*}[htbp]
  \centering
  \caption{Ablation of embedding models on OpenHermes using Qwen2.5-3B as the target model. The values 0.6B, 4B, and 8B denote the parameter sizes of the Qwen3-Embedding models. The best result in each column is highlighted in bold.}
  \label{tab:openhermes_embedding_ablation}

  \begin{subtable}{\textwidth}
    \centering
    \caption{10\% data subset.}
    \label{tab:openhermes_embedding_ablation_10}
    \resizebox{\textwidth}{!}{%
      \begin{tabular}{lcccccccccccccc}
      \toprule
      \textbf{Embedding Model} & \textbf{ARC-E} & \textbf{ARC-C} & \textbf{GPQA} & \textbf{IFEval-P} & \textbf{IFEval-I} & \textbf{HumanEval} & \textbf{TriviaQA} & \textbf{SQuAD 2.0} & \textbf{DROP} & \textbf{MATH} & \textbf{MMLU} & \textbf{GSM8K} & \textbf{HotpotQA} & \textbf{ARP} \\
      \midrule
        Full Data & 89.77 & 83.05 & 22.73 & 29.94 & 41.25 & 61.59 & 47.22 & 15.20 & 54.62 & 36.20 & 50.72 & 75.51 & 8.79 & 100.00 \\
      \midrule
        0.6B & \textbf{87.30} & \textbf{80.00} & 11.11 & 26.06 & 38.37 & 64.02 & 47.53 & 16.09 & \textbf{61.72} & \textbf{36.48} & \textbf{51.06} & \textbf{75.66} & \textbf{19.77} & \textbf{105.58} \\
        4B & \textbf{87.30} & 77.29 & 9.09 & \textbf{26.43} & 38.25 & \textbf{64.43} & 47.28 & \textbf{16.46} & 61.34 & 36.36 & 46.96 & 75.44 & 19.00 & 103.52 \\
        8B & \textbf{87.30} & 79.32 & \textbf{13.13} & \textbf{26.43} & \textbf{38.73} & 64.02 & \textbf{47.67} & 14.71 & 60.01 & 36.44 & 50.34 & 75.51 & 18.27 & 104.00 \\
      \bottomrule
      \end{tabular}%
    }
  \end{subtable}

  \vspace{10pt}

  \begin{subtable}{\textwidth}
    \centering
    \caption{20\% data subset.}
    \label{tab:openhermes_embedding_ablation_20}
    \resizebox{\textwidth}{!}{%
      \begin{tabular}{lcccccccccccccc}
      \toprule
      \textbf{Embedding Model} & \textbf{ARC-E} & \textbf{ARC-C} & \textbf{GPQA} & \textbf{IFEval-P} & \textbf{IFEval-I} & \textbf{HumanEval} & \textbf{TriviaQA} & \textbf{SQuAD 2.0} & \textbf{DROP} & \textbf{MATH} & \textbf{MMLU} & \textbf{GSM8K} & \textbf{HotpotQA} & \textbf{ARP} \\
      \midrule
        Full Data & 89.77 & 83.05 & 22.73 & 29.94 & 41.25 & 61.59 & 47.22 & 15.20 & 54.62 & 36.20 & 50.72 & 75.51 & 8.79 & 100.00 \\
      \midrule
        0.6B & \textbf{89.07} & 81.69 & \textbf{17.17} & 26.06 & 38.01 & 61.59 & 47.46 & 15.07 & \textbf{61.58} & \textbf{36.44} & \textbf{53.37} & 75.21 & 16.60 & 104.54 \\
        4B & 88.71 & \textbf{82.03} & 13.13 & \textbf{29.39} & \textbf{41.25} & \textbf{64.02} & 47.67 & 14.83 & 60.42 & 35.88 & 50.43 & 75.74 & 18.12 & 105.51 \\
        8B & 87.48 & 79.66 & 15.15 & 28.28 & 40.41 & \textbf{64.02} & \textbf{47.71} & \textbf{15.87} & 60.25 & 36.28 & 53.01 & \textbf{75.97} & \textbf{18.64} & \textbf{106.89} \\
      \bottomrule
      \end{tabular}%
    }
  \end{subtable}

  \vspace{10pt}

  \begin{subtable}{\textwidth}
    \centering
    \caption{30\% data subset.}
    \label{tab:openhermes_embedding_ablation_30}
    \resizebox{\textwidth}{!}{%
      \begin{tabular}{lcccccccccccccc}
      \toprule
      \textbf{Embedding Model} & \textbf{ARC-E} & \textbf{ARC-C} & \textbf{GPQA} & \textbf{IFEval-P} & \textbf{IFEval-I} & \textbf{HumanEval} & \textbf{TriviaQA} & \textbf{SQuAD 2.0} & \textbf{DROP} & \textbf{MATH} & \textbf{MMLU} & \textbf{GSM8K} & \textbf{HotpotQA} & \textbf{ARP} \\
      \midrule
        Full Data & 89.77 & 83.05 & 22.73 & 29.94 & 41.25 & 61.59 & 47.22 & 15.20 & 54.62 & 36.20 & 50.72 & 75.51 & 8.79 & 100.00 \\
      \midrule
        0.6B & 88.89 & 80.34 & \textbf{16.16} & \textbf{30.87} & \textbf{42.45} & 61.59 & 47.77 & \textbf{16.21} & 60.44 & \textbf{36.52} & 53.94 & \textbf{76.12} & 16.03 & 106.29 \\
        4B & 89.42 & \textbf{81.69} & 11.62 & 27.36 & 40.05 & \textbf{64.02} & 47.70 & 14.25 & 60.28 & 36.22 & 49.85 & 76.04 & 17.71 & 103.63 \\
        8B & \textbf{89.95} & 81.02 & 15.15 & 29.02 & 40.05 & 62.80 & \textbf{47.97} & 13.82 & \textbf{63.12} & 35.78 & \textbf{54.10} & 75.74 & \textbf{19.27} & \textbf{107.19} \\
      \bottomrule
      \end{tabular}%
    }
  \end{subtable}

\end{table*}

% Table 15
\begin{table*}[htbp]
  \centering
  \caption{Ablation of quality models on SmallThoughts using Qwen2.5-Math-7B as the target model. The best result in each column is highlighted in bold.}
  \label{tab:smallthoughts_quality_ablation}

  \begin{subtable}{\textwidth}
    \centering
    \caption{10\% data subset.}
    \label{tab:smallthoughts_quality_ablation_10}
    \resizebox{\textwidth}{!}{%
      \begin{tabular}{lccccccccccccccccccc}
      \toprule
      \textbf{Method} & \textbf{AMC23} & \textbf{AQuA} & \textbf{ASDiv} & \textbf{CARP-EN} & \textbf{CMATH} & \textbf{CN Middle School} & \textbf{College Math} & \textbf{GaoKao Math Cloze} & \textbf{Gaokao 2023 EN} & \textbf{GSM8K} & \textbf{MATH} & \textbf{MAWPS} & \textbf{Minerva Math} & \textbf{MMLU-STEM} & \textbf{OlympiadBench} & \textbf{SAT Math} & \textbf{SVAMP} & \textbf{TabMWP} & \textbf{ARP} \\
      \midrule
        Full Data & 17.50 & 28.30 & 56.40 & 29.70 & 58.50 & 12.90 & 11.50 & 15.30 & 24.90 & 42.00 & 26.30 & 59.00 & 6.60 & 21.20 & 7.30 & 28.10 & 47.00 & 27.30 & 100.00 \\
      \midrule
        Think & \textbf{17.50} & 19.30 & 50.80 & \textbf{31.10} & 46.20 & \textbf{21.80} & \textbf{12.00} & \textbf{18.60} & \textbf{25.70} & \textbf{39.60} & \textbf{24.40} & 55.60 & \textbf{5.90} & \textbf{25.70} & \textbf{6.80} & \textbf{25.00} & 42.10 & \textbf{23.90} & \textbf{99.51} \\
        Instruct & 15.00 & \textbf{20.10} & \textbf{52.70} & 28.90 & \textbf{47.20} & 10.90 & 11.70 & 16.10 & 23.10 & 39.30 & 23.90 & \textbf{58.20} & 5.10 & 23.60 & 5.50 & \textbf{25.00} & \textbf{46.70} & 21.80 & 90.42 \\
      \bottomrule
      \end{tabular}%
    }
  \end{subtable}

  \vspace{10pt}

  \begin{subtable}{\textwidth}
    \centering
    \caption{20\% data subset.}
    \label{tab:smallthoughts_quality_ablation_20}
    \resizebox{\textwidth}{!}{%
      \begin{tabular}{lccccccccccccccccccc}
      \toprule
      \textbf{Method} & \textbf{AMC23} & \textbf{AQuA} & \textbf{ASDiv} & \textbf{CARP-EN} & \textbf{CMATH} & \textbf{CN Middle School} & \textbf{College Math} & \textbf{GaoKao Math Cloze} & \textbf{Gaokao 2023 EN} & \textbf{GSM8K} & \textbf{MATH} & \textbf{MAWPS} & \textbf{Minerva Math} & \textbf{MMLU-STEM} & \textbf{OlympiadBench} & \textbf{SAT Math} & \textbf{SVAMP} & \textbf{TabMWP} & \textbf{ARP} \\
      \midrule
        Full Data & 17.50 & 28.30 & 56.40 & 29.70 & 58.50 & 12.90 & 11.50 & 15.30 & 24.90 & 42.00 & 26.30 & 59.00 & 6.60 & 21.20 & 7.30 & 28.10 & 47.00 & 27.30 & 100.00 \\
      \midrule
        Think & 15.00 & 21.30 & \textbf{65.60} & 31.20 & \textbf{57.80} & 9.90 & 11.90 & \textbf{18.60} & \textbf{24.70} & \textbf{44.00} & 24.20 & \textbf{70.90} & 4.80 & \textbf{26.20} & \textbf{7.60} & \textbf{28.10} & \textbf{57.50} & \textbf{28.10} & \textbf{101.38} \\
        Instruct & \textbf{20.00} & \textbf{22.00} & 62.60 & \textbf{33.10} & 55.30 & \textbf{14.90} & \textbf{12.40} & 15.30 & \textbf{24.70} & 43.30 & \textbf{24.90} & 66.30 & \textbf{6.20} & 22.00 & 6.10 & 21.90 & 56.30 & 25.50 & 100.78 \\
      \bottomrule
      \end{tabular}%
    }
  \end{subtable}

  \vspace{10pt}

  \begin{subtable}{\textwidth}
    \centering
    \caption{30\% data subset.}
    \label{tab:smallthoughts_quality_ablation_30}
    \resizebox{\textwidth}{!}{%
      \begin{tabular}{lccccccccccccccccccc}
      \toprule
      \textbf{Method} & \textbf{AMC23} & \textbf{AQuA} & \textbf{ASDiv} & \textbf{CARP-EN} & \textbf{CMATH} & \textbf{CN Middle School} & \textbf{College Math} & \textbf{GaoKao Math Cloze} & \textbf{Gaokao 2023 EN} & \textbf{GSM8K} & \textbf{MATH} & \textbf{MAWPS} & \textbf{Minerva Math} & \textbf{MMLU-STEM} & \textbf{OlympiadBench} & \textbf{SAT Math} & \textbf{SVAMP} & \textbf{TabMWP} & \textbf{ARP} \\
      \midrule
        Full Data & 17.50 & 28.30 & 56.40 & 29.70 & 58.50 & 12.90 & 11.50 & 15.30 & 24.90 & 42.00 & 26.30 & 59.00 & 6.60 & 21.20 & 7.30 & 28.10 & 47.00 & 27.30 & 100.00 \\
      \midrule
        Think & 17.50 & \textbf{22.00} & \textbf{62.30} & \textbf{32.10} & \textbf{59.20} & \textbf{12.90} & \textbf{13.60} & \textbf{14.40} & \textbf{24.40} & \textbf{46.20} & \textbf{26.30} & \textbf{64.80} & \textbf{9.60} & 24.50 & \textbf{7.90} & 12.50 & \textbf{54.30} & \textbf{26.70} & \textbf{103.04} \\
        Instruct & \textbf{20.00} & 19.70 & 57.80 & 29.30 & 54.50 & \textbf{12.90} & 11.80 & \textbf{14.40} & 22.90 & 40.60 & 23.90 & 63.20 & 7.70 & \textbf{27.70} & 7.30 & \textbf{25.00} & 50.00 & 24.70 & 99.71 \\
      \bottomrule
      \end{tabular}%
    }
  \end{subtable}

\end{table*}

% Table 16
\begin{table*}[htbp]
  \centering
  \caption{Ablation of quality models on LLaVA-CoT using Llama-3.2-11B-Vision-Instruct as the target model. The best result in each column is highlighted in bold.}
  \label{tab:llavacot_quality_ablation}

  \begin{subtable}{\textwidth}
    \centering
    \caption{5\% data subset.}
    \label{tab:llavacot_quality_ablation_5}
    \resizebox{\textwidth}{!}{%
      \begin{tabular}{lccccccccccccc}
      \toprule
      \textbf{Method} & \textbf{MATH-Vision} & \textbf{We-Math} & \textbf{LogicVista} & \textbf{DynaMath} & \textbf{MMStar} & \textbf{MME} & \textbf{MMB-EN} & \textbf{SQA-I} & \textbf{AI2D} & \textbf{ChartQA} & \textbf{InfoVQA} & \textbf{OCRBench} & \textbf{ARP} \\
      \midrule
        Full Data & 17.01 & 31.90 & 29.91 & 16.23 & 56.99 & 1549.25 & 77.92 & 95.44 & 76.94 & 85.68 & 62.29 & 73.50 & 100.00 \\
      \midrule
        Think & 15.06 & 35.06 & 29.46 & \textbf{15.71} & \textbf{53.56} & 1470.20 & \textbf{77.49} & 90.98 & \textbf{74.50} & 84.40 & \textbf{66.38} & \textbf{72.00} & 98.10 \\
        Instruct & \textbf{16.64} & \textbf{39.31} & \textbf{32.37} & 15.43 & 52.70 & \textbf{1486.13} & 76.63 & \textbf{91.03} & 74.29 & \textbf{84.44} & 65.53 & 70.50 & \textbf{100.22} \\
      \bottomrule
      \end{tabular}%
    }
  \end{subtable}

  \vspace{10pt}

  \begin{subtable}{\textwidth}
    \centering
    \caption{10\% data subset.}
    \label{tab:llavacot_quality_ablation_10}
    \resizebox{\textwidth}{!}{%
      \begin{tabular}{lccccccccccccc}
      \toprule
      \textbf{Method} & \textbf{MATH-Vision} & \textbf{We-Math} & \textbf{LogicVista} & \textbf{DynaMath} & \textbf{MMStar} & \textbf{MME} & \textbf{MMB-EN} & \textbf{SQA-I} & \textbf{AI2D} & \textbf{ChartQA} & \textbf{InfoVQA} & \textbf{OCRBench} & \textbf{ARP} \\
      \midrule
        Full Data & 17.01 & 31.90 & 29.91 & 16.23 & 56.99 & 1549.25 & 77.92 & 95.44 & 76.94 & 85.68 & 62.29 & 73.50 & 100.00 \\
      \midrule
        Think & \textbf{17.43} & \textbf{39.25} & \textbf{31.70} & \textbf{15.77} & \textbf{53.85} & \textbf{1542.21} & 77.15 & 91.42 & \textbf{75.81} & \textbf{84.96} & 65.43 & 71.70 & \textbf{101.48} \\
        Instruct & 16.28 & 31.78 & 29.69 & 15.67 & 52.44 & 1486.41 & \textbf{78.01} & \textbf{92.17} & 75.29 & 84.40 & \textbf{65.49} & \textbf{73.70} & 98.13 \\
      \bottomrule
      \end{tabular}%
    }
  \end{subtable}

  \vspace{10pt}

  \begin{subtable}{\textwidth}
    \centering
    \caption{15\% data subset.}
    \label{tab:llavacot_quality_ablation_15}
    \resizebox{\textwidth}{!}{%
      \begin{tabular}{lccccccccccccc}
      \toprule
      \textbf{Method} & \textbf{MATH-Vision} & \textbf{We-Math} & \textbf{LogicVista} & \textbf{DynaMath} & \textbf{MMStar} & \textbf{MME} & \textbf{MMB-EN} & \textbf{SQA-I} & \textbf{AI2D} & \textbf{ChartQA} & \textbf{InfoVQA} & \textbf{OCRBench} & \textbf{ARP} \\
      \midrule
        Full Data & 17.01 & 31.90 & 29.91 & 16.23 & 56.99 & 1549.25 & 77.92 & 95.44 & 76.94 & 85.68 & 62.29 & 73.50 & 100.00 \\
      \midrule
        Think & \textbf{17.27} & 37.82 & 31.92 & \textbf{15.87} & \textbf{54.56} & \textbf{1510.22} & 78.61 & 91.87 & \textbf{75.62} & \textbf{85.20} & \textbf{66.87} & \textbf{75.20} & 101.86 \\
        Instruct & 17.20 & \textbf{46.30} & \textbf{34.60} & 15.57 & 54.37 & 1502.13 & \textbf{79.21} & \textbf{92.36} & 74.77 & 84.92 & 66.67 & 74.50 & \textbf{104.45} \\
      \bottomrule
      \end{tabular}%
    }
  \end{subtable}

\end{table*}

\begin{table}[t]
\centering
\caption{
Data selection cost on Vision-Flan with 186K candidate samples.
All results are measured on four NVIDIA A6000 GPUs and reported in GPU hours.
}
\label{tab:time_cost}
\small
\setlength{\tabcolsep}{6pt}
\renewcommand{\arraystretch}{0.95}
\begin{tabular}{lc}
\toprule
\textbf{Method} & \textbf{GPU Hours} \\
\midrule
QMAC (Ours) & 25.00 \\
XMAS & 35.10 \\
COINCIDE & 23.60 \\
CVS & 21.00 \\
ScalSelect & 16.18 \\
PRISM & 14.73 \\
\bottomrule
\end{tabular}
\end{table}

\end{document}